\documentclass[oneside,12pt]{Classes/RoboticsLaTeX}

\usepackage{amsmath}
\usepackage{amsfonts}
\usepackage{algorithm}
\usepackage{algorithmic}
\usepackage{lettrine}
\usepackage{subfig}
\usepackage{csquotes}
\usepackage[utf8]{inputenc} %
\usepackage[T1]{fontenc}
\usepackage[bitstream-charter]{mathdesign}
\usepackage{graphicx,kantlipsum,setspace}
\usepackage[format=plain, font={it, stretch=1}]{caption}
\usepackage{graphicx}
\usepackage[table,xcdraw]{xcolor}
\usepackage{geometry}

\newcounter{fileAlgorithm}[fileAlgorithm]
\makeatletter
\newenvironment{fileAlgorithm}[1][htb]{%
	\let\c@algorithm\c@fileAlgorithm %
	\renewcommand{\ALG@name}{File}%
	\begin{algorithm}[#1]%
	}{\end{algorithm}}
\makeatother

\hypersetup{colorlinks=true, linkcolor=blue, citecolor=blue}

\ifpdf
    \pdfcatalog {/PageMode (/UseOutlines)
                 /OpenAction (fitbh) }
\fi

\title{Cooperative Assembly with Autonomous Mobile Manipulators in an Underwater Scenario}

\ifpdf
  \author{\href{mailto:toridebraus@gmail.com}{Davide Torielli}}
  \collegeordept{\href{http://www.dibris.unige.it}{DIBRIS - Department of Computer Science, Bioengineering, Robotics and System Engineering}}
  \university{\href{http://www.unige.it}{University of Genova}}
  \crest{\includegraphics[width=30mm]{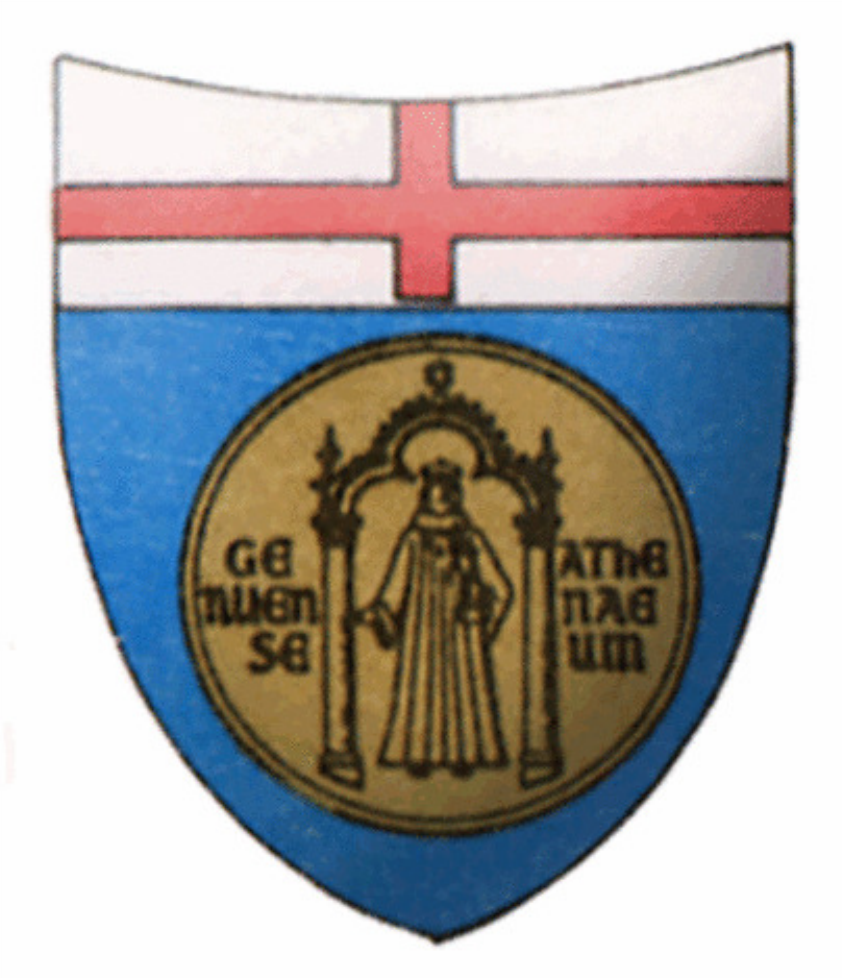}}
\else
  \author{Davide Torielli}
  \collegeordept{DIBRIS - Department of Computer Science, Bioengineering, Robotics and System Engineering}
  \university{University of Genova}
  \crest{\includegraphics[width=30mm]{logo_unige.pdf}}
\fi

\degree{Robotics Engineering}
\degreedate{30th August, 2019}

\begin{document}

\frontmatter
\maketitle
\newpage\null\thispagestyle{empty}\newpage

\setcounter{secnumdepth}{3}
\setcounter{tocdepth}{3}

\begin{dedication}

\begin{center}
	\thispagestyle{empty}
	\vspace*{\fill}
		\begin{flushright}
		``Choose a job you love,\\
		and you will never have to work a day in your life'' \\
		\textit{Confucius (\href{https://quoteinvestigator.com/2014/09/02/job-love/}{maybe})}
		
		\end{flushright}
	\vspace*{\fill}
\end{center}

\end{dedication}

\begin{acknowledgements}

I would like to thank my mother, my father, and my whole family, without who I would not be here. I would like to thank my sweetheart and my love, Sara (aka Cucurucho) which has always supported me and she always will. She was also really important for reviewing my English.\\
I would like to thank the Spanish lab, IRSLab, especially Javier, who really helped me a lot despite every morning at 10:12 he distracted me with Spanish almuerzos. I would like also to thank professors from my home country, professor Giuseppe Casalino, and professor Enrico Simetti, my two thesis supervisors. I want also to remember my Spanish supervisors, Professor Pedro J. Sanz and Professor Raul Marin Prades.\\
I think that I have to thank also my colleague and friend, Fabio. He distracted me, too, with a lot of very long telephone calls. Anyway they were useful for him to learn something from me =). \textit{Maybe}, \textit{sometimes}, they were also useful for me.\\
Last but not least, I have to thank myself, because I'm great. Bravo Tori.

\end{acknowledgements}

\newgeometry{top=2.6cm}
\begin{abstracts}

Robotics is spreading in all the relevant sectors of the human life. The importance of studying this field is confirmed by all the various applications where robots are used: exploration of space and sea, industry, healthcare, transportation and so on. This thesis aims to improve the current state of the art in a particular field: Underwater Robotics. Currently, the research in this area focuses on improving robots capabilities to make them more and more efficient in performing missions autonomously. A particular advancement is towards the cooperation between multiple agents. With cooperation the robotics systems can perform more and more difficult tasks, such as carrying a long and heavy object in an unstructured environment.\\
Specifically, the problem addressed is an \textit{assembly} one known as the \mbox{\textit{peg-in-hole}} task. In this case, two autonomous manipulators must carry \textit{cooperatively} (at kinematic level) a \textit{peg} and must insert it into an \textit{hole} fixed in the environment. Even if the \textit{peg-in-hole} is a well-known problem, there are no specific studies related to the use of two different autonomous manipulators, especially in underwater scenarios. Among all the possible investigations towards the problem, this work focuses mainly on the kinematic control of the robots. The methods used are part of the Task Priority Inverse Kinematics (TPIK) approach, with a cooperation scheme that permits to exchange as less information as possible between the agents (that is really important being water a big impediment for communication). A force-torque sensor is exploited at kinematic level to help the insertion phase. The results show how the TPIK and the chosen cooperation scheme can be used for the stated problem. The simulated experiments done consider little errors in the hole's pose, that still permit to insert the \textit{peg} but with a lot of frictions and possible stucks. It is shown how can be possible to improve (thanks to the data provided by the force-torque sensor) the insertion phase performed by the two manipulators in presence of these errors.\\
Another part of the thesis deals with computer vision algorithms: a third robot exploits  particular methods to estimate the \textit{hole}'s pose. Different techniques are compared to \textit{detect} and to \textit{track} the \textit{hole}, considering the errors they provide in the pose's estimation.\\
Even if the problem is simplified (due to its complexity), this thesis could help further works. The focus is on the particular problem stated, but the methods and tools exploited can be useful also for other applications, not only underwater-related. 

\end{abstracts}

\restoregeometry

\tableofcontents
\listoffigures

\mainmatter

\chapter{Introduction}
\label{chap:introduction}

Nowadays, Robotics is spreading in any considerable area of human life. Fields such as exploration of deep space and sea, healthcare, industrial application, and transportation, show more and more usage of robotics systems. The research has the responsibility to tackle the increasing needs that these many applications have.\\

The underwater environment is one of the fields where Robotics is in a fast escalation, being the sea so important, from industry to environmental issues. Nowadays, different kinds of robot are largely used for underwater missions. A particular type of submarine robot is the Underwater Vehicle Manipulator System (UVMS): an autonomous underwater vehicle (AUV) capable of accomplishing tasks that require a certain level of dexterity, thanks to single or multiple arms.\\
A really innovative field for underwater missions is the analysis of cooperation between multiple agents, which permits to extend their flexibility. Cooperative robots are two or more robots, identical or different, that coordinate themselves to accomplish various objectives, from mapping an area to assembling an object.\\
This thesis focuses on a totally unexplored environment: cooperative \textit{peg-in-hole} assembly with underwater manipulators. It is part of the TWINBOT project \\ \mbox{[\cite{TWINBOT2019}]}, which is devoted to make a step forward for missions in complex scenarios, by extending the robots' capabilities.\\

Part of this thesis is developed at \href{http://www.irs.uji.es/}{IRSLab} at Universitad Jaume I, Castell\`{o}n de la Plana, Spain, during the time I spent as an Erasmus+ 2018/19 student, under the supervision of Professor Pedro J. Sanz and Professor Ra\'{u}l Mar\'{i}n Prades. The IRSLab is the coordinator of the cited TWINBOT project. \\
Another part of the work is done at \href{http://www.graal.dibris.unige.it/}{GRAAL} at Universit\`{a} degli Studi di Genova, Italy, under the supervision of Professor Giuseppe Casalino and Professor Enrico Simetti.\\
The architecture is implemented in C++ and the code is avaible on GitHub at the following link:  \url{https://github.com/torydebra/AUV-Coop-Assembly}.\\
A video of the final experiment is visible at the following link: \url{https://streamable.com/kvoxq} (online; accessed 10-08-2019).\\

This thesis is structured as follows. This Chapter \ref{chap:introduction} introduces the problem, recaps the previous works in this field, and states the objectives of the whole thesis. \mbox{Chapter \ref{chap:control}} defines the theory behind the work, focusing on the Task Priority Inverse Kinematics approach. Chapter \ref{chap:method} applies the explained theory to the actual problem, introducing new methods for the stated problem. In Chapter \ref{chap:results} the simulation set-up is described and results are discussed. Chapter \ref{chap:vision} focuses on the vision part, explaining methods and examining results. In Chapter \ref{chap:conclusions}, conclusions and possible future works are given. In Appendix \ref{chap:AppendixCode} further explanations about the code are given, and in Appendix \ref{chap:AppendixVision} discarded methods (but maybe useful for other purposes) related to computer vision are briefly presented.

\section{State of the Art}
\label{sec:stateArt}
Robots have been massively introduced in various fields to help humans in different tasks. Nowadays, there are various strategies to use them in underwater environments. A manipulator (i.e. a robot's arm) is considered to be the most suitable tool for executing sub-sea intervention operations. Hence, unmanned underwater vehicles (UUVs), such as remotely operated vehicles (ROVs) and autonomous underwater vehicles (AUVs), are equipped with one or more underwater manipulators.

During the last 20 years, underwater manipulators have been used for many different sub-sea tasks in various fields, for example, underwater archaeology [\cite{IntroApp4}; \cite{IntroApp3}], marine geology [\cite{IntroApp1}; \cite{IntroApp2}], and military applications [\cite{IntroApp5}].

There are many specific tasks where the underwater manipulators are important, from salvage of sunken objects [\cite{IntroSpecApp1}] to mine disposal [\cite{IntroSpecApp2}; \cite{IntroSpecApp3}]. One particular scenario is towards the oil and gas industry, where they are used for pipe inspection, opening and closing valves, drilling, rope cutting [\cite{IntroSpecApp4}], and, in general, to reduced field maintenance and development costs [\cite{IntroSpecApp5}]. A recent survey has explored the market related to this technology for oil and gas industry [\cite{IntroSpecApp6}].

\subsection{Previous Works in Underwater Missions}
In the early '90s, research in marine robotics started focusing on the development of underwater vehicle manipulator systems (UVMS). ROVs have been largely used, but they have high operational costs. This is due to the need for expensive support vessels and highly qualified man power effort. In addition, the pilot which operates the vehicle and the arm experiences heavy fatigue in order to carry out the manipulation task. For the above reasons, the research started to increase the effort toward augmenting the autonomy level in underwater manipulation. For example, to reduce the operator's fatigue, some autonomous control features were implemented in work class ROVs [\cite{IntroTeleopRov}]. Another solution is to completely replace ROVs with autonomous underwater vehicles (AUVs).\\

Some pioneering projects in this field have been carried out in the '90s. The AMADEUS project [\cite{IntroAMADEUS1}] developed grippers for underwater manipulation [\cite{IntroAMADEUS2}] and studied the problem of dual arm manipulation [\cite{IntroAMADEUS3}].

The UNION project [\cite{IntroUNION}] was the first to perform a mechatronic assembly of an autonomous UVMS.\\

Early 2000s showed many field demonstrations. The SWIMMER project [\cite{IntroSwimmer1}] developed a prototype autonomous vehicle to deploy a ROV mounted on an AUV. This permitted to remove the need of long umbilical cables and continuous support by vessels on sea surface.

This work was followed by the ALIVE project [\cite{IntroAlive1}; \cite{IntroAlive2}], that achieved autonomous docking of an Intervention-AUV (I-AUV) in a sub-sea structure not specifically created for the AUV use.

The SAUVIM project [\cite{IntroSauvim1}; \cite{IntroSauvim2}] carried out the first autonomous underwater intervention mission. It focused on the searching and recovering of an object whose position was roughly known a priori. Here, the AUV weighted 6 tons, and the arm only 65 kg, so the dynamics of the two subsystems were practically decoupled and the two controllers were separated. The final mission consisted in the AUV performing station keeping while the arm was recovering the object.

After SAUVIM, a project called RAUVI [\cite{IntroRauvi}] took a step further. Here, the AUV performed a hook-based recovery in a water tank. Again, the control of the vehicle and of the arm were separated, even if the Girona 500 light AUV and the small 4-degrees-of-freedom (DOF) arm used had more similar masses than the ones of SAUVIM.

A milestone was the TRIDENT project [\cite{IntroTrident1}]. For the first time, the vehicle and the arm were controlled in a coordinated manner [\cite{IntroTrident2}] to recovery a black-box mockup [\cite{IntroTrident4}]. The used task priority solution dealt with both equality and inequality control objectives, although the inequality ones were only-scalar, except for the joint limits. This permitted to perform some manipulation tasks \textit{while} considering also other objectives, for example, keeping the object centred in the camera frame. In this project, only the kinematic control layer (and not also a suitable dynamic one) was implemented.

The PANDORA project [\cite{IntroPandora1}] focused on increasing the autonomy of the robot, by recognizing failures and responding to them. The work combined machine learning techniques [\cite{IntroPandora2}] and a task priority kinematic control approach [\cite{IntroPandora3}]. However it dealt with only equality control objectives, with a specific ad-hoc solution to manage the joint limits inequality task.

The TRIDENT concepts were enhanced within the MARIS project [\cite{IntroMaris0}]. The used task priority framework [\cite{IntroMaris1}] permitted to \textit{activate} and \textit{deactivate} equality/inequality control objectives of any dimension (not only scalar ones). This project also extended the problem to cooperative agents [\cite{IntroMaris2}].

TRIDENT and MARIS concentrated on using the control framework to perform not only grasping actions. A recent work  [\cite{IntroRecent}] analyses how the method can be used in different scenarios, like pipeline inspection and deep sea mining exploration.

The DexROV project [\cite{IntroDexrov}] is studying latencies problems which arise de-localizing on the shore the manned support to ROV operations.

The PROMETEO project [\cite{IntroPrometeo}] plans to improve the use of underwater robotics in archaeological sites. It investigates the manipulation capacity when occlusions of objects can occur and with a wireless communication system to use the robot without umbilical cable.

The ROBUST project [\cite{IntroRobust}] aims to explore and to map deep water mining sites, through the fusion of two technologies: laser-based in-situ element-analysing, and AUV techniques for sea bed 3D mapping.

\subsection{The Control Framework}
\label{subsec:ControlFramework}
In the '90s, industrial robotics researches focused on how to specify control objectives of a robotics system. This was done especially for redundant systems, i.e.\, systems with more degree of freedoms (DOFs) than necessary. This surplus is useful to perform multiple, parallel tasks; for example, avoiding an obstacle with the whole arm while the end-effector is reaching a goal. Given that such systems need to complete different goals, it has become important to have a simple and effective way to specify the control objectives.\\

The task-based control [\cite{IntroTpik1}], also known as operational space control [\cite{IntroTpik2}], defined the control objectives in a coordinate system that is directly linked to the tasks that need to be performed. This idea was followed by the concept of task priority [\cite{IntroTpik3}]. In this theory, a more important task is executed together with a less important task. To accomplish the whole action, the secondary task is attempted only in the null space of the primary one. This means that the secondary task is executed \textit{only if} it does not go against the accomplishment of the first.

This concept was later generalized to multiple task priority levels [\cite{IntroTpik4}]. These works putted the position control of the end-effector as the highest prioritized one, while safety tasks (like joint limits) were only \textit{attempted} at lower priority level.\\

First studies in control of redundant manipulators [\cite{IntroTpik6}; \cite{IntroTpik5}] managed the free residual DOF in such a way to solve the problem of singularity and obstacle avoidance for an industrial manipulator. Another work [\cite{IntroTpik7}] introduced the use of potential functions in industrial manipulators and mobile robots.

A different solution [\cite{IntroTpik8}] proposed a suboptimal approach. The secondary task was solved as if it was alone, but after it was projected in the null space of the higher priority one. To deal with singularities, a variable damping factor was used [\cite{IntroTpik1}]. This solution was later enhanced and called \textit{null-space-based behavioural control} [\cite{IntroTpik9}].
The approach does not deal with the problem of algorithmic singularities that can occur due to rank loss caused by the projection matrix. Further works [\cite{IntroTpik11}; \cite{IntroTpik10}] focused on this problem.\\

Since those times, the task priority framework has been applied to numerous robotic systems, other than redundant industrial manipulators. Some examples includes mobile manipulators [\cite{IntroTpik12}; \cite{IntroTpik13}; \cite{IntroTpik14}], multiple coordinated manipulators [\cite{IntroTpik15}; \cite{IntroTpik16}], modular robots [\cite{IntroTpik17}], and humanoid robots [\cite{IntroTpik18}; \cite{IntroTpik19}]. Furthermore, a stability analysis for several prioritized inverse kinematics algorithms can be found in [\cite{IntroTpik20}].\\

The problem of the classical task priority framework, evident in all the previous mentioned works, is that inequality control objectives (e.g.\ avoiding
joint limits) were never treated as such. In fact, the corresponding tasks were always active, like the equality ones. So, for example, also when the joints are sufficiently far from their limits, the fact that the task is active uselessly adds constraints and \enquote{consumes} DOFs.
Thus, without a transition, the safety control objectives like joint limits could be only considered as secondary. Otherwise, other mission tasks, like reaching a position with the end-effector, can never be accomplished. This led to an undesired situation where safety tasks have a lower priority with respect to non-safety ones.\\

The challenge in activating (inserting) or deactivating (deleting) a task is that these transitions would imply a discontinuity in the null space projector, which leads to a discontinuity in the control law [\cite{IntroTpik21}]. Thus, in the last decade, researches focused on integrate safety inequality control objectives in a more efficient way. 

A new inversion operator was introduced [\cite{IntroTpik22}] for the computation of a smooth inverse with the ability of enabling and disabling tasks in the context of visual servoing. But the work only dealt with the activation and deactivation of the rows of a single multidimensional task (so, not including the concept of different levels of priority). The extension to the case of a hierarchy of tasks with different priorities was provided successively [\cite{IntroTpik23}]. However, the algorithm requires the computation of all the combinations of possible active and inactive tasks, which grows exponentially as the number of tasks increases.

Another work [\cite{IntroTpik21}] modified the reference of each task that was being inserted or being removed, in order to comply with the already present ones, and in such a way to smooth out any discontinuity. However, the algorithm requires $m!$ pseudo-inverses with $m$ number of tasks. For this reason, the authors provided approximate solutions, which are suboptimal whenever more than one task is being activated or deactivated.

Another approach [\cite{IntroTpik25}] directly incorporated the inequality control objectives as inequality constraints in a Quadratic Program (QP). According to this, the idea was generalized to any number of priority levels [\cite{IntroTpik26}]. At each priority level, the algorithm solves a QP problem, finding the optimal solution (in a least-squares sense). Slack variables are used to incorporate inequality constraints in the minimization process. If the solution contains a slack variable different from zero, it will mean that the corresponding inequality constraint is not satisfied. Otherwise, the inequality constraints are propagated to the next level and transformed into an equality ones (to prevent lower priority tasks from changing the best least-square trade-off found). A similar process is done for the equality constraints. A drawbacks of this approach is that the cascade of QP problems can grow  in dimension rapidly. Another issue is that the activation and deactivation of tasks are not considered. This last point is important when temporal sequences of tasks are used, for example when  assembling objects [\cite{IntroTpik28}; \cite{IntroTpik27}].

Instead of a cascade of QP problems, another research [\cite{IntroTpik29}] proposed to solve a single problem finding the active set of all the constraints at the same time. Due to its iterative nature, the authors proposed to limit the number of iterations to achieve a boundary on the computation time, to be more suitable for a real-time implementation. But this solution is not optimal, and, again, activation/deactivation of equality tasks is not considered.

Improvements are made in the already cited TRIDENT project [\cite{IntroTrident1}; \cite{IntroTrident4}], where field trials proved how to consider activation and deactivation of scalar tasks. But the solution still lacks the ability to deal with activation/deactivation of multidimensional tasks, i.e.\ multiple scalar tasks at the same priority level.

The goals reached by TRIDENT are improved in the MARIS project [\cite{IntroMaris0}; \cite{IntroTpik30}], where, among the other accomplishments, task transitions were successfully implemented in the framework. In particular [\cite{IntroMaris1}], possible discontinuities that could arise are eliminated by a task-oriented regularization and a singular value oriented regularization. Plus, the original simplicity of the task priority framework is retained thanks to pseudo-inverses.

\subsection{The Peg-in-Hole Assembly Problem}
\label{sec:artPeg}
The peg-in-hole is an essential task in assembly processes in various fields, such as manufacturing lines.

This task can be performed following the classical position control method. But this is possible only if precise position of the hole is provided, and the position control error of the robot is zero.
In practice, these conditions can only be obtained in specialized scenario. In the case of more versatile robots, such as underwater manipulators, 
imprecisions and errors are unavoidable.

To deal with these problems, classical works exploit two kind of instruments: cameras and sensors. 
With camera(s), the robot can roughly recognize the objective (i.e.\ the hole) and inspect the overall process. Past researches [\cite{IntroPeg2}; \cite{IntroPeg1}] use this idea to extract boundaries of the object. Another one [\cite{IntroPeg3}] uses visual feedback for a micro-peg-in-hole task (hole of $100 \mu m$).

Other approaches perform precise assembly of the parts thanks to force/torque sensors installed on the wrist. A study [\cite{IntroPeg4}] successfully accomplishes the assembly detecting the force of contact to compensate the positional uncertainty. Newman \textit{et al.} study [\cite{IntroPeg7}] shows how sensors can be used to build map of force and torque values of each contact point.  
In another works [\cite{IntroPeg9}; \cite{IntroPeg8}], the location of the hole was estimated using the measured reaction moment occurred by the contact.
Another good aspect of the sensors  is that they can guarantee stable contact through real-time contact force feedback [\cite{IntroPeg6}; \cite{IntroPeg5}].

Other proposals [\cite{IntroPeg10}; \cite{IntroPeg11}] try to estimate the state of the contact using joint position sensor. This permits to not use the force/torque sensors on the wrist, which would need high control frequency, and would increase overall cost and operation time. 
Some researchers [\cite{IntroPeg12}] show that assembly task can be accomplished without contact force information and with inaccurate vision data. The proposed strategy  mimics the human behaviour: the peg was rubbed in a point close to the object until the relevant objects mated using compliant characteristics. The compliance allows the robot to softly adapt to the hole [\cite{IntroPeg13}; \cite{IntroPeg14}].
A similar, unexpensive, approach is tested experimentally [\cite{IntroPeg15}], without the use of force/torque sensors (i.e.\ no force feedback), nor Remote-center-compliance devices, and with inaccurate hole information.

\section{Motivation and Rationale}
Sea plays an important role in our societies. Many examples are given in the previous section \ref{sec:stateArt}. When such a kind of environment is so important, exploitation of robotic systems is necessary at different levels.\\
This thesis aims to improve the current state of the art in underwater intervention missions. Efforts in this direction can help the robots to accomplish more and more complicated tasks, substituting gradually their remotely operated versions (ROVs), and, in some cases, humans. This would help to reduce the costs, to increase the safety, to boost the performances, and, in general, to accomplish missions that before were unthinkable.

The peg-in-hole is one example of these complicated tasks. In general, robotic assembly problems have been addressed and explored widely, but, to the best of this author's knowledge, no works have been done for cooperatively assembly in underwater scenarios, except for the TWINBOT project [\cite{TWINBOT2019}], which this thesis is part of. Productions related to this problem can help to fill this current lack and can make the technology to advance. 

Cooperative agents augment the capability of the single, for example to carry an heavy object. It is important to notice that cooperation here is intended at \textit{kinematic level}. So, for example, we are not speaking about robotic swarms, where more \enquote{planning} cooperation is explored with Artificial Intelligence techniques. Instead, here cooperation means two (or, in general, more than two) robots that share (in some way) their commanded system velocities (the usual output of the kinematic layer) to move together, without, in this case, make the common tool fall or break. So, there is not an \enquote{high level} reasoning with a planner, but there are mathematical formulas with vectors and matrices to make the robots \textit{cooperate}.\\
Such improvements at kinematic level are important because they reduce the work-effort at higher levels (i.e. the planner), and they make the overall system faster.\\

At the time this thesis was being developed, the TWINBOT project was in an early stage. So, this works evolves autonomously, always keeping an eye on the main objective of the project: improving capabilities of cooperative underwater intervention robots. The aim  is to developed a kinematic control framework suitable for the cooperative underwater \textit{peg-in-hole} problem stated, considering methods that can be used also for other robotics missions.\\
The Task Priority Inverse Kinematics (TPIK) approach is exploited for the single agent kinematic control (section \ref{sec:tpik}), for the arm-vehicle coordination (section \ref{sec:armVehScheme}), and for the cooperation scheme (section \ref{sec:coopScheme}).\\
A force-torque sensor is used to have data from the collisions that happen between the \textit{peg} and the \textit{hole} during the insertion phase. This information is used by two different methods which help to achieve the final goal and to reduce frictional forces that can damage the objects or the robots.\\
The first one is a new objective called Force-Torque objective (section \ref{sec:forceTask}). Its duty is to reduce the forces and the torques that act on the peg, moving it into the opposite direction. The objective is inserted into the TPIK list, among the other objectives. This is noticeable because we exploit a \enquote{dynamic} information at kinematic level.\\
The second method is called Change Goal routine (section \ref{sec:changeGoal}). Its job is to change the origin of the goal frame (that is inside the hole) where the control drives the peg. This is done to compensate possible errors in the hole's pose. For example, if the goal is slightly on the left respect to the exact centre of the hole, lot of collisions happen with the left inner side, so lot of forces directed on the right are detected. This routine simply moves the origin of the goal frame according to the detected forces, on the right in the cited example. 
However, it must be considered that the \textit{peg-in-hole} is simplified: the problem of having too big errors, which makes the peg collide with the external face of the hole, is not considered.\\

Another part of the thesis focuses on computer vision, specifically on the pose estimation of the hole (Chapter \ref{chap:vision}). In particular, stereo-vision methods are used by a third robot, which acts exclusively for the Vision purpose.\\
The Vision routine is divided into two phases.\\
The first is the \textit{detection} (section \ref{sec:visDetect}) where the hole's structure is found in the images that the camera captures. Among all the algorithms tried, one based on shape detection and another based on template matching are discussed in the final results (section \ref{subsec:detectResult}).\\
The second phase, the \textit{tracking} (section \ref{sec:visTracking}), uses the first one as initialization for a \textit{markerless model-based} method. In this case, trials with a mono-camera, a stereo-camera, and depth stereo-camera are compared and discussed.

\chapter{Control Framework}
\label{chap:control}

\section*{Introduction}
In this section, the implemented control framework is presented, along with the necessary mathematical details. The architecture is constituted by two parts:
\begin{itemize}
	\item The Mission Manager, which job is to supervise the execution of the overall mission. In this specific case, it is only a support for the subsequent part. It creates the \textit{action}, that are a lists of control objectives that the Kinematic Control Layer must satisfy during the mission phases. 
	\item The Kinematic Control Layer (KCL) focus on provide the system velocities (i.e. vehicle and joint velocities), given the list of control objectives from the Mission Manager.
\end{itemize}
This thesis focuses mostly on the Kinematic Control Layer. Discussions about how to implement a more complete Mission Manager (which acts not only as a support for the KCL, but, for example, it deals also with the \textit{planning} of the mission phases) go out the scope of this work.\\

The framework is build from the one used in \cite{IntroMaris2}, \cite{tesiWander}, and \cite{IntroRecent}. The sections \ref{sec:definitions}, \ref{sec:controlObjectives}, \ref{sec:tpik}, \ref{sec:armVehScheme}, and \ref{sec:coopScheme} derive from these works and in the following pages their main aspects are recalled.

\section{Definitions}
\label{sec:definitions}
In this section, the necessary definitions are presented. We will consider the general case of a floating manipulator, made up of an $l$-DOF arm attached to a 6-DOF vehicle.\\

\noindent Let us define:
\begin{itemize}
	\item The system configuration vector $ \boldsymbol{c} \in \mathbb{R}^{n}$ of the robot: ~
	$\boldsymbol{c} \triangleq 
		\begin{bmatrix}
			{\boldsymbol{q}} \\ \boldsymbol{\eta}
		\end{bmatrix}$,\\
	where $\boldsymbol{q} \in \mathbb{R}^{l}$ is the l-DOF arm configuration vector and  $\boldsymbol{\eta} \in \mathbb{R}^6$ is the vehicle \emph{generalized coordinate position vector}.\\
	The first three components of $\boldsymbol{\eta}$ form the position vector $\boldsymbol{\eta}_1 \triangleq \begin{bmatrix}x \\ y \\ z\end{bmatrix}$, with components in the inertial frame $\langle w \rangle$.\\
	The last three components of $\boldsymbol{\eta}$ forms the orientation vector $\boldsymbol{\eta}_2 \triangleq \begin{bmatrix}\phi \\ \theta \\ \psi\end{bmatrix}$ expressed in terms of the three angles roll, pitch, yaw (applied in the yaw-pitch-roll sequence \cite{fossenAnglesSeq}). The singularity given by this Euler sequence that arise when $ \theta = \pi/2$ is handled by a specific control objective, the \textit{horizontal attitude} (section \ref{sec:taskList}). As the name suggests, this objective assures the vehicle to stay away from the cited singularity.\\
	From the given definition, it results that $ n = l+6 $
	
	\item The system velocity vector $\dot{\boldsymbol{y}} \in \mathbb{R}^n$ of the robot: ~
	$\dot{\boldsymbol{y}} \triangleq 
	\begin{bmatrix}\dot{\boldsymbol{q}} \\ \boldsymbol{v}\end{bmatrix}$,\\
	where $\dot{\boldsymbol{q}} \in \mathbb{R}^{l}$ are the arm joint velocities and $\boldsymbol{v} \in \mathbb{R}^{6}$ is the vehicle velocity vector.\\
	The first three components of $\boldsymbol{v}$ form the linear velocities $\boldsymbol{v}_1 \triangleq \begin{bmatrix}\dot{x} \\ \dot{y} \\ \dot{z}\end{bmatrix}$ and the last three form the angular velocities $\boldsymbol{v}_2 \triangleq \begin{bmatrix}p \\ q \\ r\end{bmatrix}$, both with components in the vehicle frame $\langle v \rangle$. \\
	The vehicle is considered fully actuated, so the vector $\dot{\boldsymbol{y}}$ coincides with the control vector that is the output of the kinematic layer.
\end{itemize}
	
\section{Control Objectives}
\label{sec:controlObjectives}
Let us consider what the robot has to achieve. An \textit{objective} is a job that the robot must accomplish during the mission. Different objectives can be requested at the same time, for example we want the joints to stay away from their physical limits and the robot to maintain an horizontal attitude, while the end-effector is reaching a desired pose. \\
In the following, specific discussions about these objectives are made.
\subsection{Control Objectives Classification}
\label{sec:coClass}
Control objectives can be classified depending on their importance (i.e. \textit{priority}) in the context of the mission. In fact, some of them can be more important than others. For example, usually we want to avoid an obstacle if this one is in the trajectory given by the reaching goal objective. So, we want \textit{before} to avoid the obstacle and \textit{then} to continue towards the goal. In general, the idea is that the more important objectives have to been satisfied first, and then, \textit{if possible}, also the less important ones. 
A general classification based on the \textit{priority} is given here (from the more important class to the less important one):
\begin{itemize}
	\item \textit{Physical Constraints} objectives. This class includes the interactions with the environment (e.g. not pushing against a rigid surface, imposing a cooperative tool velocity when two robots are carrying together the tool).
	\item \textit{System Safety} objectives. Usually (but not always), the safety is more important than the accomplishment of the Mission. As the name suggest, in this class there are objectives which aim to not damage anything (robot or objects), anyone (humans) and, in general, to not make the whole mission fail. Examples can be staying away from joint mechanical limits, or avoiding an obstacle.
	\item \textit{Prerequisite} objectives. This class is for objectives needed to accomplish the actual mission. An example for a grasping Mission could be focusing the camera on the object to be grasped.
	\item \textit{Mission} objectives, the actual objective that define the mission, like reaching an end-effector position.
	\item \textit{Optimization} objectives. This class is to choose, among all the possible solutions (if multiple ones exist) the best one. For example, this category can be used to improve the energy consumption.
\end{itemize}
The classes of \textit{priority} of these objectives are only a general classification. Depending on the application, objectives can have a different ordering.

\subsection{Equality and Inequality Control Objectives}
\label{sec:eqIneqObj}
Difference between equality and inequality control objectives is another important classification. We consider a variable  $ \boldsymbol{x}(\boldsymbol{c}) \in \mathbb{R}^m $, dependent on the robot configuration vector $ \boldsymbol{c}$, with $ m $ \textit{dimension} of the control objective. \\
The control objective can be of two types:
\begin{itemize}
	\item \textit{Equality control objective}, the requirement, for $t \to \infty$, that \\ \mbox{$\boldsymbol{x}(\boldsymbol{c}) = \boldsymbol{x}_0$}.
	
	\item \textit{Inequality control objective}, the requirement, for $t \to \infty$, that \\ \mbox{$\boldsymbol{x}(\boldsymbol{c}) < \boldsymbol{x}_{max}$ ~ or ~ $\boldsymbol{x}(\boldsymbol{c}) > \boldsymbol{x}_{min}$ ~or ~$ \boldsymbol{x}_{min} < \boldsymbol{x}(\boldsymbol{c}) < \boldsymbol{x}_{max}$}.
\end{itemize}
Please note that here symbols $= , < , >$ refers to element-by-element comparison of the vectors.\\
To make the notations lighter, from now on the dependency of the variable $\boldsymbol{x}(\boldsymbol{c})$ on the configuration vector $\boldsymbol{c}$ will be omitted.\\
The importance of the subdivision between \textit{equality} and \textit{inequality} will be clarified later.

\subsection{Reactive and non Reactive Control Task}
\label{sec:reactNonReact}
Mathematically speaking, the aim of a control objective is to drive the variable $\boldsymbol{x}(\boldsymbol{c})$ toward a point $ \boldsymbol{x}^* $ where the requisite (introduced in the previous section \ref{sec:eqIneqObj}) is satisfied.
For this scope, each control objective has always associated a \textit{feedback reference rate} (in this thesis also denotes as \textit{reference velocity}).\\
An example of \textit{feedback reference rate} is:
\begin{equation}
	\label{feedbackRate}
	\boldsymbol{\dot{{\bar{x}}}} (\boldsymbol{x}) \triangleq \gamma (\boldsymbol{x}^* - \boldsymbol{x}),\quad \gamma > 0
\end{equation}
That is a simple proportional law, where $\gamma$ is a positive gain proportional to the desired convergence rate for the considered variable. From now on, the \textit{feedback reference rate} will have always this formulation.\\

To link the considered variable $ \boldsymbol{x}(\boldsymbol{c})$ to the system velocity vector $\dot{\boldsymbol{y}}$, the following relationship is used:
\begin{equation}
\label{eq:CartJacVel}
	\dot{\boldsymbol{x}} = \boldsymbol{J} \dot{\boldsymbol{y}}
\end{equation} 
that expresses how the system velocity vector $\dot{\boldsymbol{y}}$ influences the rate of change of the variable $\boldsymbol{x}$. $ \boldsymbol{J} \in \mathbb{R}^{m \times n}$ is the so-called \textit{task-induced Jacobian}.\\
The aim of having the actual $\dot{\boldsymbol{x}}$ as much as possible equal to the desired reference $\dot{\bar{\boldsymbol{x}}}$ is called a \textit{reactive control task}.\\

There are situations where a task has not an associated control objective. It happens when an external agent (e.g. an human operator with a console) provides directly the reference rate. So, there is no \textit{feedback reference rate} to calculate, because it is generated by the external agent. In the case of the human operator, it is this one that, seeing (in some way) how the system is behaving, adjust the command, making in its brain a sort of law like the one in formula \eqref{feedbackRate}.  When no control objective is present, we speak about \textit{non-reactive control task}.

\subsection{Inequality Control Objectives Activation and Deactivation}
\label{sec:activations}
During the execution of a mission, not always each inequality control objective is relevant. As an example, maintaining joints away from their mechanical limits is a safety objective which is needed only when the joints are actually near their limits. When a joint is sufficiently far away, there is no necessity to overconstrain the system imposing an additional velocity.\\
To deal with this, we speak about \textit{activation} and \textit{deactivation} of control objectives and of their relative control tasks.
Let us define the following \textit{activation function}:
\begin{equation}
	a(x) \in [0,1]
\end{equation}
as a continuous, sigmoidal, function, which assumes $0$ values within the validity region of the control objective. The validity region is intended as an interval where the objective is satisfied and it is far from not being satisfied any more (with satisfying we intend to accomplish the requirement explained in section \ref{sec:eqIneqObj}). At the margin of this region, a smooth transition from 0 to 1 is present, to gently activate/deactivate the control objective when necessary.\\
For example, considering a scalar ($ m = 1 $) inequality control objective with the requirement $x(\boldsymbol{c}) > x_{min}$ the \textit{activation function} can be defined as:
\begin{equation}
	\label{eq_activation_f}
	a(x) \triangleq
	\begin{cases}1,& x(\boldsymbol{c}) < x_{min}\\
	s(x),& x_{min} \leq x(\boldsymbol{c}) \leq x_{min} + \Delta\\
	0, & x(\boldsymbol{c}) > x_{min} + \Delta\\
	\end{cases}
\end{equation}    
where $s(x)$ is a smooth decreasing function joining the two extreme value $1$ and $0$, and $\Delta$ a value to create a zone where the inequality is satisfied but we want the objective to be activated a bit because we are near the margins. This is also important to prevent chattering problems that would occur if the objective was activated instantaneously.\\
Similar definitions of the activation function can be done for the other two kinds of inequality control objectives.\\

In general, for multidimensional control objectives ($m > 1$), the activation takes the form of a diagonal matrix:
\begin{equation}
\boldsymbol{A} \triangleq
	\begin{bmatrix}
	a_1 & & \\
	& \ddots & \\
	& & a_m	 
	\end{bmatrix}
\end{equation}
where the diagonal elements $a_i$ are activation functions defined similarly to the one in formula \eqref{eq_activation_f}.

Obviously, for the equality control tasks the activation function is a constant equal to $1$, because these tasks are always \enquote{active}. In the multidimensional case, the activation is an identity matrix.\\
Same reasoning can be done for the \textit{non-reactive} control tasks, being absent the variable $x(\boldsymbol{c})$, and being the reference velocities directly given.

\section{Task Priority Inverse Kinematics}
\label{sec:tpik}
In this section, the core of the control architecture is explained.\\
We describe an \textit{Action} $\mathcal{A}$ as a list of prioritized control objectives (with their associated tasks), each one positioned at a defined priority level $k$ (where lower $k$ means higher priority). With this notation, the following symbols are defined:
\begin{itemize}
	\item $\dot{\bar{\boldsymbol{x}}}_k \triangleq \begin{bmatrix}\dot{\bar{x}}_{1,k} & \cdots & \dot{\bar{x}}_{k_m,k}\end{bmatrix}^T$ is the vector of the reference velocities for the control task $k$, with a task dimension $k_m$.
	\item $\dot{\boldsymbol{x}}_k \triangleq \begin{bmatrix}\dot{x}_{1,k} & \cdots & \dot{x}_{k_m,k}\end{bmatrix}^T$ is the current rate of change of the task $k$.
	\item $\boldsymbol{J}_k$ is the Jacobian relationship which relates the current rate-of-change $\dot{\boldsymbol{x}}_k$ with the system velocity vector $\dot{\boldsymbol{y}}$ as in equation \eqref{eq:CartJacVel}.
	\item $\boldsymbol{A}_k \triangleq \textrm{diag}(a_{1,k},  \cdots,  a_{k_m,k})$ is the diagonal matrix composed by the activation functions described in section \ref{sec:activations}.
\end{itemize}
It is important to notice that, in the practice, different objectives can have the same priority. In this case, it is possible to simply stack the vectors and matrices to obtain a objective and a related task $k$ that includes both objectives. For simplicity, but without loss of generality, different objectives will be considered always at different priority levels.\\

The aim of the kinematic layer is to find a system velocity vector $\dot{\bar{\boldsymbol{y}}}$ that satisfies \textit{as much as possible} the requirements of each objective of the action $\mathcal{A}$. Given the presence of different objectives with different priorities, we have to satisfy first the higher priority ones, and then, \textit{if possible}, the lower priority ones. To do this, a sequence of nested minimization problems must be solved: 
\begin{equation}\label{eq:rminproblem}
S_k \triangleq \left\{ \arg \mathrm{R\textrm{-}}\min_{\dot{\bar{\boldsymbol{y}}} \in S_{k-1}} \left\| \boldsymbol{A}_k (\dot{\bar{\boldsymbol{x}}}_k - \boldsymbol{J}_k \dot{\bar{\boldsymbol{y}}}) \right\|^2 \right\},\, k = 1, 2, \ldots, N,
\end{equation}
where $S_0 \triangleq \mathbb{R}^n$, $\:S_{k-1}$ is the manifold of solutions of all the previous tasks in the hierarchy, and $N$ is the total number of priority levels. The notation $\mathrm{R\textrm{-}}\min$ is introduced in \cite{IntroMaris1}, and it indicates a series of regularization to avoid singularities.\\
This problem is the so called \textit{Task Priority Inverse Kinematics} (TPIK). To solve the formula  \eqref{eq:rminproblem}, the so called \textit{iCAT} (inequality Constraints And Task transitions) framework uses the following algorithm \ref{alg:icat}:
\begin{algorithm}[H]
	\caption{iCAT} \label{alg:icat}
	\begin{algorithmic} [1] {\large
		\STATE{$\boldsymbol{\rho}_{0} = \boldsymbol{0}$}
		\STATE{$\boldsymbol{Q}_{0} = \boldsymbol{I}$}
		\vspace{5px}
		\FOR{k=1 \TO N}
		\vspace{5px}
		\STATE{$\boldsymbol{W}_k =  \boldsymbol{J}_k \boldsymbol{Q}_{k-1} (\boldsymbol{J}_k \boldsymbol{Q}_{k-1})^{\#,\boldsymbol{A}_k,\boldsymbol{Q}_{k-1}}$}
				\vspace{5px}
		\STATE{$\boldsymbol{Q}_k = \boldsymbol{Q}_{k-1} (\boldsymbol{I} - (\boldsymbol{J}_k \boldsymbol{Q}_{k-1})^{\#,\boldsymbol{A}_k,\boldsymbol{I}} {\boldsymbol{J}_k \boldsymbol{Q}_{k-1}})$}
		\vspace{5px}
		\STATE{$\boldsymbol{\rho}_k = \boldsymbol{\rho}_{k-1} + \textrm{Sat}\left( \boldsymbol{Q}_{k-1} (\boldsymbol{J}_k \boldsymbol{Q}_{k-1})^{\#,\boldsymbol{A}_k,\boldsymbol{I}} \boldsymbol{W}_k \left(\dot{\bar{\boldsymbol{x}}}_k - \boldsymbol{J}_k \boldsymbol{\rho}_{k-1} \right) \right)$}
		\vspace{5px}
		\ENDFOR 
		\vspace{5px}
		\STATE{  $\dot{\bar{\boldsymbol{y}}} = \boldsymbol{\rho}_N$  }
	}
	\end{algorithmic}
\end{algorithm}

\noindent The special pseudo inverse operator $\boldsymbol{(\cdot)}^{\#,\boldsymbol{A},\boldsymbol{Q}}$ [\cite{IntroMaris1}] manages some invariance problems of \eqref{eq:rminproblem}; the function Sat($\cdot$) [\cite{antoSat}] controls the variable saturations. Details of the procedure can be found again in [\cite{IntroMaris1}].

\subsection{Notes on Conflicting Objectives}
From section \ref{sec:coClass}, it should be understood that lower priority tasks are not always satisfied. The problem with this arises when the main mission objective (e.g. reaching a point with the end effector), that is not at the higher priority, can't be never accomplished, thus failing the general mission. This can be the case when an obstacle is met: the robot may stuck in a point of \textit{local minima} (that is however better than crash into it). This means that the robot can't find a trajectory toward the goal and, \textit{at the same time}, avoids the obstacle. So, it remains still because avoiding the obstacle is an objective with more priority. This is a general problem of all reactive control methods. The solution must be found at the mission manager level, which should plan another trajectory (e.g. with intermediate way-points) and/or another sequence of Actions. This problem is not considered in this work.

\section{Arm-Vehicle Coordination Scheme}
\label{sec:armVehScheme}
\begin{figure}[H]
	\begin{center}
		\includegraphics[width=0.90\columnwidth]{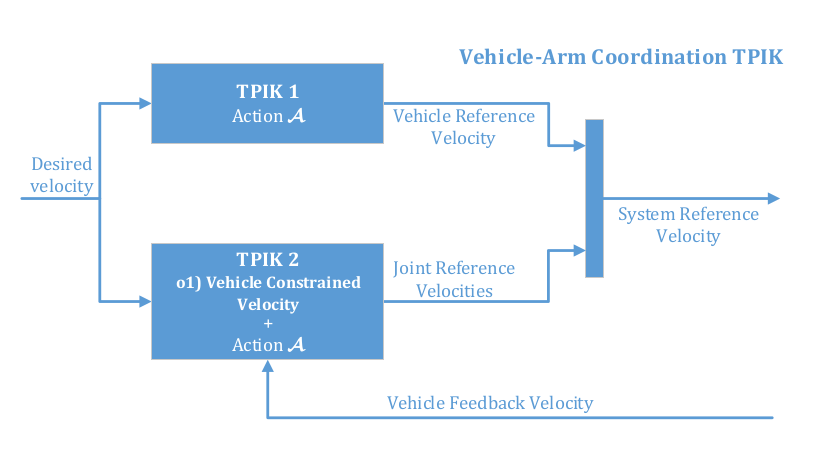}
		\caption[Arm-Vehicle Coordination Scheme in the TPIK]{A scheme showing the two Task Priority Inverse Kinematics blocks for the arm-vehicle coordination}\label{fig:veharmcoord}
	\end{center}
\end{figure}
Inaccuracies in velocity tracking of the vehicle can have effects on the arm. 
A relevant problem arises when disturbances of the floating base, caused by thrusters and/or its large inertia, propagate and affect the end effector motions [\cite{IntroMaris2}].
To solve this, a kinematic decoupling of the arm and the base is done, implementing it within the task priority approach.\\
As in the previous sections, we consider an Action $\mathcal{A}$, that is a list of prioritized objectives to be satisfied.
The idea is to have two TPIK running in parallel as shown in figure \ref{fig:veharmcoord}:
\begin{itemize}
	\item \textbf{TPIK 1}. It considers the vehicle together with the arm as a whole full controllable system. From its output $\dot{\bar{\boldsymbol{y}}}$, only the vehicle velocity component are taken (discarding the arm ones).
	\item \textbf{TPIK 2}. It considers the vehicle as totally non controllable. So, a \textit{non-reactive} task (\ref{sec:reactNonReact}) is added at the top of the priority list $\mathcal{A}$ to \textit{constrain} the output vehicle velocity to the real one (measured in some way). The other objectives of $\mathcal{A}$ remain unchanged. From its output $\dot{\bar{\boldsymbol{y}}}$, only the arm part is taken.
\end{itemize}
At the end of the procedure, the two parts of $\dot{\bar{\boldsymbol{y}}}$ are put togheter to compose the final system reference velocity vector.\\
Thanks to the TPIK 2, the joint velocities are \textit{optimized}, in the sense that they follow \textit{at best} the objectives of the action $\mathcal{A}$ considering also the \textit{measured} vehicle velocity and its influence on the objectives.	

For real mobile manipulators, in general, a multi-rate control of arm and vehicle is used, which means that velocities for the arm and for the vehicle are given at different frequency. This is common because usually the arm can be controlled more precisely and its performance are better than the base. The coordination schema proposed here is suitable for such an implementation: the TPIK 2 can run at higher frequency, updating the commanded arm velocities more frequently than the vehicle ones.

\section{Cooperation Scheme}
\label{sec:coopScheme}
\begin{figure}[H]
	\centerline{
		\includegraphics[width=0.8\columnwidth]{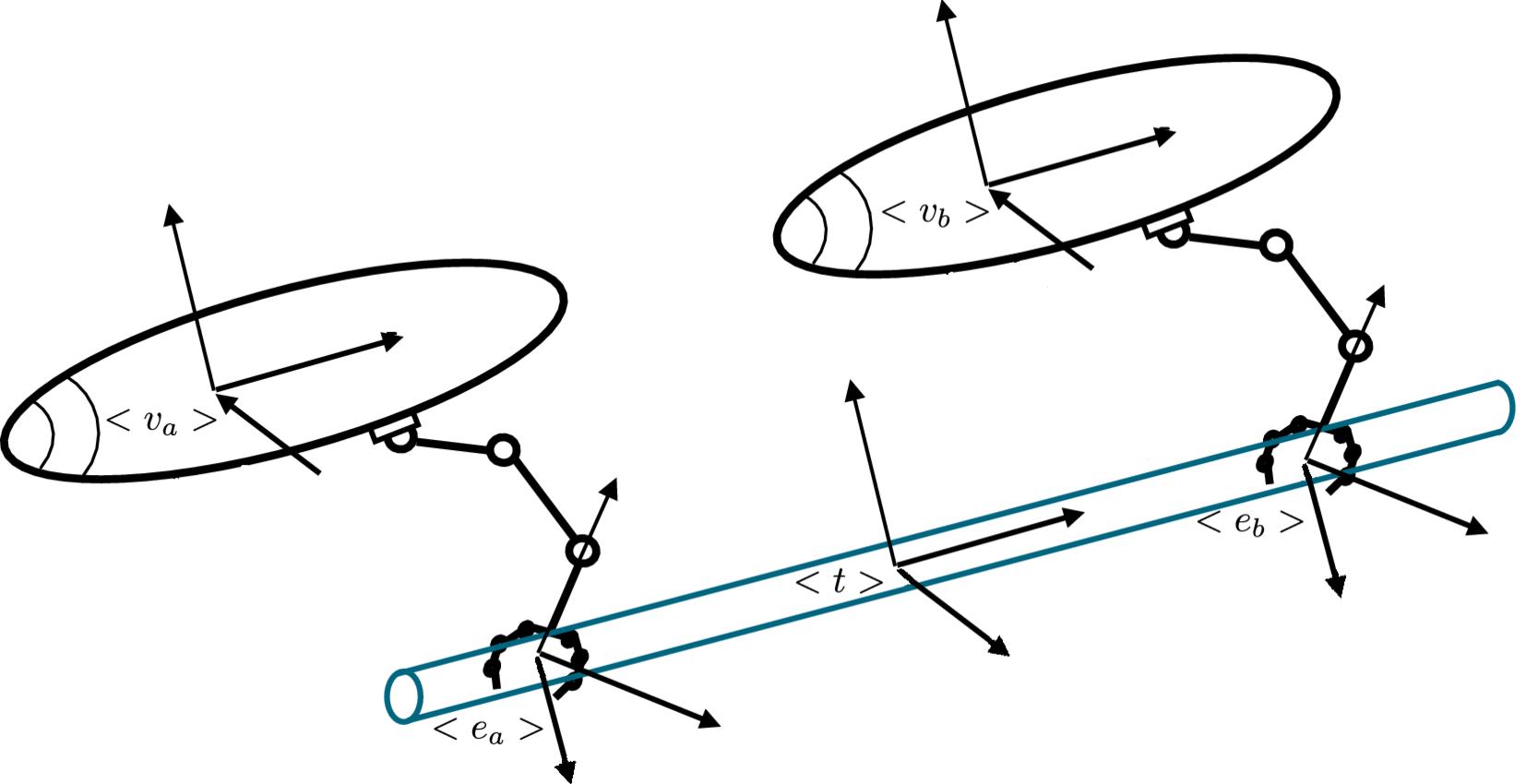}
	}
		\caption[Relevant frames for the cooperation scheme]{The frames of the two cooperative vehicles carrying a common object}\label{fig:coopFrames}
\end{figure}

In this section, the discussion about cooperation is explained. We limit the explanations to only two cooperative robotic systems, but, in general, more agents can be considered.\\

The \textit{cooperation} is used to carry a common tool with the two manipulators, without making it fall or break. This is done at kinematic level: the scheme provides suitable system velocities $\dot{\bar{\boldsymbol{y}}}_a$ and $\dot{\bar{\boldsymbol{y}}}_b$ for the two robots considering the constraint given by the carried common object. So, both system velocities $\dot{\bar{\boldsymbol{y}}}_a$ and $\dot{\bar{\boldsymbol{y}}}_b$ must cause same Cartesian velocity $\dot{{\boldsymbol{x}}}_t$ to the tool.\\
The coordination policy that will be presented takes care of the bandwidth restriction typical of underwater scenarios. Thus, it deals with the cooperation in a decentralized manner, limiting as much as possible the amount of data exchanged.\\
Furthermore, this scheme is different from the classical \textit{leader-follower} ones, because, as we will see, the \enquote{leadership} changes based on the difficulties in tracking the ideal tool velocity that one robot can meet.\\

It is assumed that the object is held firmly by both agents, so no sliding happens during the missions. The two robots agree on a shared fixed frame, so, their respective tool frames $\langle t_a \rangle$ and $\langle t_b \rangle$ and the object frame $\langle o \rangle$ are coincident: 
\begin{equation} 
\langle t \rangle \triangleq \langle t_a \rangle = \langle t_b \rangle = \langle o \rangle
\end{equation}
In figure \ref{fig:coopFrames} the main frames related to the cooperation are shown.\\

\noindent The firm grasp assumption imposes that:
\begin{equation}\label{eq:coopintro}
	\dot{\boldsymbol{x}}_t = \boldsymbol{J}_{t,a} \dot{\boldsymbol{y}}_a = \boldsymbol{J}_{t,b} \dot{\boldsymbol{y}}_b
\end{equation}
with $\dot{\boldsymbol{x}}_t$ the object velocity with component on $\langle t \rangle$; $\dot{\boldsymbol{y}}_a$, $\dot{\boldsymbol{y}}_b$ the system velocity vectors of agents $a$ and $b$ (introduced in section \ref{sec:definitions}); and $\boldsymbol{J}_{t,a}$, $\boldsymbol{J}_{t,b}$ the system Jacobians of agents $a$ and $b$ with respect to $\langle t \rangle$. These Jacobians tell how the tool velocity $\dot{\boldsymbol{x}}_t$ is affected by the system velocities $\dot{\boldsymbol{y}}_a$ and $\dot{\boldsymbol{y}}_b$. Due to the firm grasp assumption, the tool velocities caused by $\dot{\boldsymbol{y}}_a$ and $\dot{\boldsymbol{y}}_b$ must be equal.\\

\noindent Let us rewrite the second part of equation \eqref{eq:coopintro} as:
\begin{equation}\label{eq:coopintro2}
	\begin{gathered}
	\begin{bmatrix}
	\boldsymbol{J}_{t,a} & -\boldsymbol{J}_{t,b}
	\end{bmatrix}
	\begin{bmatrix}
	\dot{\boldsymbol{y}}_a \\ \dot{\boldsymbol{y}}_b
	\end{bmatrix}
	\triangleq \boldsymbol{G}\dot{\boldsymbol{y}}_{ab}=0 \quad \Longleftrightarrow \quad \dot{\boldsymbol{y}}_{ab} \in ker(\boldsymbol{G})
	\end{gathered}
\end{equation}
$ker(\boldsymbol{G})$ represents the subspace where $\dot{\boldsymbol{y}}_{ab}$ is constrained to lay for the firm grasp assumption.\\

We could consider the two manipulators as a unique system simply stacking correctly vectors and matrices. To transport cooperatively the tool, an additional \textit{physical constraint} objective would be added to the TPIK list, to ensure the control outputs a command $\dot{\bar{\boldsymbol{y}}}$ which satisfies the constraint \eqref{eq:coopintro2}. In practice, with this new objective, in the minimization problems of \eqref{eq:rminproblem} we would have $S_1 = ker(\boldsymbol{G})$.\\
The problem with following this way is that we are not considering that the two vehicles are separate entities. This idea would be feasible when the agent is a single robot with two arms. Instead, in this case, exchanging all the vectors and matrices between the robots during the TPIK procedure would not be possible, especially in an underwater situation. So, another method must be considered.\\

\noindent The equation \eqref{eq:coopintro} can be expressed in the Cartesian space as:
\begin{equation}
	\dot{\boldsymbol{x}}_t = \boldsymbol{J}_{t,a} \boldsymbol{J}^\#_{t,a} \dot{\boldsymbol{x}}_t =  \boldsymbol{J}_{t,b} \boldsymbol{J}^\#_{t,b} 
	\dot{\boldsymbol{x}}_t 
\end{equation}
\begin{equation}
\label{eq:constrainMatrixC}
	(\boldsymbol{J}_{t,a} \boldsymbol{J}^\#_{t,a} - \boldsymbol{J}_{t,b} \boldsymbol{J}^\#_{t,b}) 
	\dot{\boldsymbol{x}}_t \triangleq \boldsymbol{C} \dot{\boldsymbol{x}}_t = \boldsymbol{0}
\end{equation}
\begin{equation}
	\dot{\boldsymbol{x}}_t \in ker(\boldsymbol{C}) = Span(\boldsymbol{I} - \boldsymbol{C}^\#\boldsymbol{C})
\end{equation}
$\boldsymbol{C}$ is a particular matrix called \textit{Cartesian Constraint Matrix}. The kernel of $\boldsymbol{C}$ expresses the space of achievable object velocities at the current configurations of the two robots.\\

The idea of the scheme is to put a non-reactive task at the top of the hierarchy, to constrain the desired object velocity $\boldsymbol{\dot{\tilde{x}}}$ in this subspace. After this constraint, we are sure, at kinematic level, that both agents can follow this desired object velocity despite the possible different situations caused by the other objectives and by the different robots configurations.\\
\begin{figure}[H]
	\centering
	\centerline{
		\includegraphics[width=1.22\columnwidth]{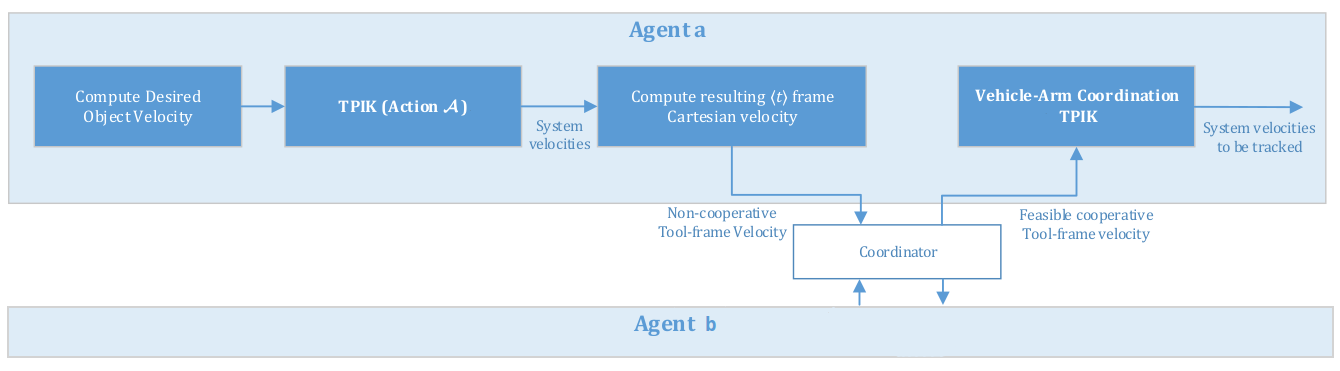} }
	\caption[Cooperation Scheme in the TPIK]{The cooperation scheme with its different steps. The Agent b block is equal to the Agent a one.}
	\label{fig:coopScheme}
\end{figure}
\vspace{10px}
\noindent The scheme, sketched in figure \ref{fig:coopScheme} proceeds as follows:
\begin{itemize}
	\item In the first step, the two agents calculate system velocities (using the TPIK explained in  section \ref{sec:tpik}) as if they were alone. So, we have:
	\begin{equation}
		\dot{\boldsymbol{x}}_{t,a} = \boldsymbol{J}_{t,a} \dot{\boldsymbol{y}}_{a} , \qquad 
		\dot{\boldsymbol{x}}_{t,b} = \boldsymbol{J}_{t,a} \dot{\boldsymbol{y}}_{b}	
	\end{equation}
	where, in general, the two \textit{non cooperative} tool velocities are different:\\ \mbox{$\dot{\boldsymbol{x}}_{t,a} \neq \dot{\boldsymbol{x}}_{t,b}$.}
	
	\item The tool velocities are exchanged (i.e. they are sent to the coordinator) and a \textit{cooperative} tool velocity $\dot{\hat{\boldsymbol{x}}}_t$ is computed as:
	\begin{equation}\label{eq:weightsum}
		\dot{\hat{\boldsymbol{x}}}_t = \dfrac{1}{\mu_a + \mu_b} (\mu_a \dot{\boldsymbol{x}}_{t,a}  + \mu_b \dot{\boldsymbol{x}}_{t,b}), \qquad
		\mu_a , \mu_b > 0	
	\end{equation}
    \begin{equation}
		\begin{gathered}
			\mu_a = \mu_0 + \| \dot{\bar{\boldsymbol{x}}}_t - \dot{\boldsymbol{x}}_{t,a} \| \triangleq \mu_0 + \| \boldsymbol{e}_a \|, \\
			\mu_b = \mu_0 + \| \dot{\bar{\boldsymbol{x}}}_t - \dot{\boldsymbol{x}}_{t,b} \| \triangleq \mu_0 + \| \boldsymbol{e}_b \|, \\
			\mu_0 > 0
	    \end{gathered}
	\end{equation}
	where $\dot{\bar{\boldsymbol{x}}}_t$ is the ideal velocity that, if applied, would asymptotically take the tool to the desired goal.\\
	The \textit{cooperative} tool velocity $\dot{\hat{\boldsymbol{x}}}_t$  is a \textit{weighted} compromise between the two \textit{non cooperative} ones. The \textit{weights} $\mu_a, \mu_b$ give more freedom to the robot which meet the highest error $\boldsymbol{e}$. This error is a way to understand how much one robot is in difficult in tracking the \textit{ideal} tool velocity $\dot{\bar{\boldsymbol{x}}}_t$.
	
	\item The new \textit{cooperative} tool velocity $\dot{\hat{\boldsymbol{x}}}_t$ is not, in general, a \textit{feasible} velocity that both vehicle can provide to the tool. So, an additional passage is required:
	\begin{equation}
		\dot{\tilde{\boldsymbol{x}}}_t \triangleq \big( \boldsymbol{I} - \boldsymbol{C}^\# \boldsymbol{C} \big) \dot{\hat{\boldsymbol{x}}}_t
	\end{equation}
	with $\boldsymbol{C}$ defined in \eqref{eq:constrainMatrixC}.
	
	\item Each agent runs a new TPIK procedure, with a objective list identical to the first one, but with a \textit{non-reactive} control objective at the top to track the \textit{feasible cooperative} velocity $\dot{\tilde{\boldsymbol{x}}}_t$. The outputs of this procedure, $\dot{\hat{\boldsymbol{y}}}_a$ and $\dot{\hat{\boldsymbol{y}}}_b$ will be the final velocities which the kinematic layer provides.\\
	Moving the equality control objective to make the end effector reach the goal at the top of the hierarchy does not influence the safety tasks. This property is proven in \cite{tesiWander}.
\end{itemize}
\vspace{15px}
The method assumes that the Coordinator can calculate the ideal tool velocity $\dot{\bar{\boldsymbol{x}}}_t$, so it must know the transformation matrix between the goal frame $\langle g \rangle $ and the tool frame $ \langle t \rangle $.\\

It can be noticed how the only information that the agents must exchange are the 
\textit{non-cooperative} velocities $\dot{\boldsymbol{x}}_{t,a}$ and $\dot{\boldsymbol{x}}_{t,b}$, the matrices $\boldsymbol{J}_{t,a} \boldsymbol{J}^{\#}_{t,a}$ and $\boldsymbol{J}_{t,b} \boldsymbol{J}^{\#}_{t,b}$ (to build the Cartesian Constraint Matrix $\boldsymbol{C}$), and the feasible velocity $\dot{\tilde{\boldsymbol{x}}}_t$. Less data can be exchanged if we have Jacobians expressed analytically. In this case, instead of sharing the two $6 \times 6$ matrices, we can share only two configuration vectors $\boldsymbol{c}$ ($n \times 1$), and make the coordinator calculate the Jacobians from their analytical expressions.\\
Even less data can be shared if the \textit{coordinator} is a procedure that runs on a robot, and it is not on an external node. In this case, practically only half of the amount of data must be shared through water.

\chapter[Control Architecture: Methods]{Control Architecture: Methods}

In this chapter, the theory explained in the previous Chapter \ref{chap:control} is exploited to deal with the scenario stated for this thesis.\\
In Section \ref{sec:forceTask}, a new control objective, called \textit{Force-Torque} objective, is added to the Action list of the TPIK approach. The aim of this new objective is to drive the peg \enquote{away} from collisions that may happen during the insertion phase. Thanks to force-torque data given by a sensor, this will help the mission, reducing the amount of collisions between the peg and the hole. It is important to notice that we are exploiting force-torque information at kinematic level.\\
In section \ref{sec:taskList}, a list of objectives, including the new Force-Torque one, is presented. This composes the Action $\mathcal{A}$ that must be accomplished for the stated mission.\\
Section \ref{sec:changeGoal} describes another (additional) method which exploits the data given by the force-torque sensor. In brief, this new \textit{routine}, called Change Goal, shifts the origin of the goal frame (which is inside the hole) according to the forces detected on the peg. The aim is to compensate the error given by a not perfectly estimation of the hole's pose.

\section{Force-Torque Objective}
\label{sec:forceTask}
Information from a force torque sensor can be exploited at kinematic level, inserting an additional control objective into the TPIK procedure. The aim of this objective is to zeroing the forces and torques acting on the peg. This is done by generating properly joints and vehicle velocities to drive the peg in such a way the forces and torques decrease. This objective is similar to the one used for pipeline weld inspection in \cite{IntroRecent}.\\
If we visualize the resultant of the forces on the peg caused by the collisions as a vector, moving \textit{linearly} the peg along this vector will cause the forces to decrease. The same idea can be use with torques, \textit{rotating}, along the resultant vector instead of moving linearly.\\

\noindent The \textit{feedback reference rate} for this objective will be:
\begin{equation}
	\label{eq:refForceTask}
	\boldsymbol{\dot{\bar{x}}}_{ft} \triangleq \begin{bmatrix}{\dot{\bar{x}}}_f \\ {\dot{\bar{x}}}_m \end{bmatrix} \triangleq 
	\begin{bmatrix} \gamma_f \\ \gamma_m \end{bmatrix} 
	\begin{bmatrix}
		0 - \| \boldsymbol{f} \| \\ 0 -\| \boldsymbol{m} \|
	\end{bmatrix} \qquad 0 < \gamma_f < 1, \quad 0 < \gamma_m < 1
\end{equation}
where $\| \boldsymbol{f} \|$ and $\| \boldsymbol{m} \|$ are the norms of the forces and torques vectors $\boldsymbol{f}$ and $\boldsymbol{m}$. Gains smaller that $1$ are necessary to reduce the amount of speed requested. In fact, for example, if a force with a norm of 1 N (which is relatively small, so common to be detected) was present, a gain equal to 1 would mean to request a tool speed of 1 m/s, that is an exaggeration for this case.\\
It can be noticed that, instead of the full 3-dimensional vectors $\boldsymbol{f}$ and $\boldsymbol{m}$, the norms $\| \boldsymbol{f} \|$ and $\| \boldsymbol{m\textbf{}} \|$ are used. This is done to not overconstrain the system and to let more freedom to lower priority task. Even with norms, the collisions are reduced, because when we bring to zero the norms, also each component of the vector tends to zero.\\

The \textit{feedback reference rate} of equation \eqref{eq:refForceTask} is a velocity that the tool must follow. So, the Jacobian must be built considering this fact. For the \textit{task-induced} Jacobian (section \ref{sec:tpik}) of this new task, we have to split the linear and the angular part of the tool Jacobian $\boldsymbol{J}_t$. Then, due to the fact that we are considering the norms, we have to pre-multiplying the two parts for the normal vector of $\boldsymbol{f}$ and $\boldsymbol{m}$ transposed:
\begin{equation}
	\label{eqJacobFor}
	\boldsymbol{J}_{ft} \triangleq \begin{bmatrix}{\boldsymbol{J}}_f \\[1em] {\boldsymbol{J}}_m \end{bmatrix} \triangleq 
	\begin{bmatrix} \left( - \; \dfrac{\boldsymbol{f}}{\| \boldsymbol{f} \|}\right)^T \enspace ^{lin}\boldsymbol{J}_t  \\[1.5em]
		\left( - \; \dfrac{\boldsymbol{m}}{\| \boldsymbol{m} \|}\right) ^T \enspace ^{ang}\boldsymbol{J}_t  \end{bmatrix} 
\end{equation}
where $\boldsymbol{J}_f, \boldsymbol{J}_m \in \mathbb{R}^{1\times l}$; $\;\;\; \boldsymbol{J}_t$ is the Jacobian which express how the Cartesian tool velocity $\dot{\boldsymbol{x}}_t$ is affected by the system velocity vector  $\dot{\boldsymbol{y}}$; $\;\;lin, ang$ superscripts refer to \textit{linear} (top three rows) and \textit{angular} (bottom three rows) parts of $\boldsymbol{J}_t$.\\

This objective can be considered as a \textit{pre-requisite} one (section \ref{sec:coClass}), because it is better to reduce the collisions \textit{before} going on with the insertion, also to avoid stuck. In truth, this kind of objective could be also considered as a Physical Constraints one, like in \cite{IntroRecent}. In this case, the first choice is made. In both cases, it is always put at higher priority than the \textit{reaching goal objective}. This will cause the robot to, first, try to nullified the forces and torques (if collisions happened), and only after (i.e. \textit{if possible}) to move the peg towards the goal.\\
Deactivating the task is necessary when the forces and/or torques are zero, to not generate system velocities for this task when they are not necessary. So a smooth activation function $\boldsymbol{A} \in \mathbb{R}^{2 \times 2}$ is used (similarly to the generic one of section \ref{sec:activations}):
\begin{equation}
	\begin{gathered}
		\boldsymbol{A}_{ft} \triangleq
		\begin{bmatrix}
			a_f & 0 \\
			0 & a_t \\
		\end{bmatrix} \\
		\vspace{10px}
		a_f(\| \boldsymbol{f}\|) \triangleq
		\begin{cases}
			0,& \| \boldsymbol{f}\| = 0\\
			s(\| \boldsymbol{f}\|), & 0 < \| \boldsymbol{f}\| \leq 0 + \Delta\\
			1, & \| \boldsymbol{f}\| > 0 + \Delta\\
		\end{cases} \\
		\vspace{10px}
		a_m(\| \boldsymbol{m}\|) \triangleq
		\begin{cases}
			0,& \| \boldsymbol{m}\| = 0\\
			s(\| \boldsymbol{m}\|), & 0 < \| \boldsymbol{m}\| \leq 0 + \Delta\\
			1, & \| \boldsymbol{m}\| > 0 + \Delta\\
		\end{cases}
	\end{gathered}
\end{equation}  
where $s(\cdot)$ is a smooth \textit{increasing} function from 0 to 1, and $\Delta$ a constant to create the smooth zone.\\
In this case, the activation function is also important for a mathematical detail. In fact, we can see from the Jacobian formula \eqref{eqJacobFor}, that the norms are in the denominator. When they are near to zero, numerical issue (such as too big values in the Jacobian) may happen. This is a common problem when a task is used with the norm. But an easy solution is to deactivate the task when the norm is below a little value $\epsilon$. In this case, for the force part:
\begin{equation}
		a_f(\| \boldsymbol{f}\|) \triangleq
		\begin{cases}
			0,& \| \boldsymbol{f}\| \leq \epsilon \\
			s(\| \boldsymbol{f}\|), & \epsilon < \| \boldsymbol{f}\| \leq 0 + \Delta\\
			1, & \| \boldsymbol{f}\| > 0 + \Delta\\
		\end{cases} \\
\end{equation}
The activation for the torque is similar.\\

Considering the minimization problem presented with the formula \eqref{eq:rminproblem}, for this specific objective the equation will be:
\begin{equation}
	S_k \triangleq \left\{ \arg \mathrm{R\textrm{-}}\min_{\dot{\bar{\boldsymbol{y}}} \in S_{k-1}} \left\| \boldsymbol{A}_{ft} (\dot{\bar{\boldsymbol{x}}}_{ft} - \boldsymbol{J}_{ft} \dot{\bar{\boldsymbol{y}}}) \right\|^2 \right\}
\end{equation}
with $k$ that depends on the order of priority chosen for the new objective.

\section{Objectives Prioritized List}
\label{sec:taskList}
In this section, the objectives inserted into the TPIK procedure, to form the Action $\mathcal{A}$ suitable for the mission, are listed and briefly explained.\\

The first task prioritized list, the one where the two robot act independently to each other, is:
\begin{itemize}
	\item \textbf{Joint Limits avoidance} (\textit{reactive, inequality, safety}). This objective keeps joint away from their mechanical limits. It must be at an high priority because it is a \textit{safety} objective. It is also an \textit{inequality} one to not overconstrain the system when joints are away from their limits.
	
	\item \textbf{Horizontal Attitude} (\textit{reactive, inequality, safety}). This objective is to maintain the vehicle horizontal respect to the water surface. Most of the underwater vehicle are passively stable (and also not controllable) in roll and pitch, so this objective would be useless. In this thesis, where buoyancy is not simulated and the vehicle has full DOF, this objective is necessary. It is also important to not occur in the singularity given by the Euler Sequence used to describe the orientation (section \ref{sec:definitions}), which arises when the pitch is equal to $\pi/2$. The objective is consider a \textit{safety} one because it is more important than the accomplishment of the mission; and it is an \textit{inequality} for the same reason of the previous. 
	
	\item \textbf{Force-Torque} (\textit{reactive, inequality, pre-requisite}). This objective is to reduce the forces and torques acting on the peg during the insertion. This objective is detailed in section \ref{sec:forceTask}.
	
	\item \textbf{Tool position control} (\textit{reactive, equality, mission}). This objective is the one that defines the real mission. It is used to bring the tool towards the defined goal (i.e. inside the hole), so it always must be active.
	
	\item \textbf{Preferred Arm Shape} (\textit{reactive, inequality, optimization}). This is a low priority objective to maintain the arm in a predefined shape. This shape permits the arm to have good dexterity (if the shape is chosen wisely) but it is also useful to transport the peg in a natural way. Being only an \textit{optimization} objective, it is put at low priority.
		
\end{itemize}

\noindent The categories (written in italic) are explained in section \ref{sec:controlObjectives}. Please note that in the code there is also an additional \textit{last task} which is used to cancel out any practical discontinuities during task activations [\cite{IntroMaris1}].\\

After the first TPIK procedure is run for the presented list, it is called other two more times. One is for the cooperation between the two robots (section \ref{sec:coopScheme}), and the other for the vehicle-arm coordination (section \ref{sec:armVehScheme}). Respectively, two \textit{non-reactive} objectives are put at the top of the hierarchy listed above, as explained in the cited sections.\\

Please note that some important objectives related to safe transportation (e.g. obstacle avoidance, minimum altitude from seafloor, minimum distance between robots), grasping (e.g. camera centring object) are not considered because they are not necessary in the particular experiment chosen, and also because they are explored in other works [\cite{IntroMaris2}; \cite{tesiWander}; \cite{IntroRecent}].\\

\section{Change Goal Frame Routine}
\label{sec:changeGoal}
In general, the frame where the tool is driven to by the control architecture (i.e. the \textit{goal} frame) is known with some errors. This is the case when, for example, we have some computer vision algorithm to estimate the hole pose. 
This error between ideal goal and estimated one, both in the linear and in the angular components, can cause the peg to collide a lot with the hole. If the peg clashes against the hole structure face (i.e. outside the proper hole), it is pushed back and a stuck may happen. The result is that the peg will go on bouncing back and forth forever. In the literature, various methods (cited in \ref{sec:artPeg}) have been explored to deal with the problem of \textit{finding the hole}, moving the peg on the surface. In this work, this is not explored.\\

This thesis focuses only the final part of the \textit{peg-in-hole}, i.e. when the peg is inside the hole, but bad alignment causes a lot of collisions with internal hole's  walls. In this case, usually, the peg does not stuck in a intermediate position, because the forces and torques acting on the peg \textit{naturally} drive it inside the cavity. But, in such a way, the peg continuously scrapes along the hole's walls, possibly damaging itself, the hole and also the robot which can suffer some stress. In practice there is a chattering problem: the peg continuously \textit{bounces} 
because the control wants to drag it towards the erroneous pose, while the hole's walls cause forces in a different direction.\\

The method explained in this section try to solve this problem, modifying the goal accordingly to the forces acting on the peg.\\
Let us consider the Cartesian coordinate of the origin $^{w}\boldsymbol{g} \in \mathbb{R}^{3}$ of the goal frame, projected on the world frame. According to the force detected, we modify this origin, providing $^{w}\boldsymbol{g'} \in \mathbb{R}^{3}$ as:
\begin{equation}
	\label{eq:changeGoal}
	\begin{gathered}
	^{w}\boldsymbol{g'} = \,^{w}\boldsymbol{g} + \, ^{w}\boldsymbol{\tilde{f}} \\
	 ^{w}\boldsymbol{\tilde{f}} =  \,^{w}\boldsymbol{R}_t \,^{t}\boldsymbol{\tilde{f}} \\
	 ^{t}\boldsymbol{\tilde{f}} = k \, [ 0, \, f_y, \, f_z ], \qquad 0 < k < 1
	 \end{gathered}
\end{equation}
where $[ 0, \, f_y, \, f_z ]$ is the vector which represent the resultant of the forces acting on the peg, projected on the \textit{tool} frame, but with the component along x put to zero; $^{w}\boldsymbol{R}_t$ is the rotation matrix from world to tool; k a positive gain smaller than 1.\\
The component on $x$ axis of the force is neglected. This is done because the x axis of the tool frame $ \langle t \rangle $ is the one along the length of the peg. So, we do not want that this component modifies the goal because it would change the wanted depth of insertion.\\

To proper utilize the data given by the sensor, that is the force vector, we must use a gain $k$. This gain is less than $1$ for a similar reason as the one explained for the Force-Torque objective (section \ref{sec:forceTask}). For example, when a little force of 1N is detected, obviously we don't want to move the goal frame of 1 m or more. Instead, we want very little modifications, done every time a force (not null) is detected by the sensor. So the formula \eqref{eq:changeGoal} shifts the goal at the same frequency of the provided sensor data. Obviously we can also update the goal less often, maybe with a bigger gain.\\ 

To understand better the method, we can take as an example the situation where the estimated goal is a bit on the left respect to the centre of the hole, but not so much to make the peg miss the hole. In such a case, the control architecture drives the peg on the left of the centre, causing a lot of collisions with the left side of the inner hole. So, the peg suffers forces with an important component in the right direction. Thus, this method shifts the goal to the right.\\
From this example should be noticed that this approach could cause the same problem with the opposite side of the cavity, if the goal is shifted too much on the right. For this reason, setting a suitable $k$ is important.

%%%%%%%%%%%%%%%%%%%%%%%%%%%%%%%%%%%%%%%%%%%%%%%%%%%%%%%%%%%%%%%%%%%%%%%%%%%%%%%%
%2345678901234567890123456789012345678901234567890123456789012345678901234567890
%        1         2         3         4         5         6         7         8
% THESIS CHAPTER

\chapter[Control Architecture: Simulation Results]{Control Architecture: \\ Simulation Results}
\label{chap:results}

\begin{figure}[H]
	\centering
	\includegraphics[width=14.5cm]{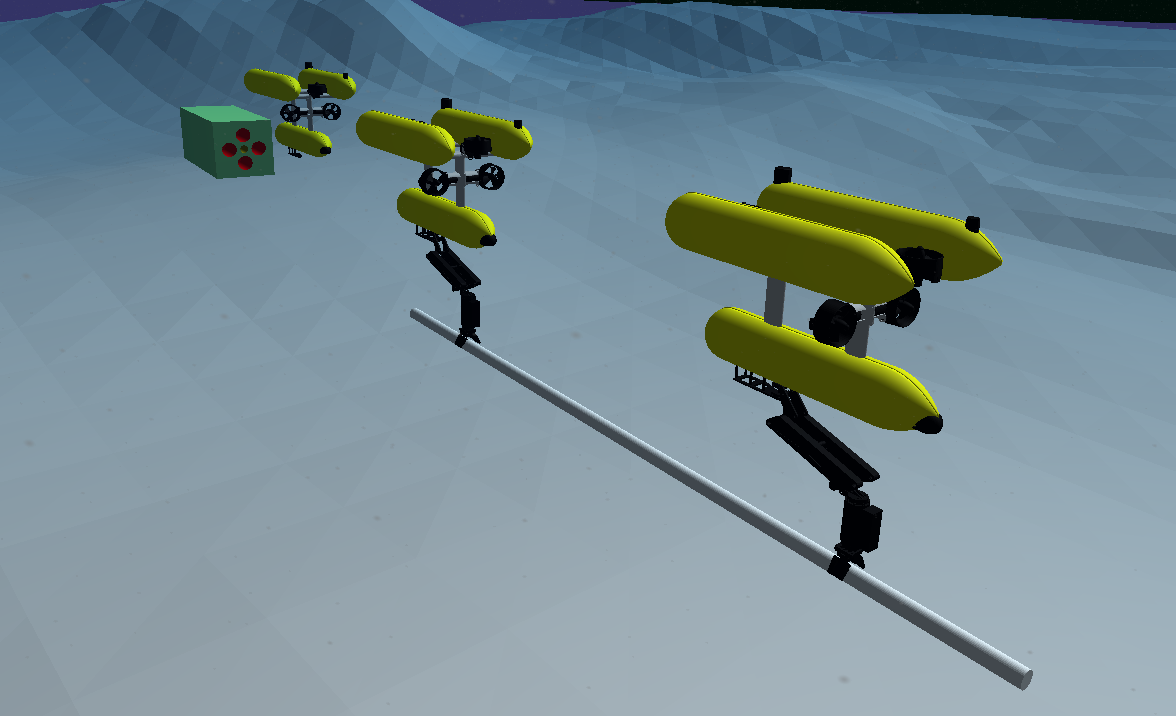}
	\caption[The Scenario with the two robot carrying the peg and the Vision robot watching the hole]{The Scenario of the experiment. The two twin robots are carrying the peg, while the third robot is watching the hole to estimate its pose.}
	\label{fig:method_uwsim}
\end{figure}

In this Chapter, experimental set-up is described, and results are given and discussed. The code for the whole architecture is available at the following link: \url{https://github.com/torydebra/AUV-Coop-Assembly}; some details about it are discussed in section \ref{sec:controlLoop} and in appendix \ref{chap:AppendixCode}.
A video of the final experiment is visible at the following link: \url{https://streamable.com/kvoxq} (online; accessed 10-08-2019).\\

The scenario is made up of two \href{https://cirs.udg.edu/auvs-technology/auvs/girona-500-auv/}{Girona 500} I-AUV's, each one equipped with a CSIP Robot arm5E (a 4 DOF arm with a parallel yaw gripper). The final goal is to successfully coordinate the robots in such a way that the peg, hold by both manipulators, is inserted correctly in the hole, fixed in the environment. In the literature, this problem is known as \textit{peg-in-hole}.\\
One robot is equipped with a force-torque sensor that permits to understand forces applied on the peg, caused by collisions during the insertion phase. This information is provided to both robots. A third robot is equipped with two cameras to estimates the hole's pose. The figure \ref{fig:method_uwsim} shows what has just been described.\\
The chosen strategy divides the problem in two phases: Hole Detection and Insertion. In the first, preliminary steps are done to detect the hole. The third robot, not used for manipulation, is in charge of exploiting computer vision algorithms to estimate the pose of the hole. Details about this are given in section \ref{chap:vision}.
The second phase explores the problems inherent to transportation of the tool, the interaction between the peg and the hole, and the communication between the carrying agents. This is described in this Chapter.

\section{Choosing the Simulator}
\label{sec:simulators}
Some effort has been spent to choice a suitable simulator for the case. At the end, \href{http://www.irs.uji.es/uwsim/}{UWSim} [\cite{uwsim}] was chosen. It is a simulator largely used for this kind of scenarios, where underwater robots must accomplish some particular tasks. It provides a different variety of useful sensors (e.g. the used force-torque sensor and the cameras), and also personalized ones can be added. It uses ROS as the simulator interface, which makes it really easy to use. Through ROS messages, we can send commands to the robots and we can receive information from the going-on test. Contact physics is implemented using \href{https://github.com/mccdo/osgbullet}{OSGBullet} to integrate the physics engine \href{https://pybullet.org/wordpress/}{Bullet} with the 3D graphics toolkit \href{http://www.openscenegraph.org/}{OSG}. For what concerns the collisions, these are calculated taking into account the compenetration between 3D models. The more the models are compenetrated, the more the forces and torques have big magnitudes. To know further details about how the simulator is built, especially for the contact physics part, please refer to the documentation of the cited software.\\

The cons of UWSim is that the simulation is fully kinematic, so no dynamic interactions are present (expect for contact physics). This means, for example, that velocities sent to the robot are immediately accomplished, that buoyancy is not present, and that it is not simulated the physic related to the object grasping. For the scope of this thesis, this lack is not important because dynamic is not considered. Furthermore, how the collisions between the tool and the hole affect the whole manipulator chain, can be simulated thanks to the information provided by the force-torque sensor, as explained in section \ref{sec:forceConsideration}.\\

To fill the UWSIM lack of dynamics, a good alternative can be \href{https://github.com/freefloating-gazebo/freefloating_gazebo}{FreeFloatingGazebo} [\cite{freeFloatingGazebo}]. In truth, this simulator is a plug-in for Gazebo and UWSim; it integrates them in order to achieve both dynamics (thanks to Gazebo) and visually realistic I-AUV simulation (thanks to UWSim). It is easy to use as UWSim, being ROS always the adopted interface, but also because the same scene (described by an \textit{.xml} file) can be used. With this plug-in, we can simulate features as buoyancy and coupling dynamics between arm and vehicle (i.e. how arm movements affect the base).
FreeFloatingGazebo has been taken into consideration for dynamic tests, but at the end it is not used due to the lack of time. However, further works toward dynamic simulations can begin from here.\\

Another simulator is \href{http://gazebosim.org/}{Gazebo}, widely known in all robotics fields. It is the de-facto standard simulator for ROS. Due to its generic purpose, it is not a ready-to-use simulator for an underwater environment, so it can be only a starting point to develop a software specific for this particular scenario (as it is done by FreeFloatingGazebo).\\
Also other similar simulators, \href{http://www.coppeliarobotics.com/index.html}{V-REP} [\cite{vrep}] and \href{https://cyberbotics.com/}{Webots} [\cite{webots}] have been taken into consideration, but then they have been discarded for the same \enquote{not ready-to-use} reason like Gazebo.\\

Another interesting simulator is \href{https://github.com/disaster-robotics-proalertas/usv_sim_lsa}{USV} [\cite{usvsim}], a really recent and in-development project. It takes the best from UWSim, Gazebo and FreeFloatingGazebo to implement realistic simulations. However, it is focused more on surface vessels dynamics.\\

\begin{figure}[H]
	\centering
	\includegraphics[width=14cm]{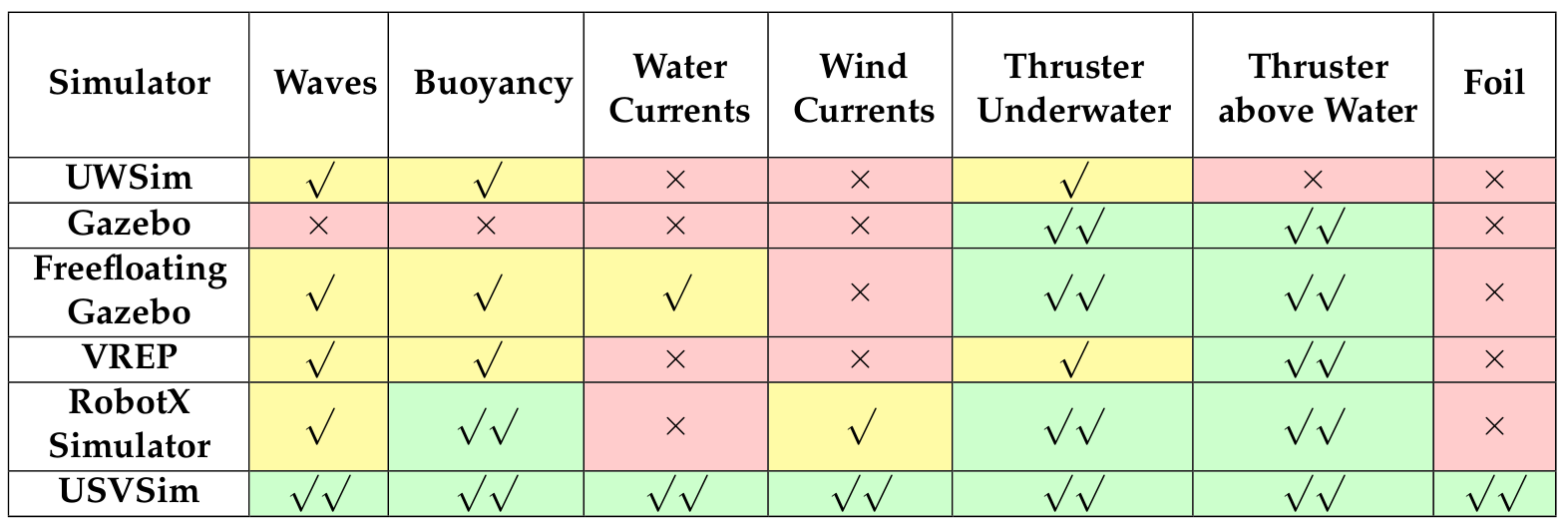}
	\caption[Table of simulators comparison]{Schematic recap of the simulation comparison taken from \cite{usvsim}. $\times$ stands for no implemented feature; ~ $\surd\,$  for a feature that is a discrete representation of the real one; ~ $\surd\surd\,$ for a good feature reproduction of the real one. More details on how each feature is evaluated are available in the original paper.}
	\label{fig:simComparison}
\end{figure}

More details on these and other simulators are available in \cite{simComparisonCook} and \cite{usvsim}. From \cite{usvsim}, a schematic comparison is taken and shown in figure \ref{fig:simComparison}.

\section{Simulating the Firm Grasp Constraint}
\label{sec:firmGrasp}
Being UWSim (introduced in section \ref{sec:simulators}) only a kinematic simulator, some additions have to be made.\\
Without dynamics, simulating correctly the peg grasping is impossible. The simulator permits to fake it with an \textit{object picker} sensor: when an object is sufficiently near to the point where this sensor is, it becomes \enquote{grasped} and, from then on, it will rigidly move with the whole robot. The problem here is that we have two robots that must take the tool, so the object can't rigidly move with both, but only with the first which catches it. \\
Furthermore, external forces applied to an object (grasped or not) can't be detected with the force-torque sensor, because, in the way it is implemented, it only detects forces acting on a vehicle part.\\

To solve this issue, a peg is modelled as an additional fixed joint attached to the end-effector of each robot. In this way, each peg is rigidly attached to its own robot. Now, the problem is how to maintain the two pegs perfectly overlapped during the whole mission, because, obviously, in real situation the tool is unique. This is also needed because the control architecture assumes a \textit{firm grasp} of the tool, without any slipping. This means that the end-effector does not move respect to the peg, and, consequently, the end-effectors of the robots do not move respect themselves. It is also important to report that the force-torque sensor does not detect collisions between the two tools, so we don't have problem for this point o view.\\

In the simulation, collisions between the \enquote{pegs} and the hole cause the tools to drive apart. This happens because collisions are propagated to the robots with a formula which use Jacobian (detailed in section \ref{sec:forceConsideration}). Jacobian derives from approximation of non-linear relationship, so results are not perfect. Thus, during the transportation, but especially during the collision propagation in the insertion phase, the two pegs distance themselves a bit. This causes that the control point for one robot is in a different position of what it expects, increasing the errors.\\

In real scenario, a firm grasp acts like a \enquote{glue}: if the end-effector tends to go away from the grasping point, friction acts to maintain it to the contact point. This is true for very small errors; if the cooperation's performance is not good, the common tool falls down or something breaks.\\

In the simulation, to fake the firm grasp, an additional routine is implemented. It simply calculates the distance between the two pegs, and it generates robot velocities to nullify this gap. It is important to notice that this is an aid that we would have also in a real scenario, as explained before. The only difference is that, in real scenario, if the errors are too big the end-effector begins to slip, and it will never return to its original grasping point. In this case, it returns always to the initial point. \\

The velocities generated by this routine are not so big to hide bad cooperation; so the tests are suitable to evaluate the proposed architecture, and to simulate real behaviours.

\section{Simulating the Collision Propagation}
\label{sec:forceConsideration}
When a robot interacts with the environment, each contact generates forces on it. Missions related to assembling any objects can't be studied in a properly manner without some considerations about these forces. In a \textit{peg-in-hole} mission, collisions between the peg and the hole will be transferred through the whole kinematic chain until the floating base, causing disturbances to the whole robotic system. Thus, it is necessary to simulate these behaviours. Being UWSim a kinematic-only simulator, an additional feature is implemented to cope with these kind of collisions.\\ 

\noindent Let us define $\boldsymbol{f} \in \mathbb{R}^3$ and $\boldsymbol{m} \in \mathbb{R}^3$ as:
\begin{equation}
\boldsymbol{f} = \begin{bmatrix}f_x \\ f_y \\ f_z\end{bmatrix} \qquad
\boldsymbol{m} = \begin{bmatrix}m_x \\ m_y \\ m_z\end{bmatrix}
\end{equation}
being $\boldsymbol{f}$ and $\boldsymbol{m}$ the resultant force and the resultant torque (projected on the tool frame $ \langle t \rangle $) of all the forces and torques acting on the tool. \\
These vectors generate a disturbance on the whole system as a velocity $ \dot{\boldsymbol{y}}_{\delta} \in \mathbb{R}^n$. This velocity can be written as [\cite{bookSiciliano}]:
\begin{equation}
	\label{eq:forTor}
	\dot{\boldsymbol{y}}_{\delta} \triangleq 
	\begin{bmatrix} \dot{\boldsymbol{q}}_{\delta} \\ \boldsymbol{v}_{1 \delta} \\ \boldsymbol{v}_{2 \delta} \end{bmatrix}
	= \begin{bmatrix} k_q \\ k_{v1} \\ k_{v2} \end{bmatrix} \, \begin{bmatrix}\;(^{lin}\boldsymbol{J}_t)^T \boldsymbol{f} + \;(^{ang}\boldsymbol{J}_t)^T \boldsymbol{m}\end{bmatrix} 
	\qquad 0 < k_q, k_{v1}, k_{v2} < 1 
\end{equation}
where $\boldsymbol{J}_t$ is the Jacobian which expresses how the Cartesian tool velocity $\dot{\boldsymbol{x}}_t$ is affected by the system velocity vector  $\dot{\boldsymbol{y}}$; $\;lin$, $ang$ superscripts refer to \textit{linear} part (top three rows) and \textit{angular} part (bottom three rows) of $\boldsymbol{J}_t$; $\;\dot{\boldsymbol{q}}_{\delta} \in \mathbb{R}^l$ are the joints velocities caused by the collisions; $\;\boldsymbol{v}_{1\delta} \in \mathbb{R}^3$ and $\boldsymbol{v}_{2\delta} \in \mathbb{R}^3$  are the linear and angular vehicle velocity caused by the collisions; $\;k_q, k_{v1}, k_{v2}$ are positive gains smaller than $1$, and in general different from each other  because we are considering different types of velocities.\\
Similarly to the Force-Torque objective (section \ref{sec:forceTask}) and for the Change Goal routine (section \ref{sec:changeGoal}), gains smaller than $1$ are necessary when dealing with forces and torques to not generate too high velocities.

\section{Experiment's Assumptions}
\label{sec:expAssumption}
It is important to detail the assumptions made during the simulation. In fact, some problems, that must be taken into account in a real environment, are not explored. This is necessary due to the difficulties of the particular mission analysed. \\
So, in this section, the main assumptions are summarized.

\begin{itemize}
	\item Simulation is kinematic-only. This implies, for example, that the commanded velocity to the vehicle and the arm are accomplished \textit{instantaneously} and \textit{perfectly}. Another implication is that the movements of arm and of the vehicle don't influence each other at all. The only exceptions are the Firm Grasp constraint (section \ref{sec:firmGrasp}) and the Collision Propagation (section \ref{sec:forceConsideration}).
	
	\item The initial configuration shows the peg already grasped \textit{correctly} by both robots. Also, the point where the end-effectors have grasped the tool and the peg's dimensions are known. This implies that the relative position between each robot and peg's tip is \textit{perfectly} known.\\
	Such an initial configuration has been chosen because the grasping phase and problems arising during cooperative transportation have been explored in other projects like MARIS and ROBUST (e.g in the work \cite{IntroMaris2}) and in the on-going TWINBOT.

	\item A common reference frame (denoted as $\langle w \rangle$ - \textit{world} in the whole thesis) is used to know the relative poses among objects, carrying robots and the Vision robot. This assumption is mostly needed to make the Vision robot share correctly the estimated hole's pose.\\
	In real situation, the underwater location of something is always an issue and it is never really precise. Good precision can be provided, for example, after some preliminary works in mapping the seafloor. Another method can be the exploitation of helper support vessels, for example, as explored the WiMUST project [\cite{wimust}]. This can provide a common reference point somewhere.\\
	Please note that, for this work, it is not important that the common frame $\langle w \rangle$ is located above the sea surface. The important thing is only to have a common point, that can also be underwater.\\

	\item No real communication problems between the two cooperative robots are taken into account.\\
	The presence of water gives relevant issues in a real situation. A \textit{full-duplex} communication (i.e. sharing data \textit{at the same time}) is impossible. Also, in general, data exchange is much slower respect to the air. Some experiments in simulated environment with different methods of underwater communication are detailed in \cite{IntroMaris2}.\\
	However, these communication issues are considered by the cooperative scheme (as explained in section \ref{sec:coopScheme}); in fact it permits to exchange very few information between the two carrying agents.
	
	\item The two robots firmly grasp the peg. There is no sliding caused by robot movements. This point is detailed in section \ref{sec:firmGrasp}.
	
	\item During the insertion phase, the control architecture tries to resolve alignment errors \textit{only if} the peg is inside the hole. No methods are implemented to deal with the problem of \textit{looking for} the hole on the surface. So, if the peg touches the external hole surface (due to a too big hole's pose estimation error), it will bounce back and forth forever.
	
	\item The force-torque sensor is positioned on the tip of the peg, and it provides the resultant force and the resultant torque of collisions on the whole peg.\\
	This would obviously not possible in real applications. In real scenario, the sensor is usually put on the arm's wrist and it provides forces and torques respect to this point.\\
	In this case, we could have projected the force-torque sensor information on the wrist frame, to make more realistic simulation, but for simplicity this is not implemented.\\
	Furthermore, both robots have access to the sensor data, at the same frequency, and without uncertainties (except errors due to how almost all physics engines compute collisions, that it is done with approximations to improve the performance).
	
\end{itemize}

\noindent Others assumptions, more related to the vision part, are detailed in section \ref{sec:visioAssumption}.

\section{Control Loop}
\label{sec:controlLoop}
\begin{figure}[H]
	\centering
	\includegraphics[width=12cm]{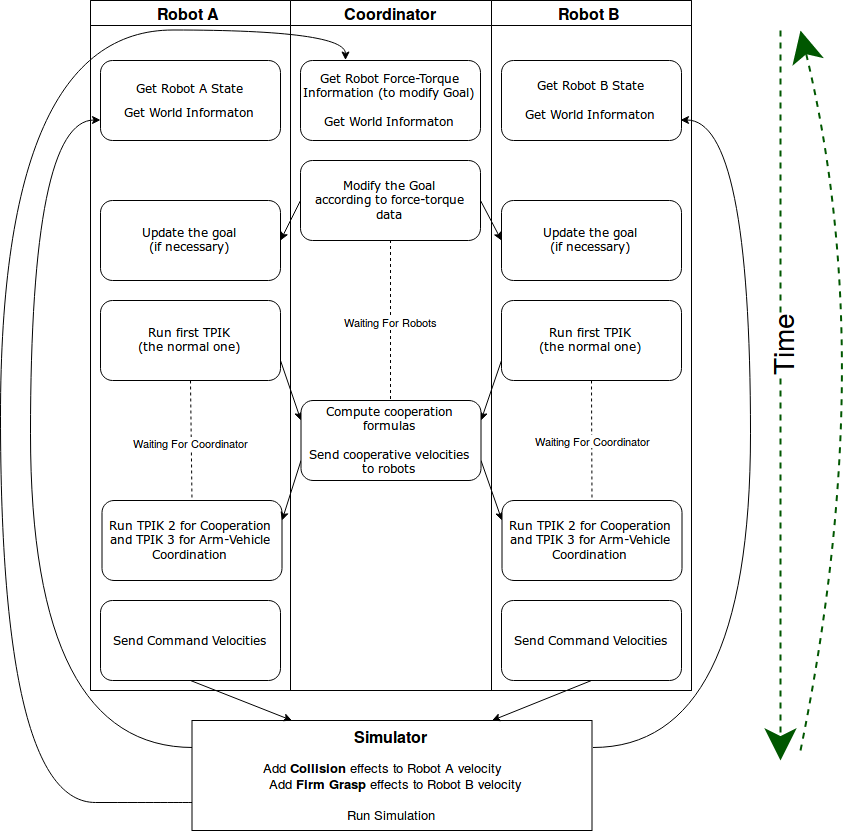}	
	\caption[Flow Scheme of the Control Loop]{A flow scheme showing the main steps of a single control loop for the robots and for the coordinator. Blocks at the same horizontal level are executed at the same time. Arrows indicate sharing of data between blocks of different nodes. Note here that the Vision robot is not considered.}
	\label{fig:flowScheme}
\end{figure}

This section is written with the scope of giving a better idea on how the control architecture works.\\
The Vision robot, which job is to estimate the hole's pose, acts in a preliminary phase. It \textit{tracks} the hole, thanks to stereo-cameras, and it sends the estimated pose to the Coordinator. Due to its nature, no complicated control is implemented for this agent: when the pose is sent, we simply move the robot away from the hole with keyboard (like a ROV) to not interfere with the insertion mission. This part is described in Chapter \ref{chap:vision}.\\
After the coordinator receive the hole's pose, it sends it together with a signal to the two carrying robots to make them begin the mission.\\

The two carrying robots are fully autonomous: as soon they get the hole pose, they proceed \textit{without user intervention}. There are three nodes running at the same time: the two Robots (\textit{A} and \textit{B}) and the Coordinator. The latter is not a real \textit{physical} agent: it is only a software routine. So, it can be physically inside a robot, from now on, the Robot A. In this way, communication issues (due to the underwater scenario) occur only between the two robots, and not among all the three nodes.\\

At the beginning of the mission, the Agent A, the Agent B and the Coordinator synchronize themselves, i.e. each one waits that the other two are ready. After this phase, the normal routine starts. In figure \ref{fig:flowScheme}, the main instructions of the control loop are depicted.

\begin{itemize}
	\item At the beginning of the control loop, each node gets the updated simulation state, e.g. pose of the robots, pose of the tool, information from force-torque sensor, and so on.
	
	\item The Coordinator, which (as said previously and without loss of generality) is a software routine inside the Robot A, modifies the goal's linear position (as explained in section \ref{sec:changeGoal}), if some forces are detected. If the goal is updated, the two agents get this new information.
	
	\item In the third block's row, the Robots run the first TPIK procedure. Then, they send the necessary data to the Coordinator, which computes the cooperative velocities and sends them back to the robots. Finally, the two Agents run another two TPIK procedures, one for the cooperation (section \ref{sec:coopScheme}) and the other for the vehicle-arm coordination (section \ref{sec:armVehScheme}).
	
	\item At the end, the two Agents send the system velocity vectors provided by TPIK to the simulation.
	
	\item Before sending the velocities to the \enquote{real} simulation, some disturbances must be added to the commanded system velocity vectors. For the Robot A, this means adding effects of collisions between the peg and the hole (section \ref{sec:forceConsideration}). Instead, for the Robot B, effects of the firm grasp constraint are added (\mbox{section \ref{sec:firmGrasp}).}
	
	\item After the simulator performs a step, the loop starts again.
	
\end{itemize}

\noindent It can be noticed that the two added physical interactions (collisions and firm grasping) are added only for one robot (A and B, respectively) and not for both.\\
This is done to not add simulation errors that could occur, and that would not happen in real scenario. For example, in real situation there are not two coincident pegs (as in this simulation) and so they can't really distinguish themselves. Putting the firm grasp constraint only on one robot helps to reduce disturbances that in real scenario are not present. Also, it is sufficient to fake a real firm grasping.\\
About the collisions, they affect, \textit{directly}, only the first agent. In truth, they also affect the other one, \textit{indirectly}, because the latter is \textit{dragged} by the firm grasp constraint. So, practically, collisions affect the behaviours of both agents.\\
It is important to notice that these physical interactions do not hide control problems: if the control is setted badly, the whole mission fails (e.g. the two pegs diverge and/or compenetrate visibly with the hole).\\
%%%%

Another thing to notice is that, when the Force-Torque objective (Section \ref{sec:forceTask}) is used, both Robots need the sensor data. Being the force-torque sensor only on Robot A, this means that additional communications between Robot A and Robot B are needed. As known, underwater data transmission is complicated and slow, and its amount should be kept as small as possible (from this problem it derives the used coordination policy).\\ 

An alternative to sharing force-torque data can be using another sensor on the Robot B. The problem following this direction could be that different sensors give not exactly same data. So, the two robots run each TPIK with different information. This alternative is not explored here.\\

Another solution can be simply to avoid using this objective: this would decrease the performance of the mission (as we will see later) but results are good anyway.\\

It must be noticed that also to update the goal additional information has to be exchanged between the two robots. 

\section{Results}
\label{sec:resultNoVisio}
\begin{figure}[H]
	\centering
	\includegraphics[width=11.5cm]{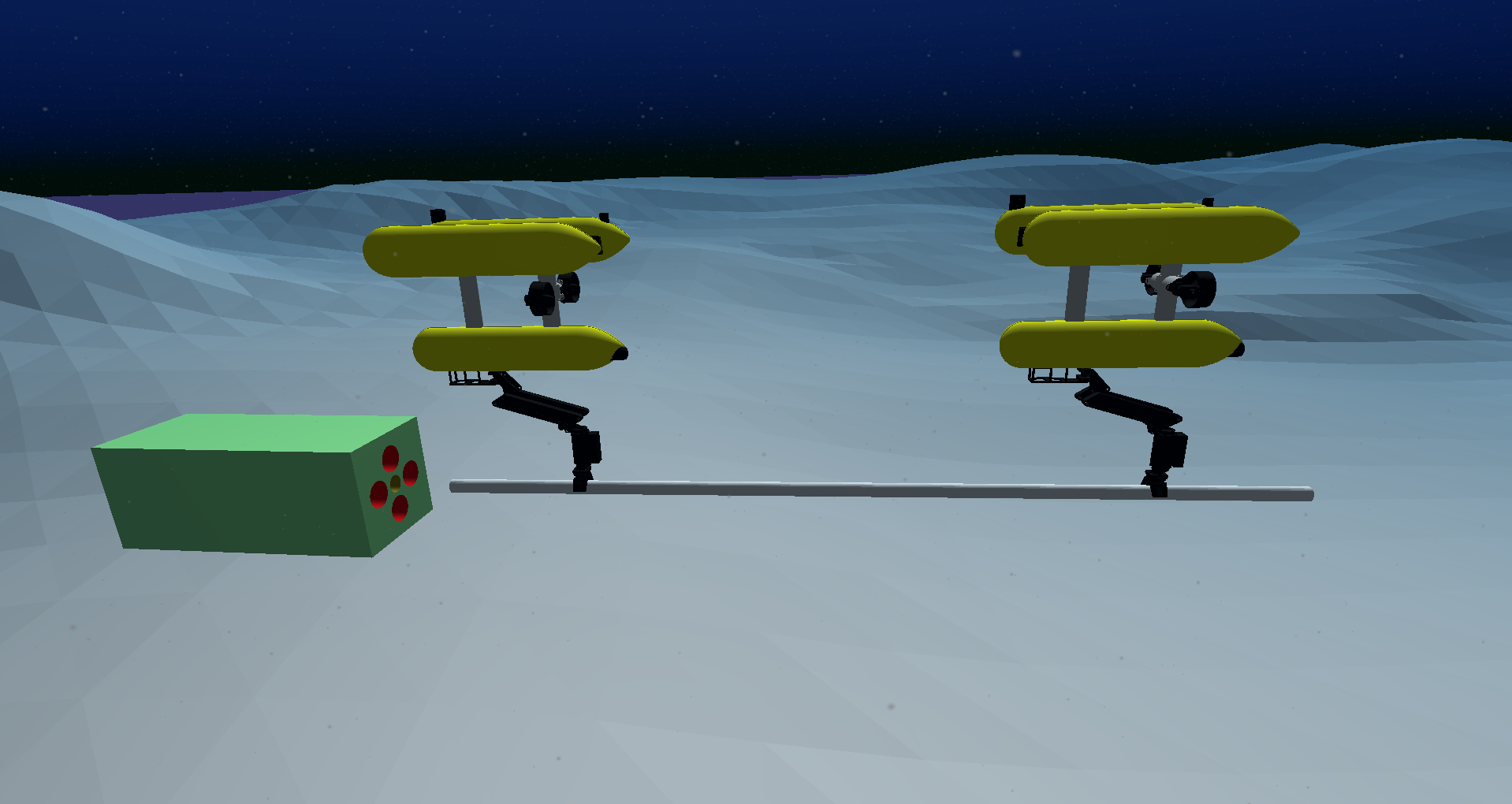}
	\vspace{5px}
	\includegraphics[width=11.5cm]{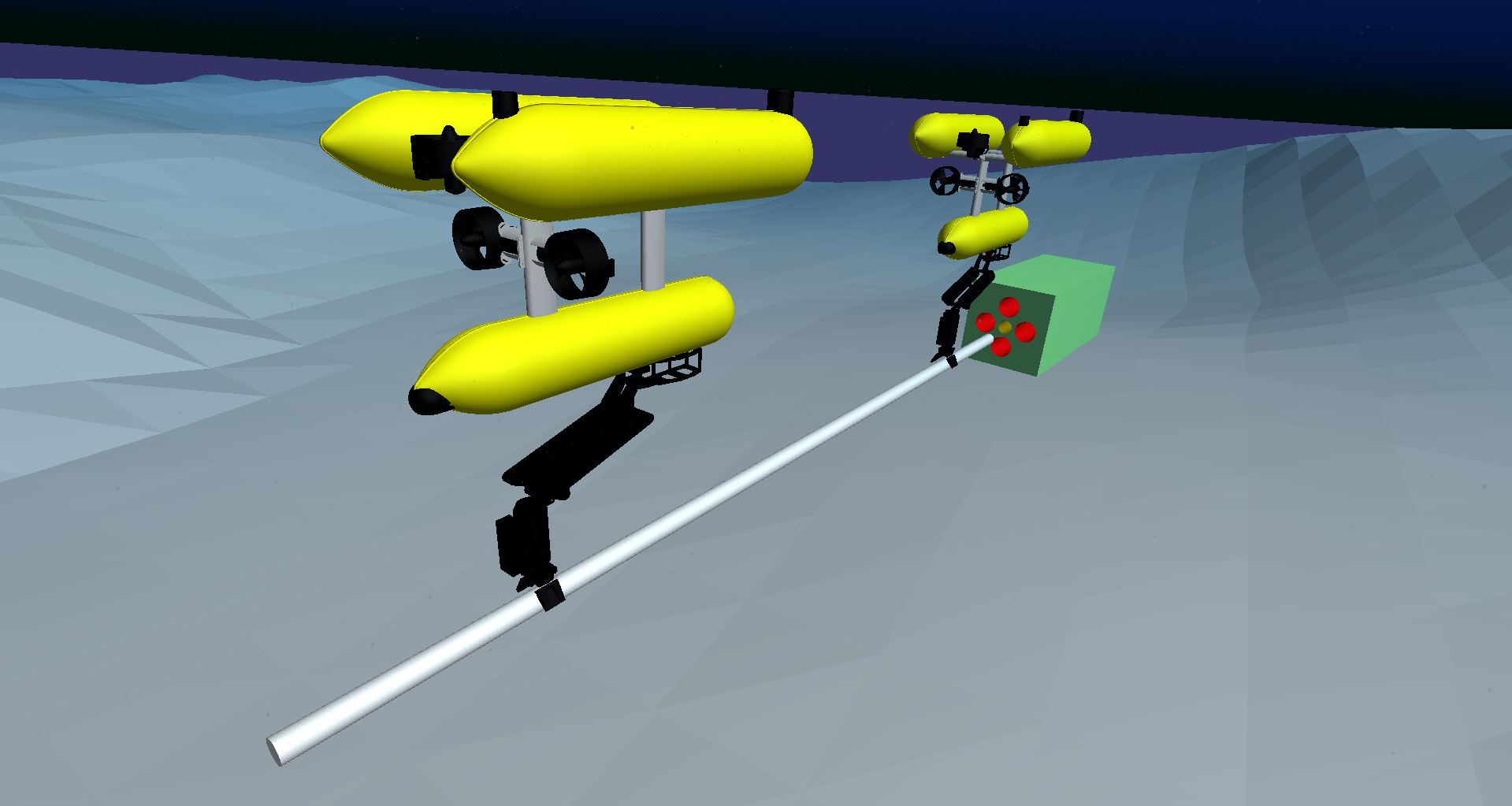}
	
	\caption[Scenario for tests without the vision part]{Two different points of view of the scenario for the results presented in this section. The pose estimation with vision is neglected here.}
	\label{fig:onlyTwin_uwsim}
\end{figure}
In this section, four different experiments are presented. An additional last experiment, comprehensive of the Vision part, is presented in section \ref{sec:finalTest}. Discussions and results analysis are given in section \ref{sec:resDisc}.\\ 

As said, in this section trials do not take into consideration the vision part. This is done to have a hole's pose error arbitrarily settable, that permits to discuss independently the performance of the control methods used.\\

\begin{figure}[H]
	\centering
	\includegraphics[width=12.5cm]{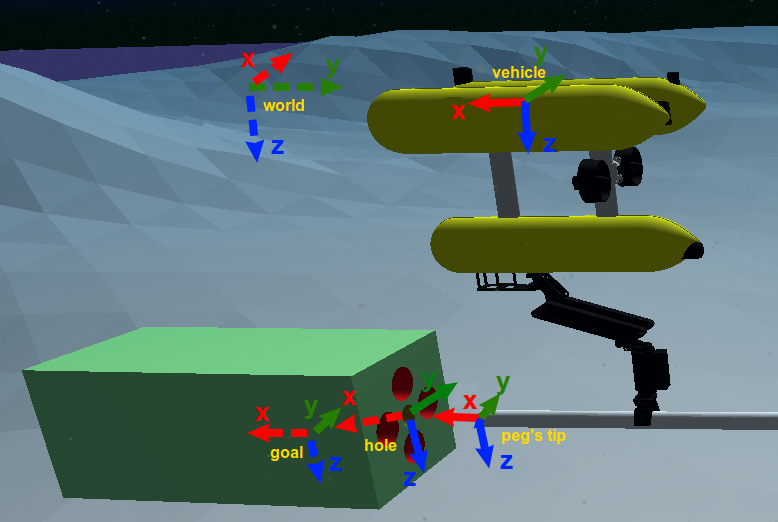}	
	\caption[Main frames related to the insertion phase]{Detail of a screenshot where the main frames are drawn. The world frame is present only to clarify its orientation; its origin is not in that point. For the hole frame, the $x$-axis goes inside the cavity. The goal frame $\langle g \rangle$ has the same orientation of the hole frame, and it is shifted of 0.2 meters in the direction of hole's $x$-axis.}
	\label{fig:scenario_frames}
\end{figure}

The figure \ref{fig:onlyTwin_uwsim} shows the robots initial position. Below, some important details about the simulations are listed:
\begin{itemize}
	\item The \textbf{Peg} is a six meters long cylinder, with a diameter of 0.10 meters.
	\item The \textbf{Hole} is a cylindrical cavity with a diameter of 0.14 meters. It is at the centre of a cuboid structure. In the figure \ref{fig:onlyTwin_uwsim}, the hole is the yellow cavity between the four red circles (that are present only to aid the vision algorithms).
	\item The initial position of the agents is near the hole: the peg's tip is almost aligned perfectly to the hole, and it is at almost 0.44 meters from it.\\
	To be precise, the vector from peg's tip to the hole has components :\\ $[0.441, -0.008, -0.018]$. The peg's tip frame has the $x$-component along the length of the peg; the \mbox{$y$-component} lies on the tip's surface and it points to the left; the $z$-component points downward (as in figure \ref{fig:scenario_frames}).\\
	The orientation from the peg's tip frame to the hole frame is described by these Euler angles: $[0, 0, 1.942]$ (\textit{roll, pitch, yaw}, in degrees).
	\item The mission's aim is to drive the peg inside the hole with a depth of 0.2 meters.
	\item All the vectors displayed in the next plots are projected in the world frame (which orientation is visible in fig. \ref{fig:scenario_frames}).
\end{itemize}

In the following results, only a few plots (fig. \ref{fig:expWithVisioVel} and fig. \ref{fig:expWithVisioVelTool}) are related to the cooperation scheme of section \ref{sec:coopScheme}, and no long discussions are made for them. This choice has been made because no robot has difficulty in tracking the ideal common tool velocity $\dot{\boldsymbol{x}}_t\,$. So, no very interesting plot occurs when \textit{non-cooperative} and \textit{cooperative} velocities are compared. Experimental results about this particular scheme used can be found in \cite{IntroMaris2} and \cite{tesiWander}.\\

The focus of the results is on the implemented methods for helping the insertion phase: the Force-Torque objective (section \ref{sec:forceTask}) and the Change Goal routine (section \ref{sec:changeGoal}). Further outcomes are presented about the added simulation procedures: the Firm Grasp constraint (section \ref{sec:firmGrasp}) and the Collision propagation (section \ref{sec:forceConsideration}). These last two extensions are important to improve the simulation in an otherwise pure-kinematic scenario.

\subsection{Perfectly known Hole's Pose}
\label{sec:testPerfectHolePose}
\begin{figure}[H]
	\centering
	\textbf{Perfectly known Hole's pose}\\
	\vspace{8px}
	\centerline{
		\includegraphics[width=9cm]{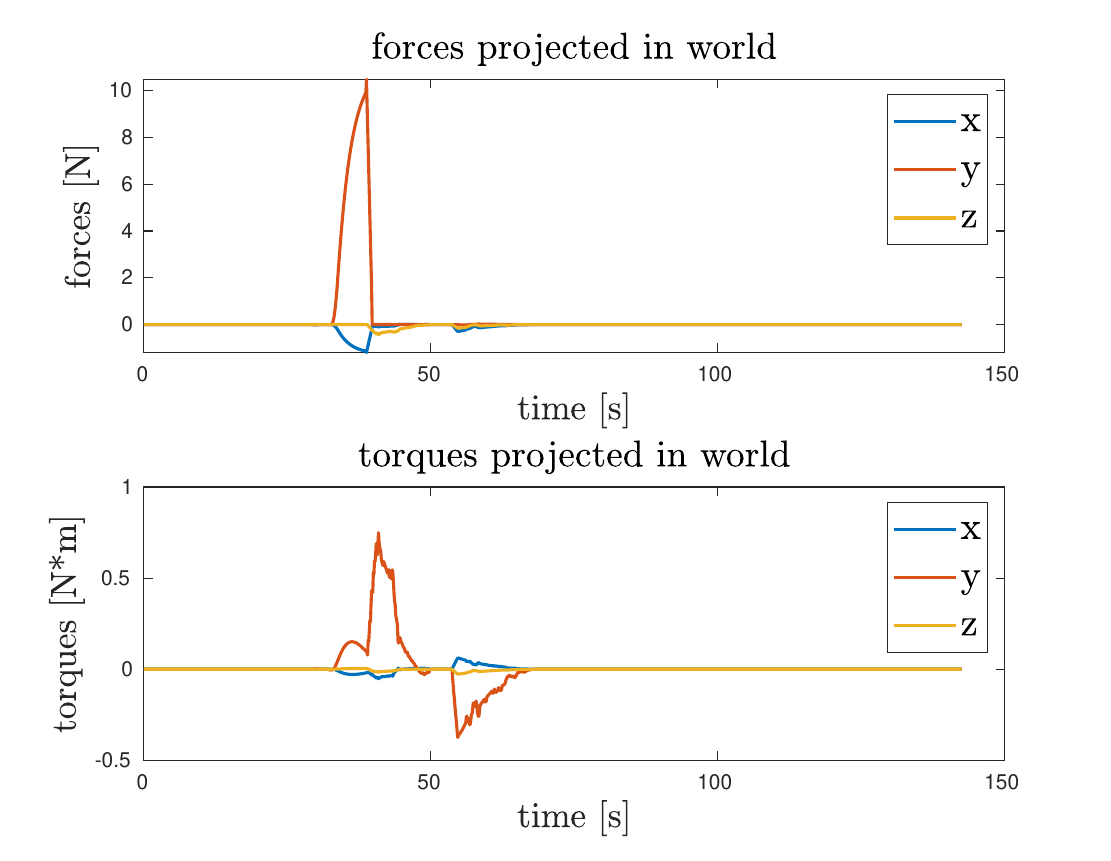}
		\includegraphics[width=9cm]{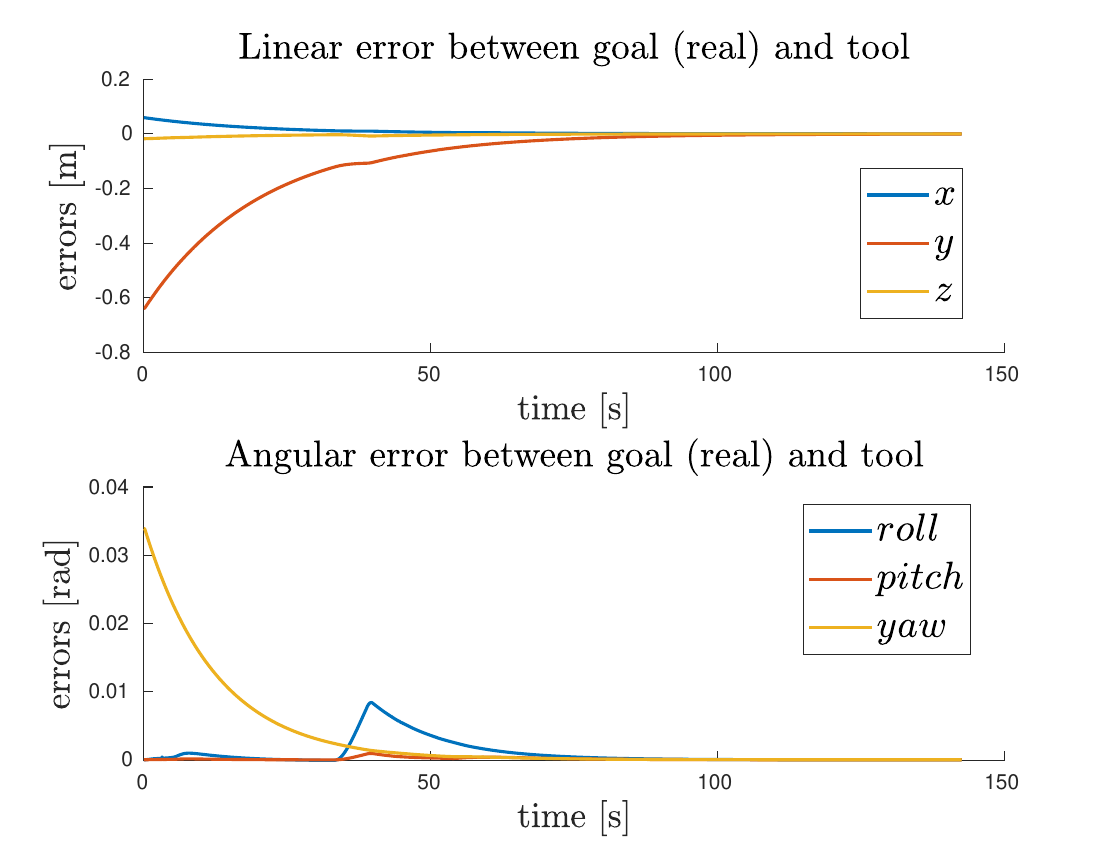}
	}
	\vspace{6px}
	\centerline{
		\includegraphics[width=9cm]{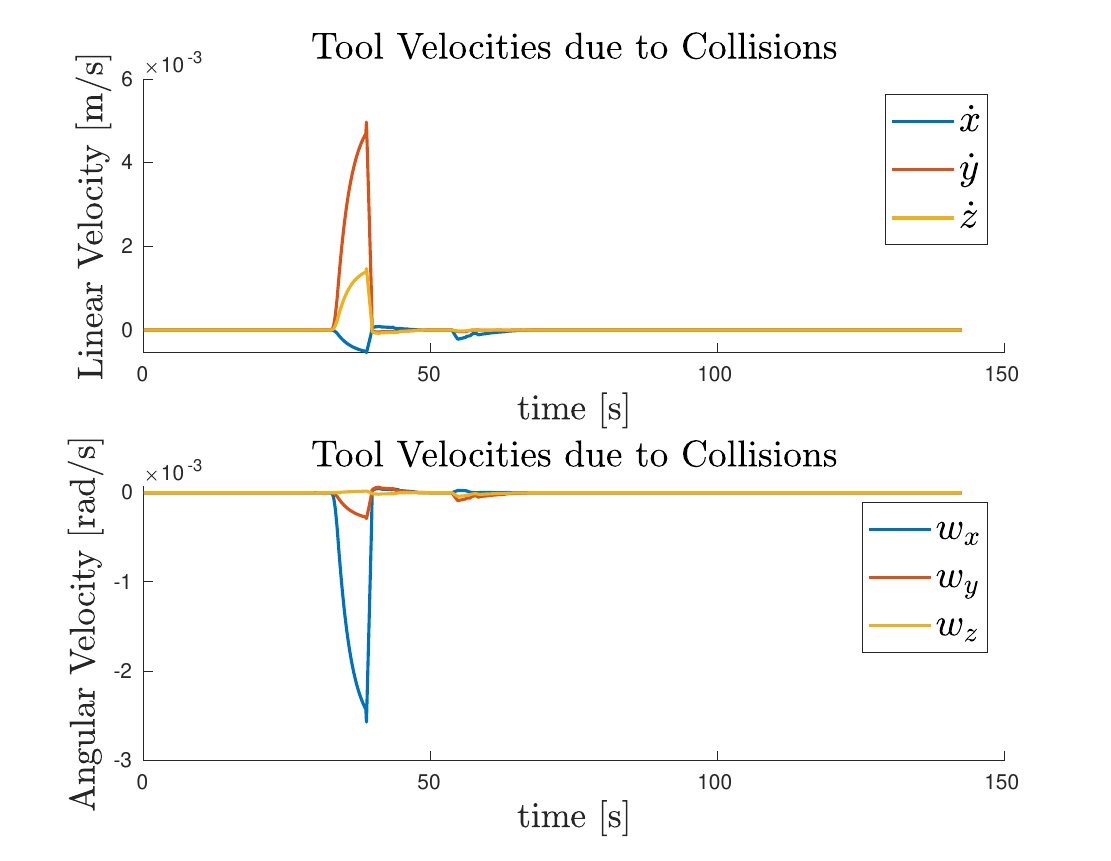}
		\includegraphics[width=9cm]{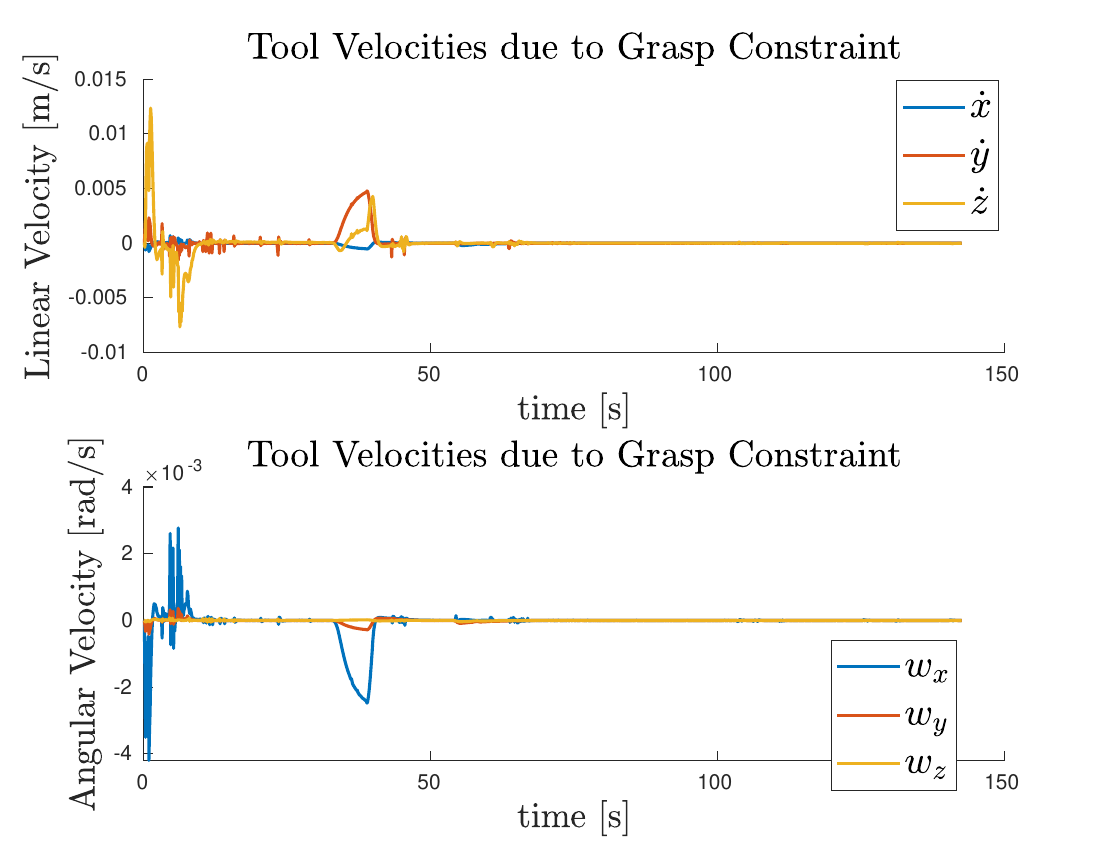}
	}
	\caption[Plots with perfectly known Hole's pose]{The results with the hole's pose known without errors. The upper left plot shows the forces and torques acting on the whole peg: they are the components of the total resultants projected on the world. The high peak in $y$ is due to the first contact between the peg and the hole surface. The upper right plot shows the convergence of the error between goal frame and tip frame. The lower left plot displays the tool velocities generated by the system motions caused by the collisions propagation. The lower right plot shows the tool velocities generated by the Firm Grasp routine.}
	\label{fig:noErrorPlots}
\end{figure}

In the first experiment, the hole's pose (and so the goal frame $\langle g \rangle$) are known without uncertainties. The plots of figure \ref{fig:noErrorPlots} show: how the forces and the torques act on the peg; the converging positional error from the goal to the peg's tip; the tool velocities generated by the collisions; the tool velocities caused by the firm grasp constraint.

\subsection{Error on the Hole's Pose}
\label{subsec:resultsControlError}
In general, a perfect pose estimation is never achievable, so the control should take into account that errors can be present. In this experiment, an error of 0.015 meter is added along the $x$-component of the goal (considering the goal projected in the world frame). So, the peg is driven a bit on the right respect to the centre of the hole, causing a lot of collisions with the right side of the cavity.\\

Three different experiments with the given error have been conducted. In the first, the control does not use any method to exploit the force-torque data (as in the previous experiment of section \ref{sec:testPerfectHolePose}). In the second, the Change Goal routine (described in section \ref{sec:changeGoal}) is used to try to correct the pose error. In the third, both the Change Goal routine and the Force-Torque objective (described in section \ref{sec:forceTask}) are included. The addition of the new objective in the TPIK list is done to try to reduce the amount of force and torques acting on the peg.

\subsubsection{Change Goal Routine Results}
As explained in section \ref{sec:changeGoal}, it has been implemented a routine to move the goal's origin according to the forces and torques detected by the sensor. A comparison of the results with and without this method is visible in figure \ref{fig:Error_nothingandgoal_plot6}. It can be seen that the goal is changing, and at the end of the experiment the added error is compensated.

\subsubsection{Force-Torque Objective Results}
\begin{figure}[H]
	\centering
	\textbf{Error of 0.015 meter on x-axis of the goal\\}
	\vspace{30px}
	\textbf{Norm of the forces and torques acting on the peg\\}
	\vspace{10px}
	\centerline{
	\hspace{5px}
	\textbf{Without Change Goal routine} 
	\hspace{35px}
	\textbf{With Change Goal routine}
	\hspace{25px}
	\textbf{With Change Goal and Force Task} 
	}
	\centerline{ 
		\includegraphics[width=6.5cm]{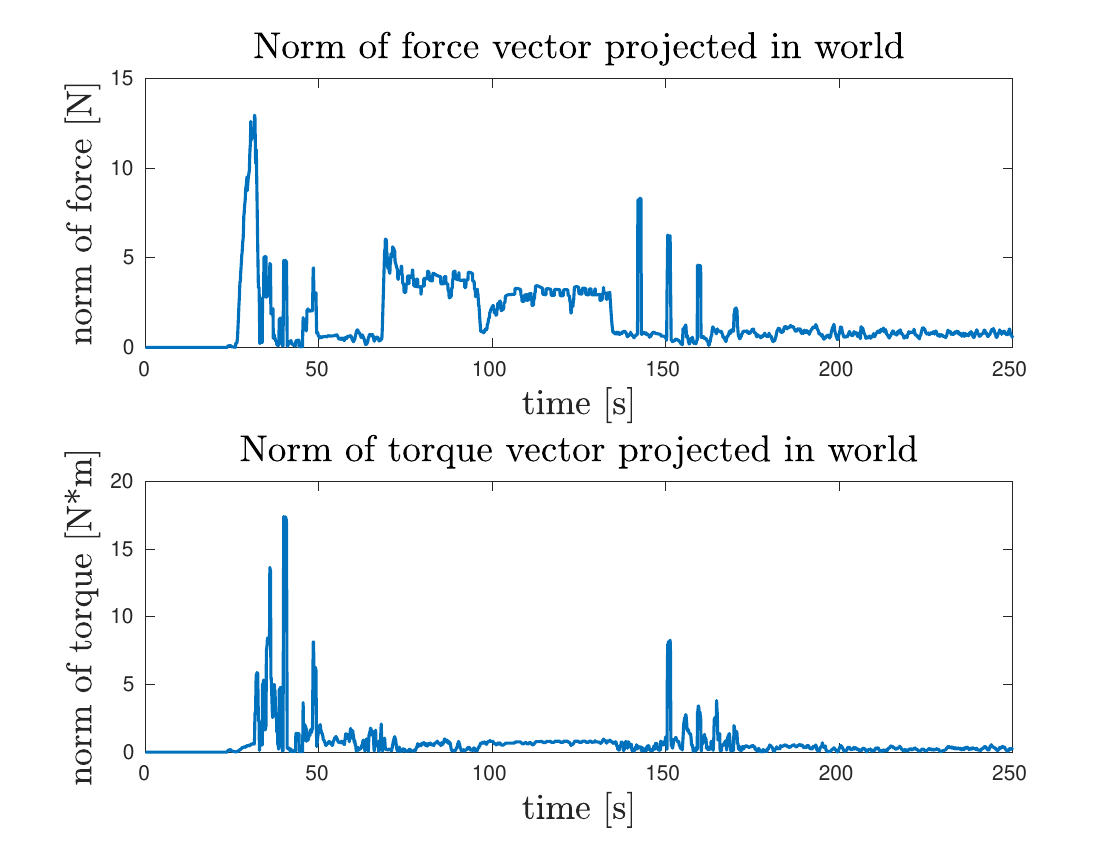}
		\includegraphics[width=6.5cm]{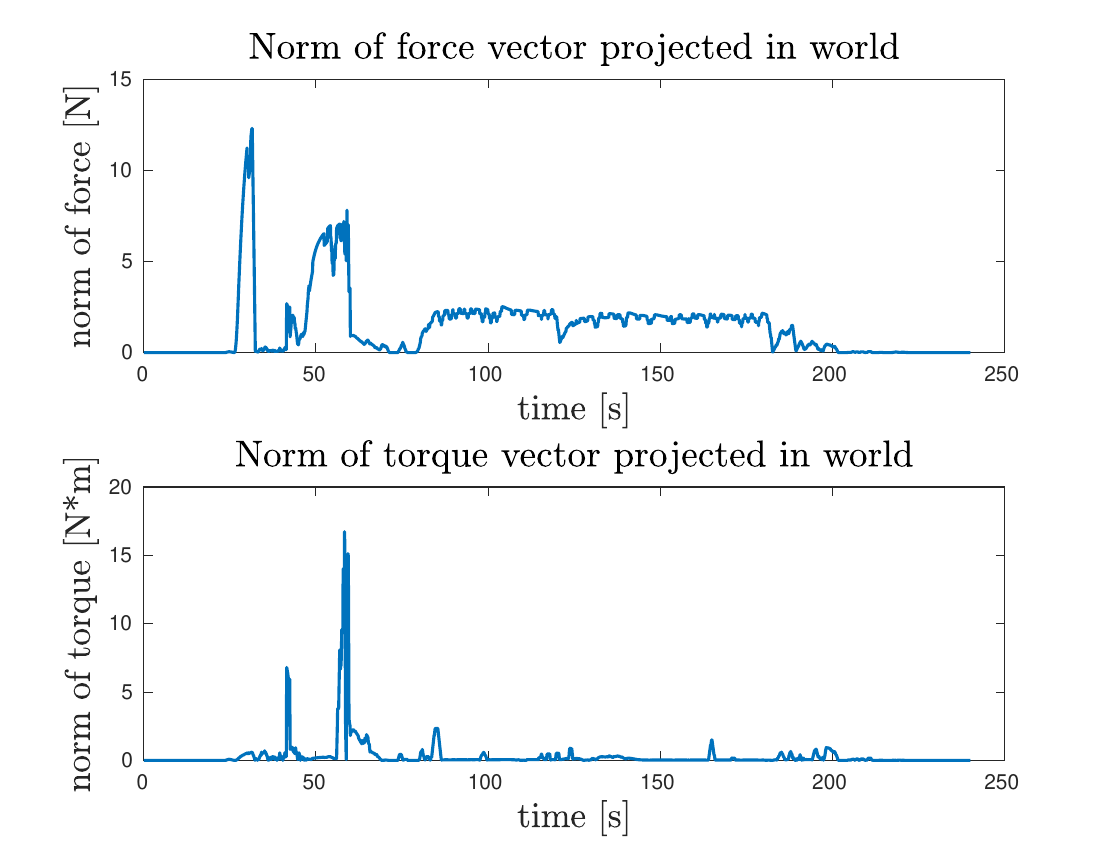}
	 	\includegraphics[width=6.5cm]{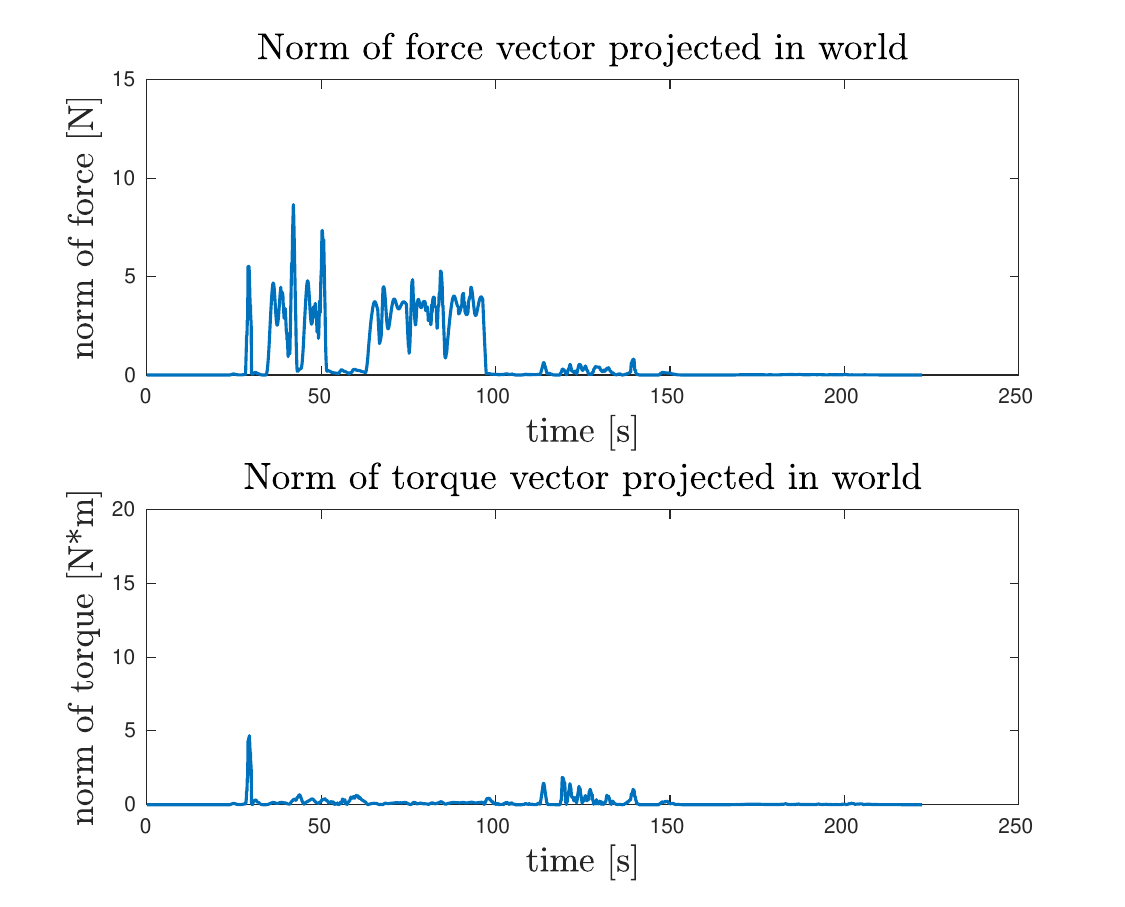}
	}
	\vspace{30px}
	\textbf{Norm of the error between ideal goal (without the added error) and tool's tip\\}
	\vspace{10px}
	\centerline{
		\hspace{5px}
		\textbf{Without Change Goal routine} 
		\hspace{35px}
		\textbf{With Change Goal routine}
		\hspace{25px}
		\textbf{With Change Goal and Force Task} 
	}
	\centerline{
		\includegraphics[width=6.5cm]{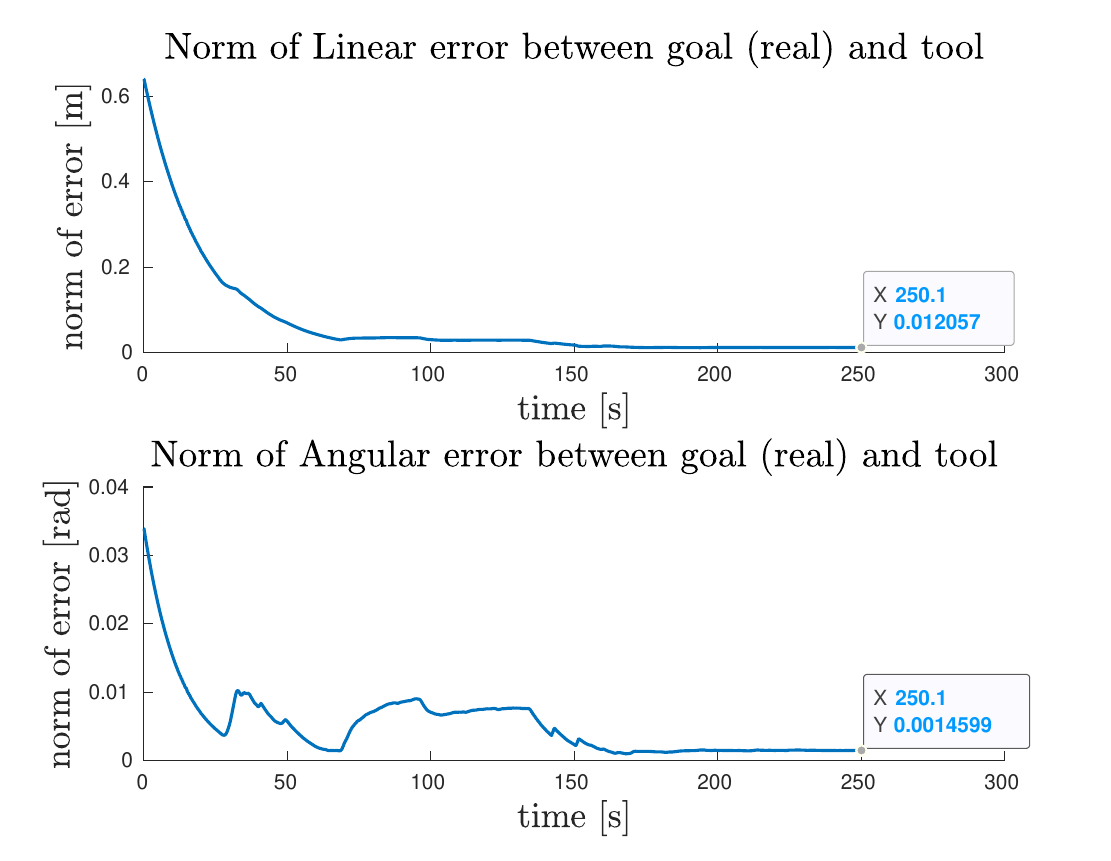}
		\includegraphics[width=6.5cm]{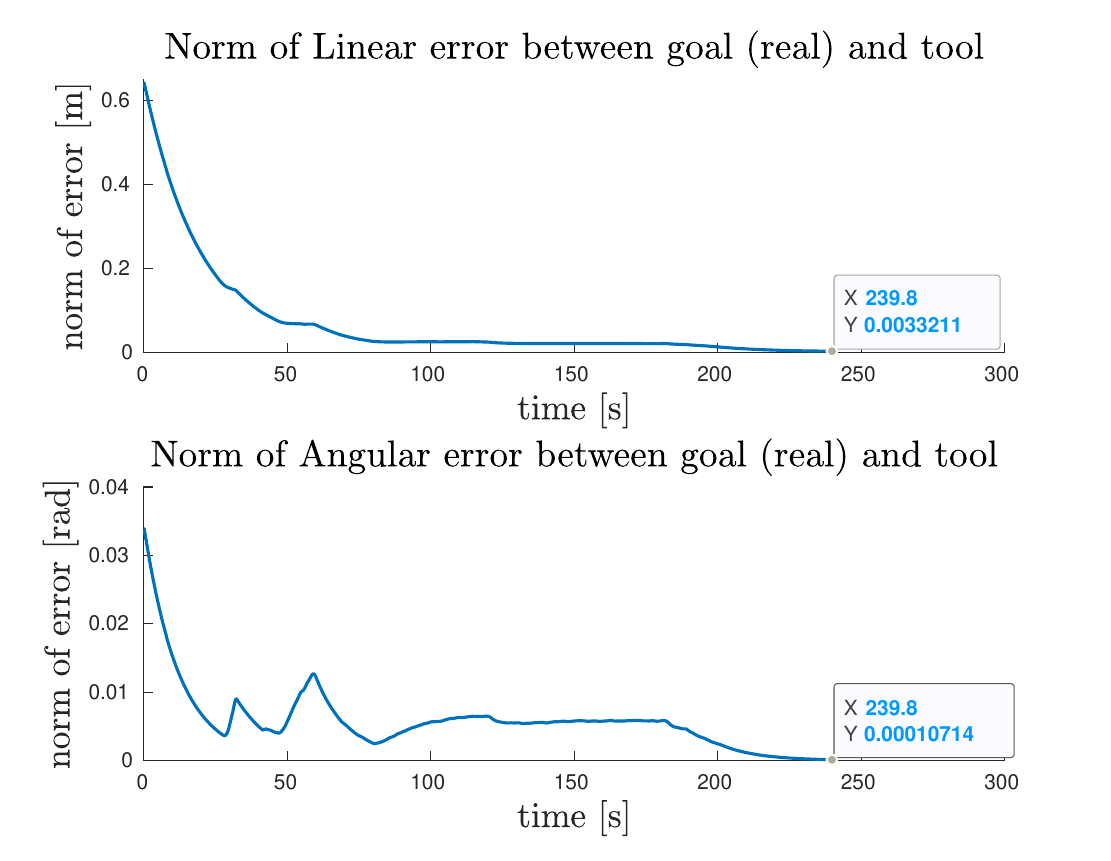}
		\includegraphics[width=6.5cm]{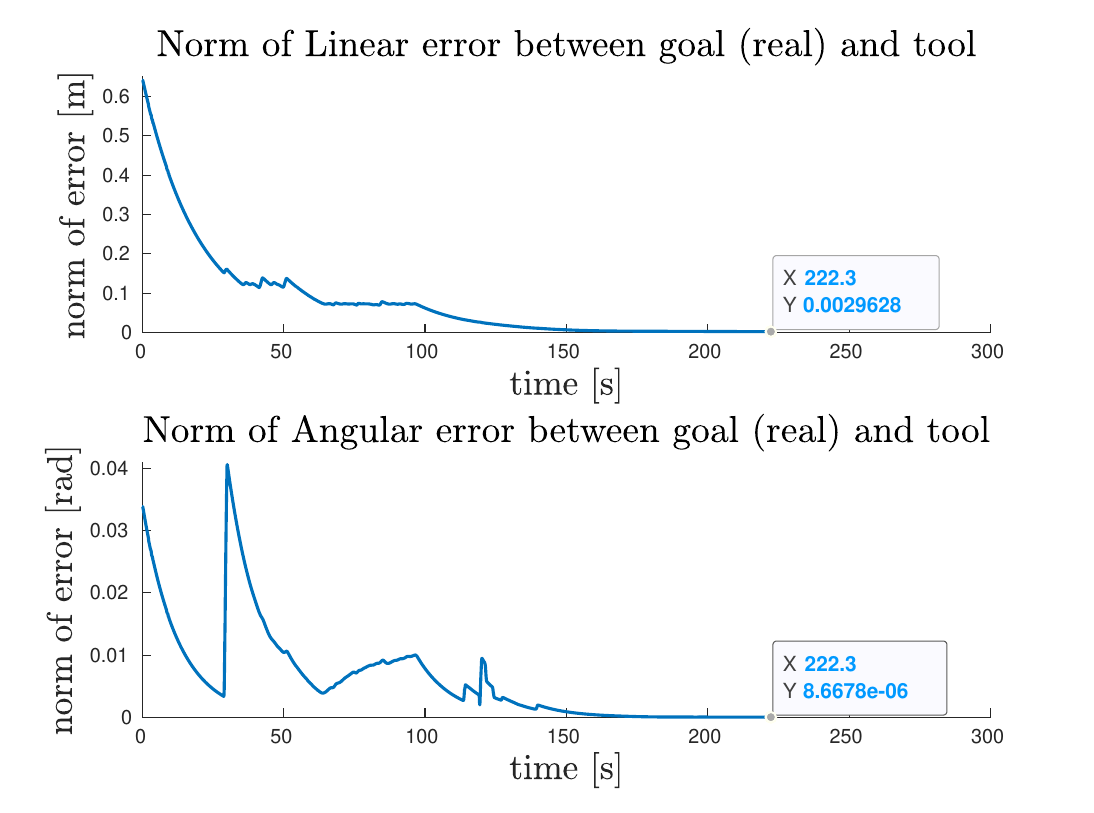}
	}
	\vspace{10px}
	\caption[Plots of comparisons with and without Change Goal and Force objective]{Comparison of results of the three methods: vanilla, Change Goal and Change Goal with Force-Torque objective. The three upper plots show the norm of the force and the norm of the torque acting on the peg. It can be noticed that, in the cases where the goal is modified, at the end their norms goes to zero. The three lower plots show the norms of the error between goal and tool's tip. In the case without the Change Goal routine, the norms converge anyway, but to a slightly bigger value than the one of the other two cases. In fact, when the Change Goal routine is used, the initial hole's pose error tends to be corrected.}
	\label{fig:comparison_final}
\end{figure}

\begin{figure}[H]
	\centering
	\textbf{Error of 0.015 meter on x-axis of the goal\\}
	\textbf{with Change Goal routine and Force-Torque objective}\\
	\vspace{13px}
	\textbf{Forces and torques acting on the peg\\}
	\vspace{3px}
	\centerline{ 
		\includegraphics[width=15.5cm]{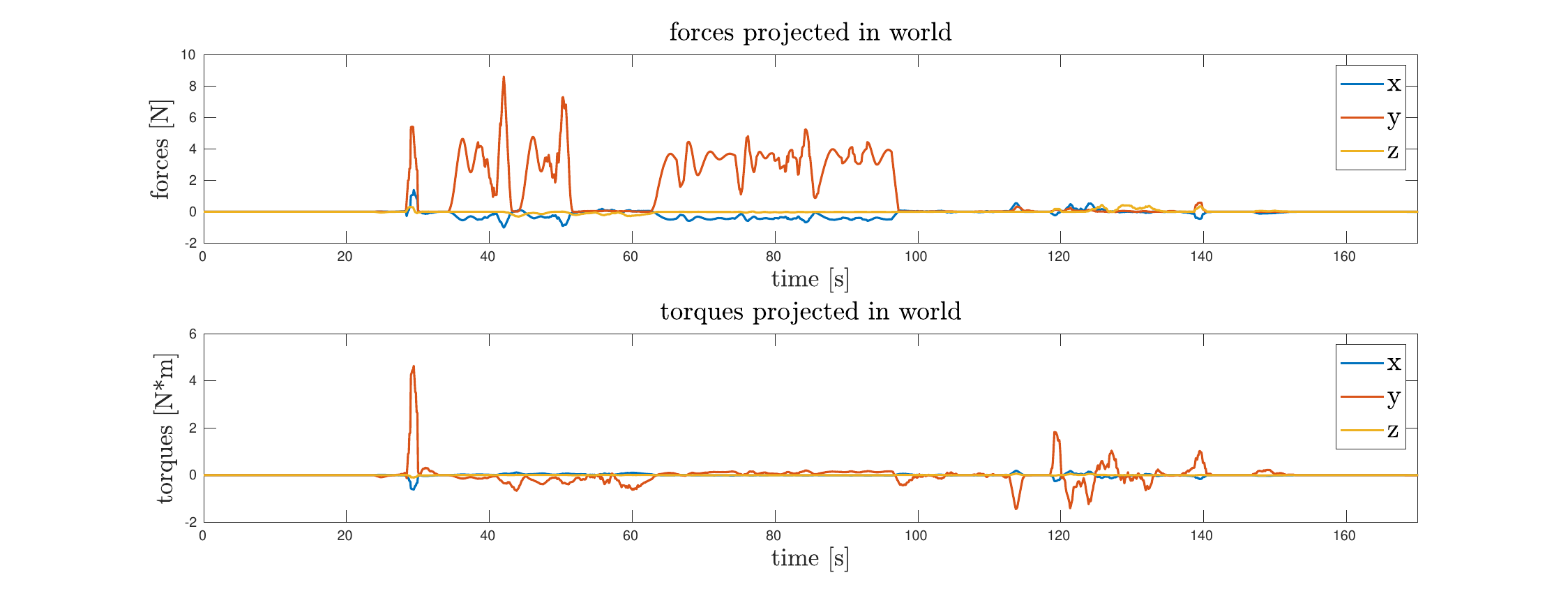}
	}
	\vspace{10px}
	\textbf{Force-Torque objective: References and Activations\\}
	\vspace{3px}
	\centerline{
		\includegraphics[width=15.5cm]{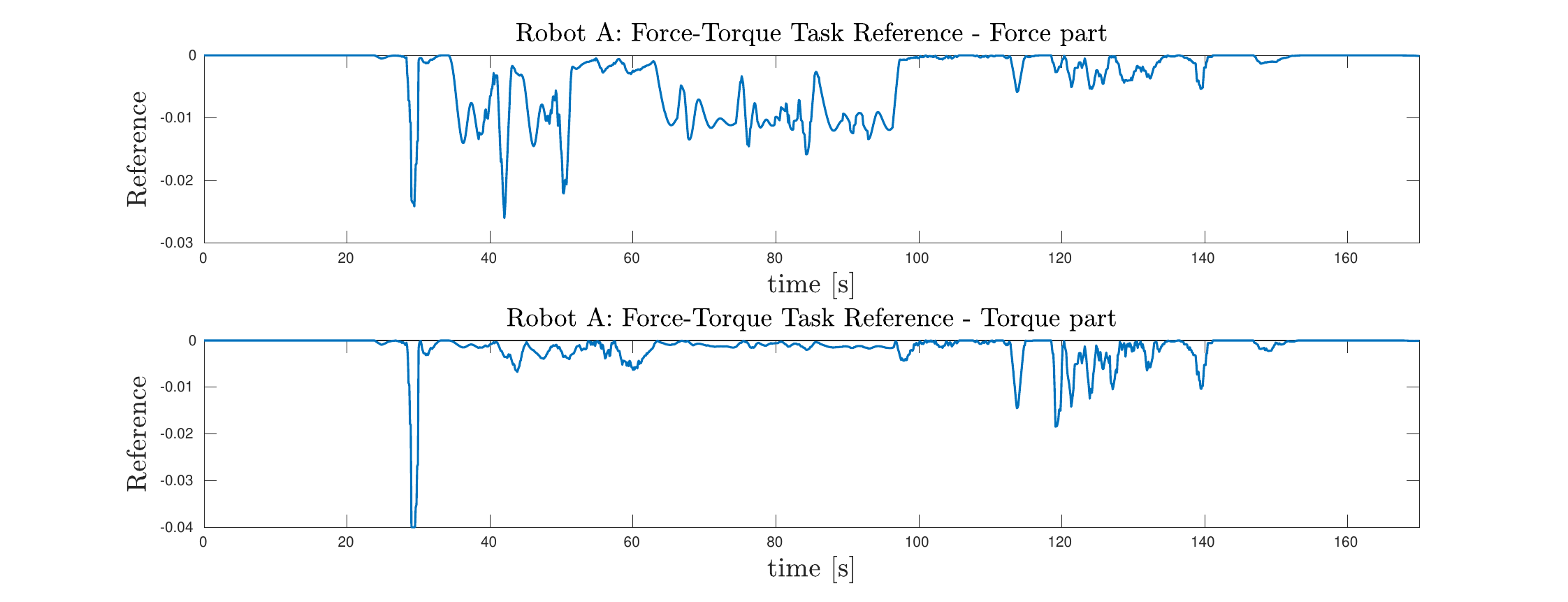}
	}
	\centerline{
		\includegraphics[width=15.5cm]{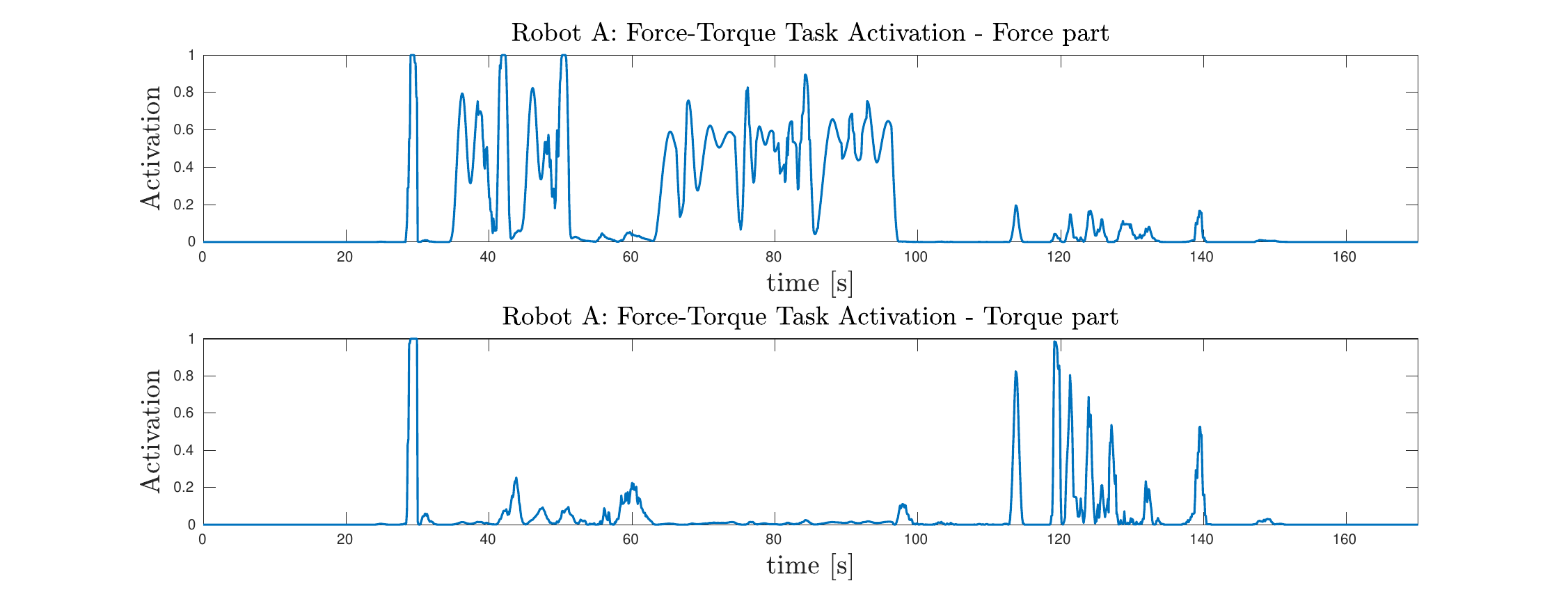}
	}

	\caption[Plots with reference and activation of the Force-Torque tobjective]{The forces and torques acting on the peg (above), and the corresponding generated references and activations  of the Force-Torque objective (below); they are calculated by robot A, but for robot B they are the same.}
	\label{fig:forceTaskActRef}
\end{figure}

\begin{figure}[H]
	\centering
	\textbf{Results with error of 0.015 meter on x-axis of the goal}\\
	\textbf{with Change Goal routine and Force-Torque objective}\\
	\vspace{15px}
	\textbf{Velocities generated by the Force-Torque objective only\\}
	\vspace{5px}
	\centerline{ 
		\includegraphics[width=9cm]{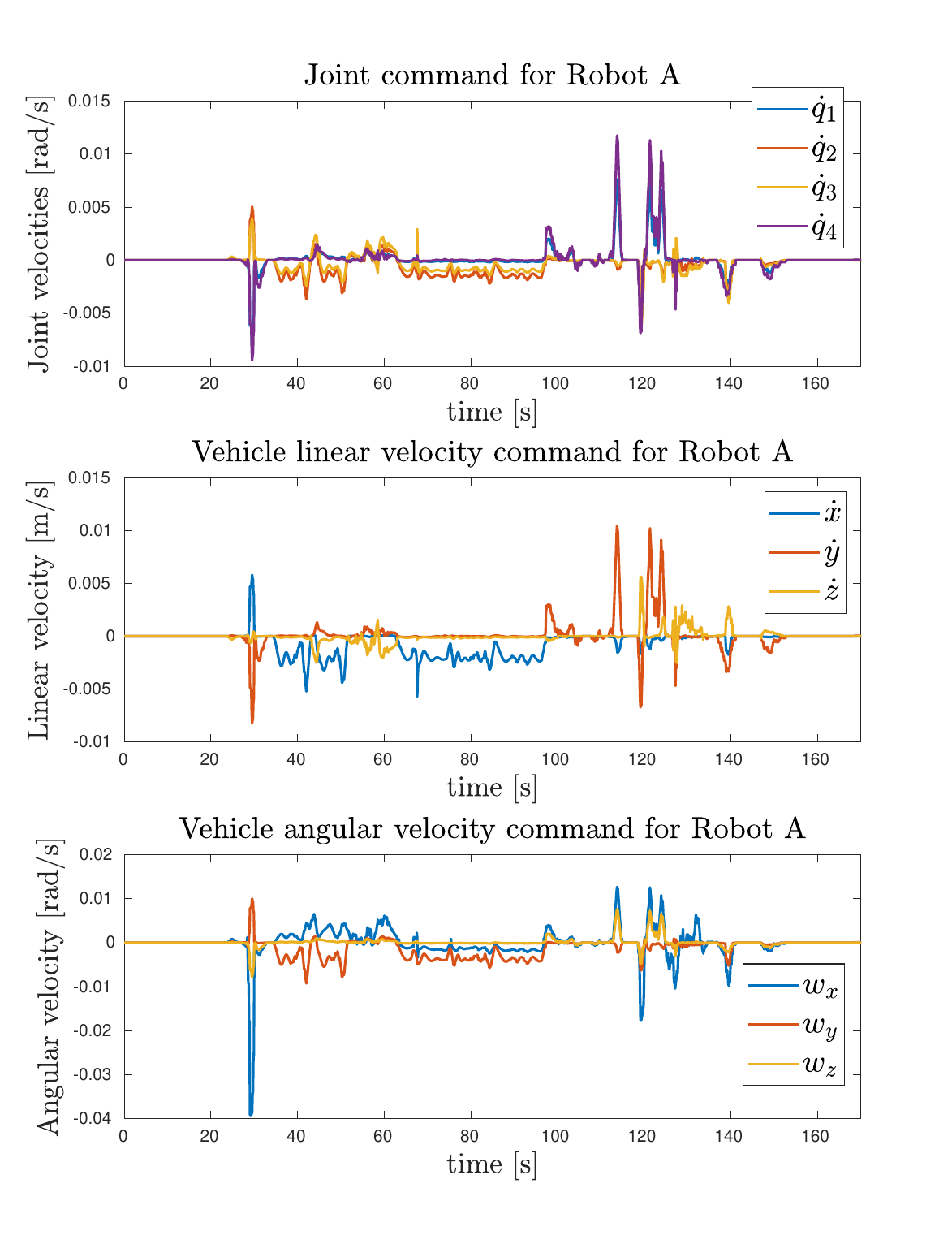}
		\includegraphics[width=9cm]{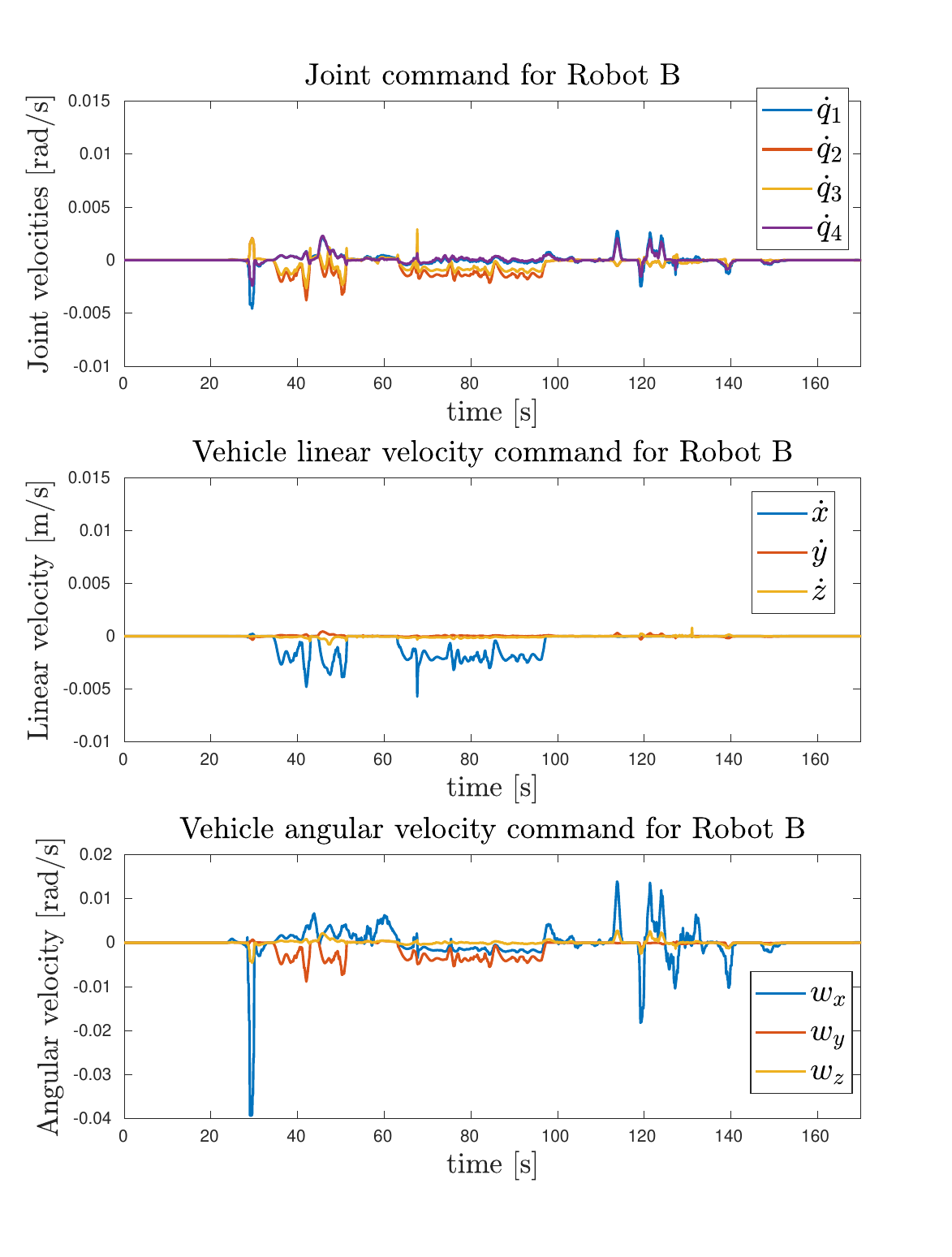}
	}
	
	\caption[Plots of the velocity command generated by the Force-Torque objective]{The velocity command generated by the Force-Torque objective, for the Robot A. These are the velocities that the task generates; the vectors depicted are simply the result of $\; \boldsymbol{J}^{\#}_{ft} \; \dot{\bar{\boldsymbol{x}}}_{ft}\;$ to show how the objective works. So they are not the real one applied to the system because with this formula higher priority objectives are not taken into consideration. For the two robots, the reference $\dot{\bar{\boldsymbol{x}}}_{ft}$ and the activation $\boldsymbol{A}_{ft}$ (visible in figure \ref{fig:forceTaskActRef}) are the same because they act with the same data; it is the Jacobian $\boldsymbol{J}^{\#}_{ft}$ which is obviously different and which makes the two plots dissimilar.}
	\label{fig:forceTaskVelocities}
\end{figure}

Besides changing the goal, it is useful to exploit the force-torque sensor also at kinematic level, using the provided information in the TPIK approach. The new added objective is described in section \ref{sec:forceTask}.\\
In figure \ref{fig:comparison_final} the results of the three different experiments are compared. It is visible that, when also the new objective is used, the forces and the torques have the smallest peaks. Meanwhile, the convergence of the error between the goal and the peg's tip is maintained as good as in the second experiment thanks to the presence of the same Change Goal routine.\\
In figures \ref{fig:forceTaskActRef} and \ref{fig:forceTaskVelocities} details on how this new objective works are shown. When some collisions happen, the reference and the activation grow to make the tool move in a way to reduce the force and the torque magnitudes. The figure \ref{fig:forceTaskVelocities} shows the velocities generated by the objective \textit{as if it was the only one} in the TPIK list, so they are not the real velocities given to the system.

\begin{figure} [H]
	\centering
	\textbf{Error of 0.015 meter on x-axis of the goal}\\
	\textbf{Linear and Angular component of the goal frame and the tool's tip frame}\\
\vspace{20px}
\textbf{Without Change Goal routine\\}
	\centerline{
		\includegraphics[width=22cm]{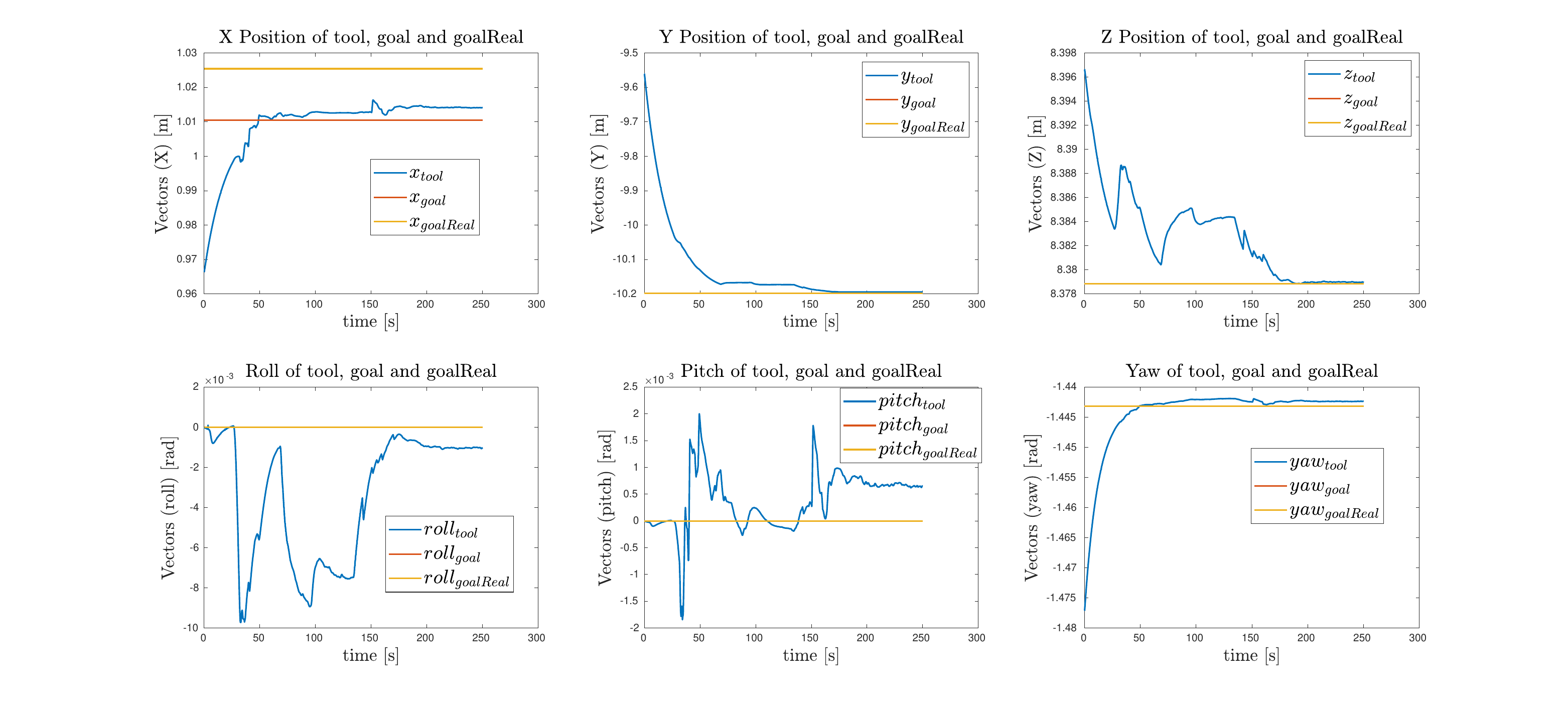}
	}
	\vspace{15px}
\textbf{With Change Goal routine}
	\centerline{
		\includegraphics[width=22cm]{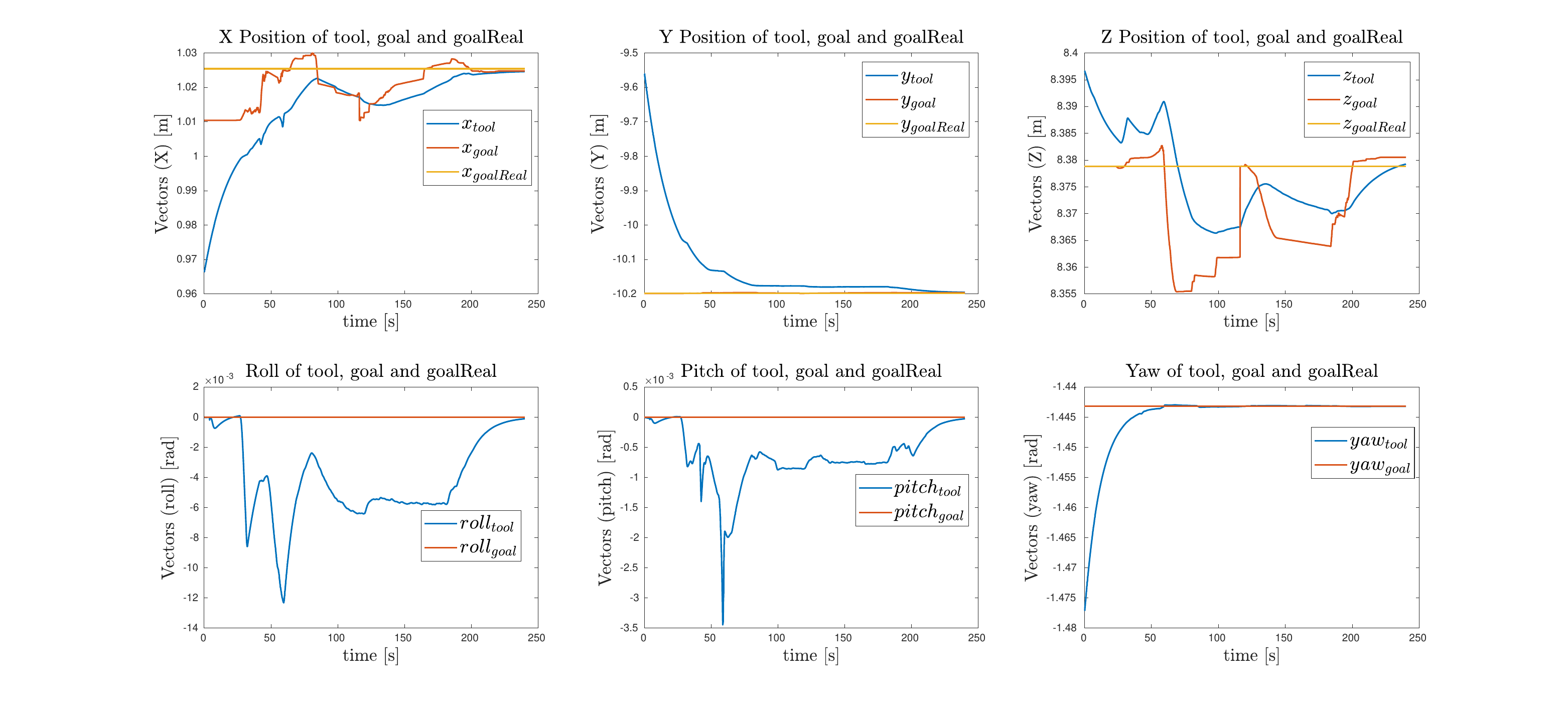}
	}
\end{figure}
\begingroup 
\captionof{figure}[Plots of Tool and Goal frames with and without Change Goal]{Results of the two experiments without and with the Change Goal routine (both without the Force-Torque objective). The plots show the pose of the goal frame $ \langle t \rangle$ and of the tool's tip frame divided into their linear and angular components. The upper six plots show results without the Change Goal routine, the bottom ones show results with the routine. All the components are respect to the world frame. For the linear part, the yellow lines represent the position of the goal without errors. The red lines represent the position of the goal that the controller uses. Please note that in some plots the red and yellow lines are coincident because the component is know without errors \emph{and} it does not change. It is clearly visible that, when the goal is modified, the red lines go toward the yellow ones, \textit{correcting} the initial hole's pose error.}
\label{fig:Error_nothingandgoal_plot6}
\endgroup

\section{Results with the Hole's Pose Estimation by Vision}
\label{sec:finalTest}
\begin{figure}[H]
	\centering
	\includegraphics[width=11cm, height=5.3cm]{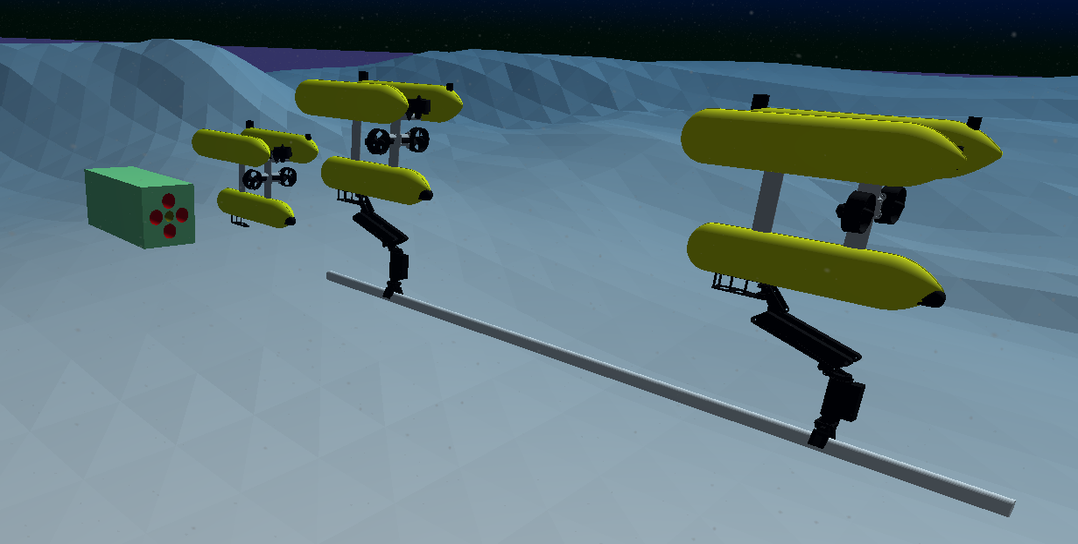}
	\includegraphics[width=11cm, height=3.2cm]{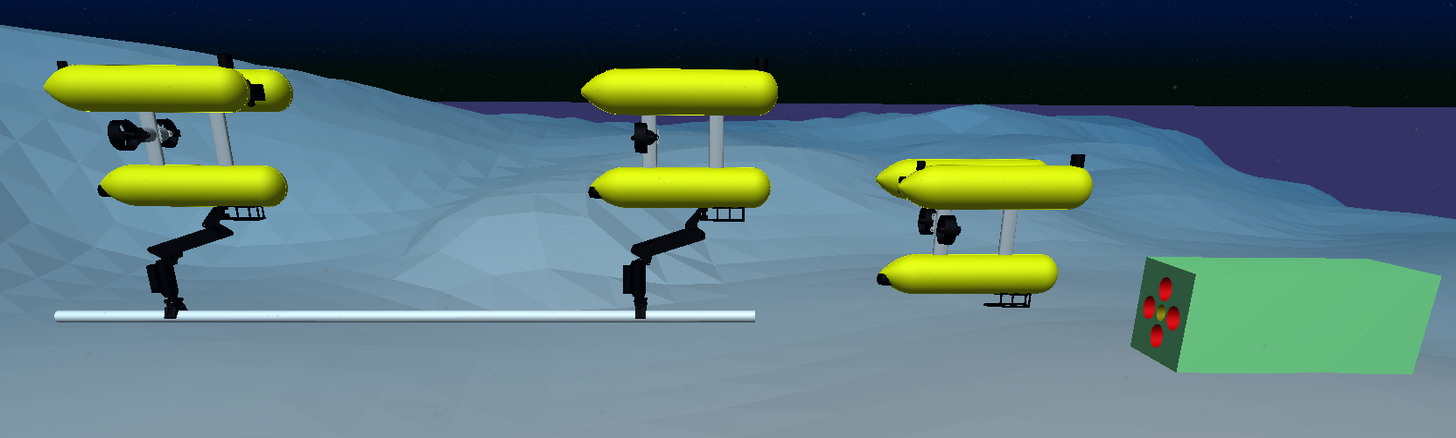}
	
	\caption[Scenario for the final test with Vision]{The starting position of the robots for the final experiment, where it is used the preliminary phase to estimate the hole's pose with vision.}
	\label{fig:uwsim_expAll}
\end{figure}

\begin{figure}[H]
	\centering
	\textbf{Screenshots from the final experiment }\\
	\vspace{5px}
	\centerline{
		\includegraphics[width=6cm, height=3.5cm]{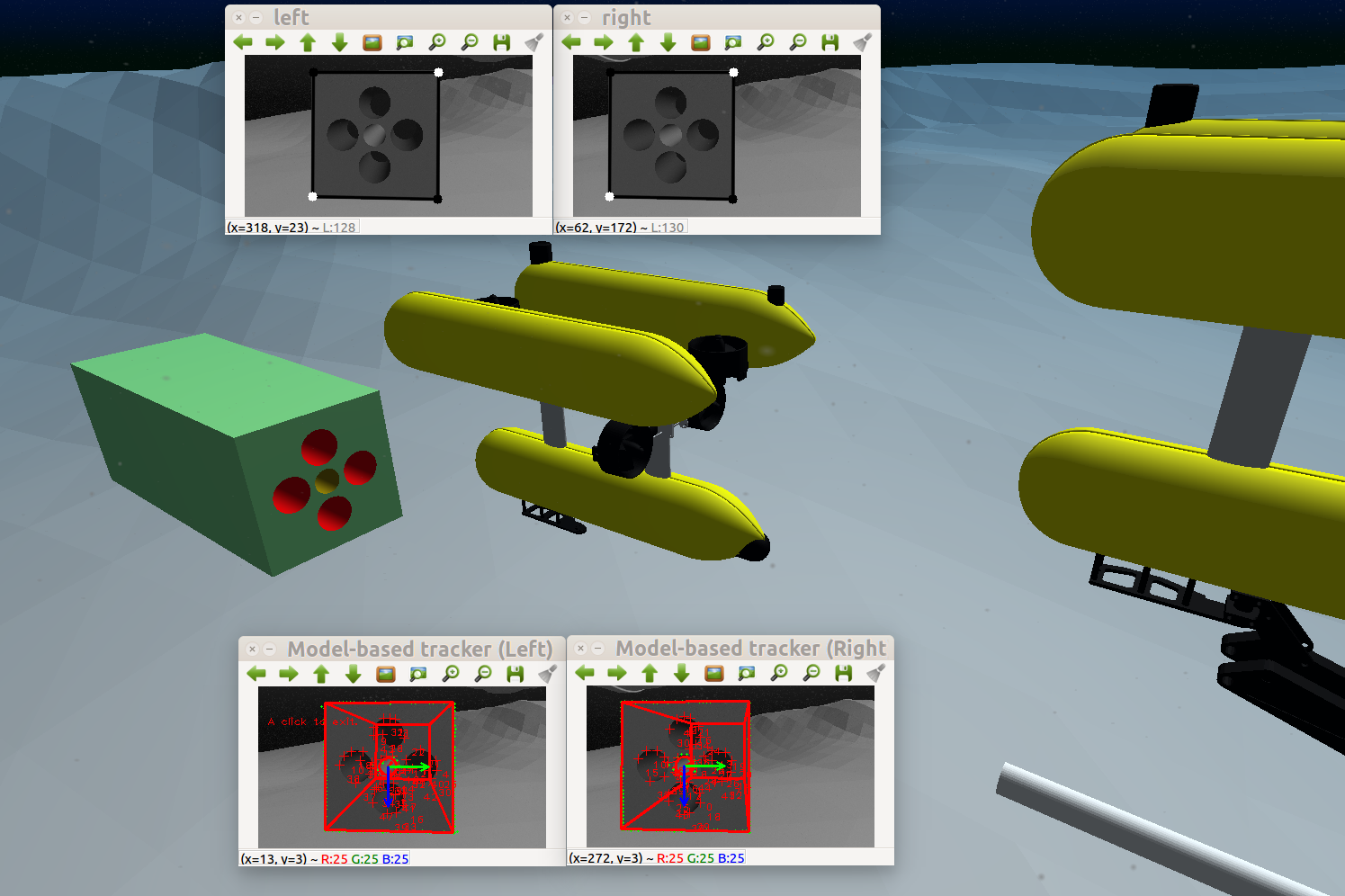}
		\includegraphics[width=6cm, height=3.5cm]{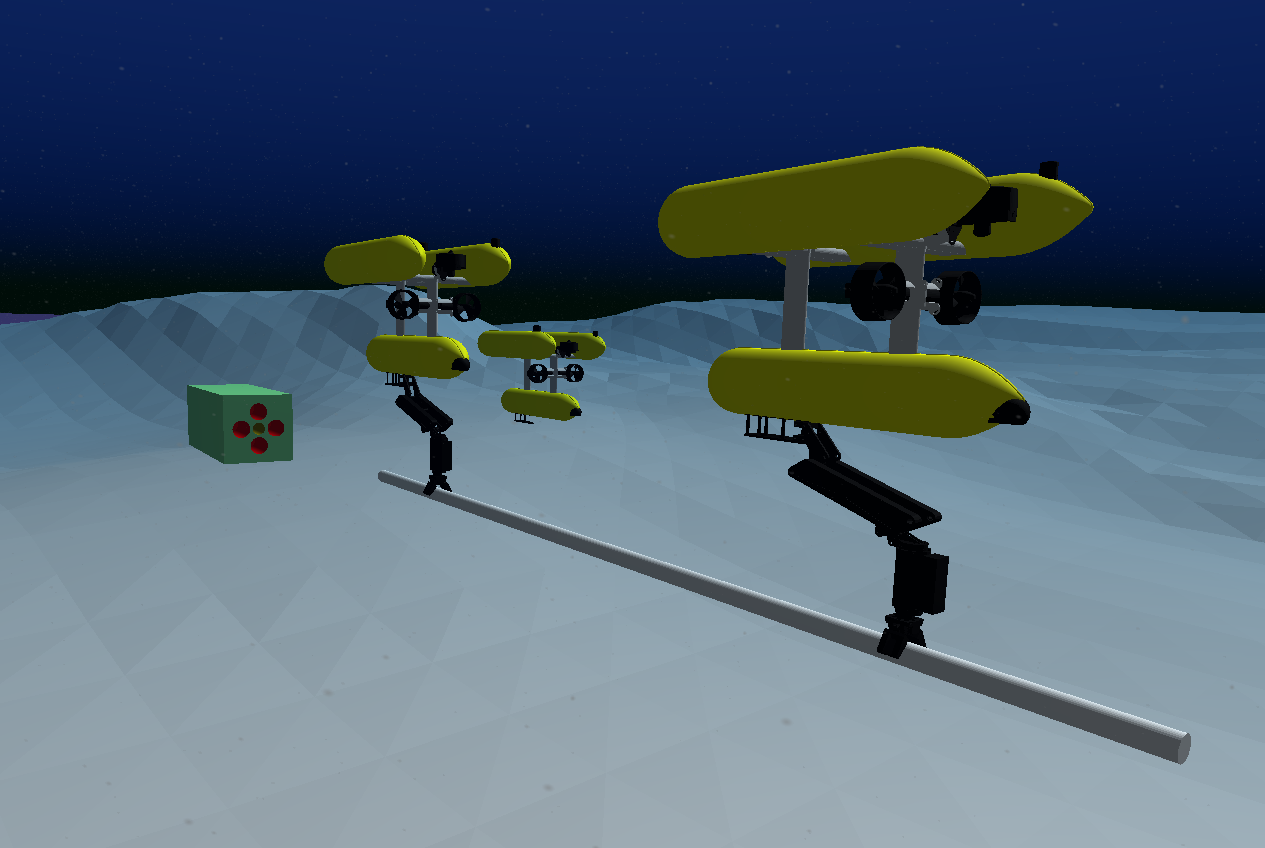}
	}
	\vspace{1px}
	\centerline{
		\includegraphics[width=6cm, height=3.5cm]{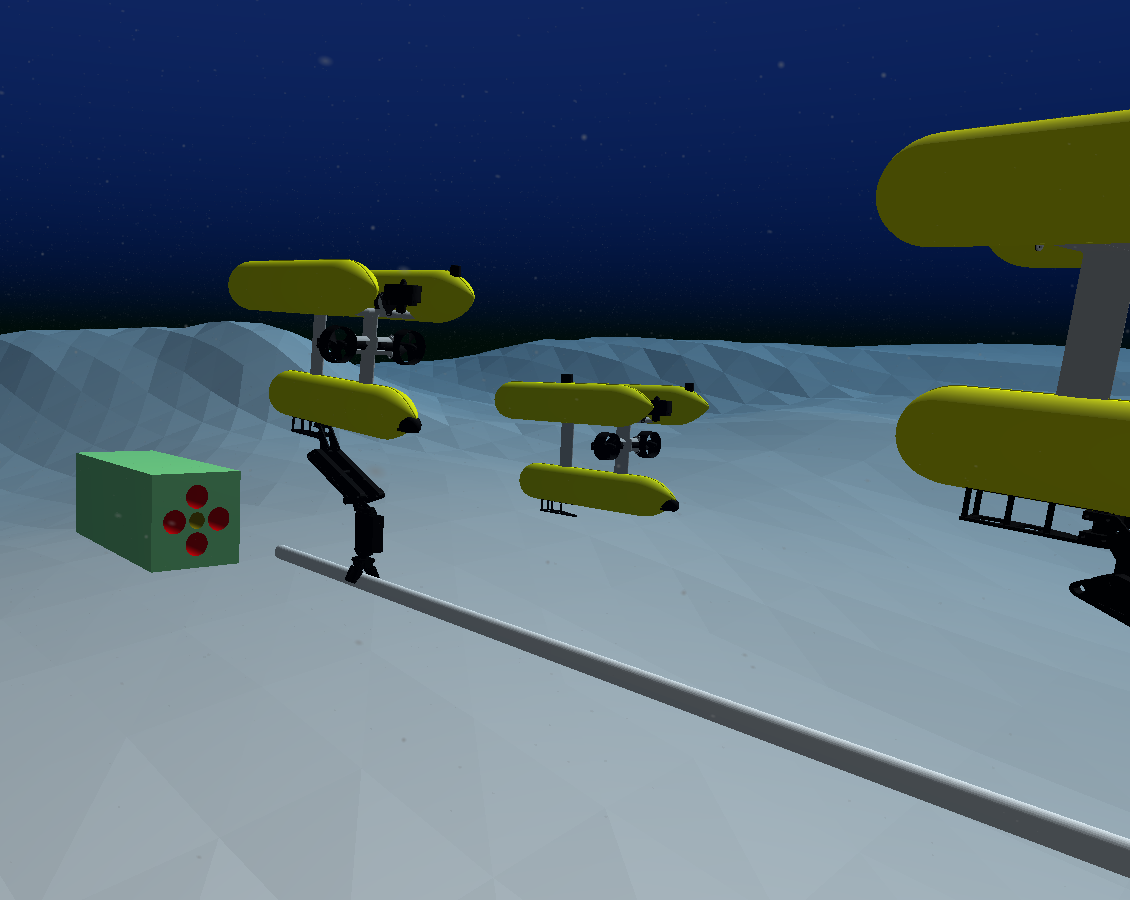}
		\includegraphics[width=6cm, height=3.5cm]{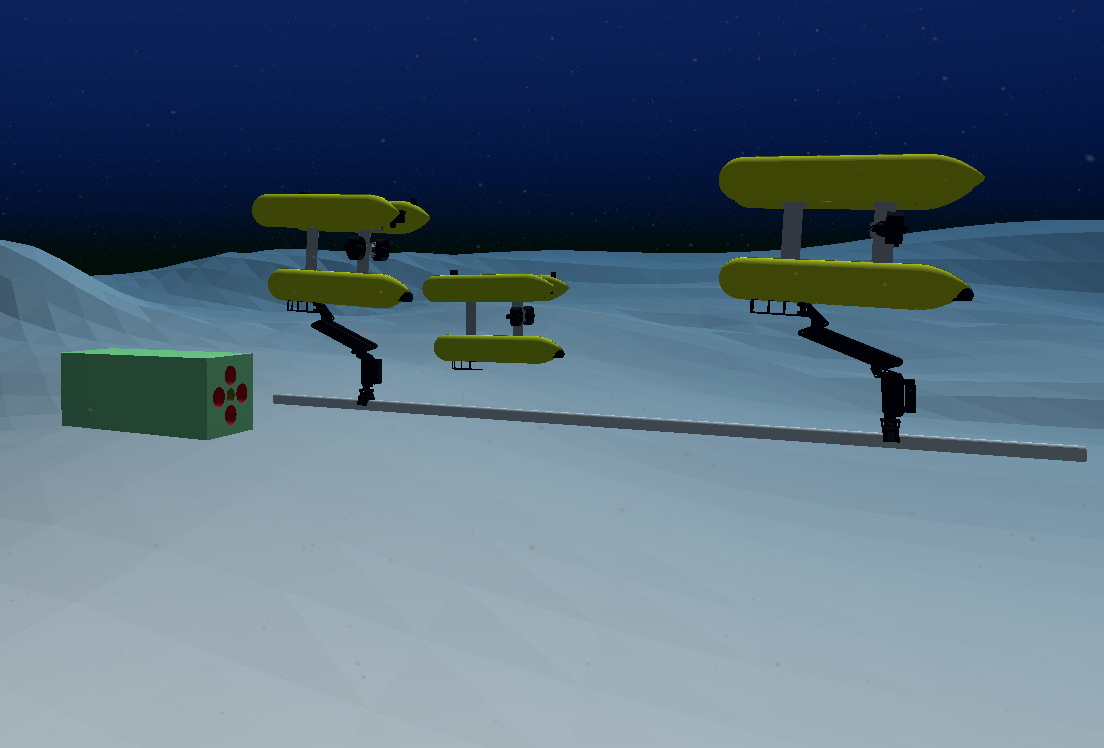}
	}
		\vspace{1px}
	\centerline{
		\includegraphics[width=6cm,height=3.5cm]{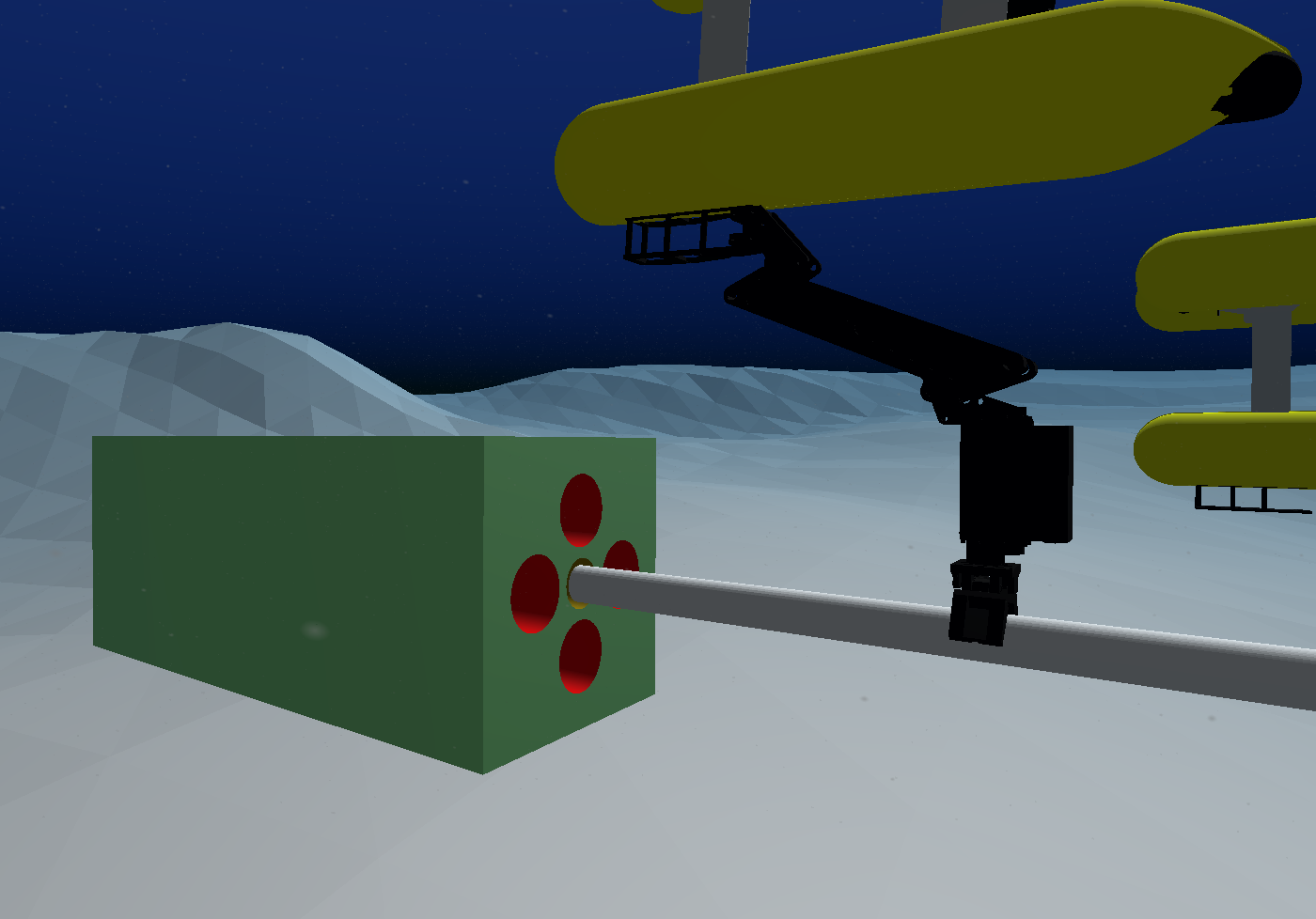}
		\includegraphics[width=6cm, height=3.5cm]{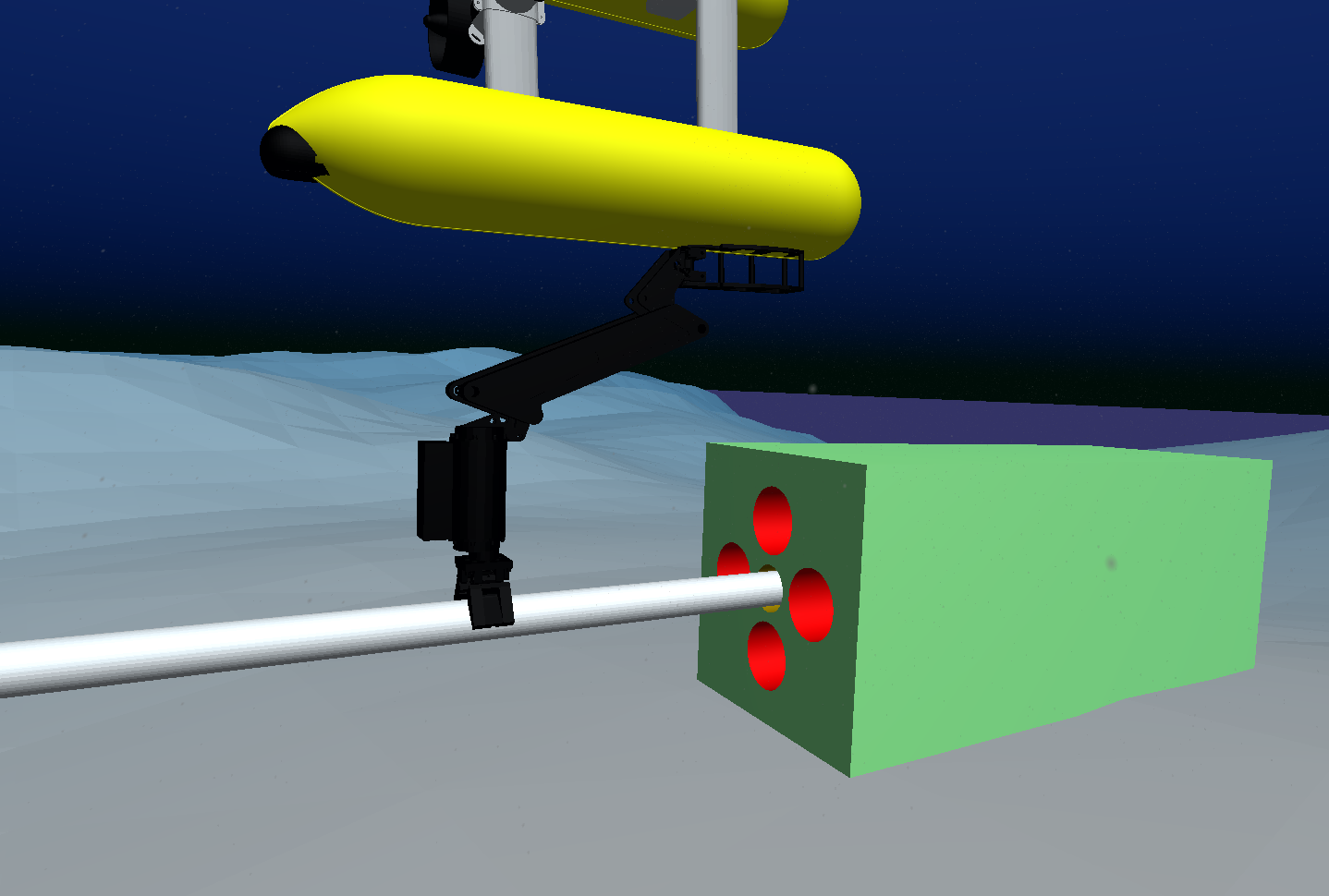}
	}
		\vspace{1px}
	\centerline{
		\includegraphics[width=6cm, height=3.5cm]{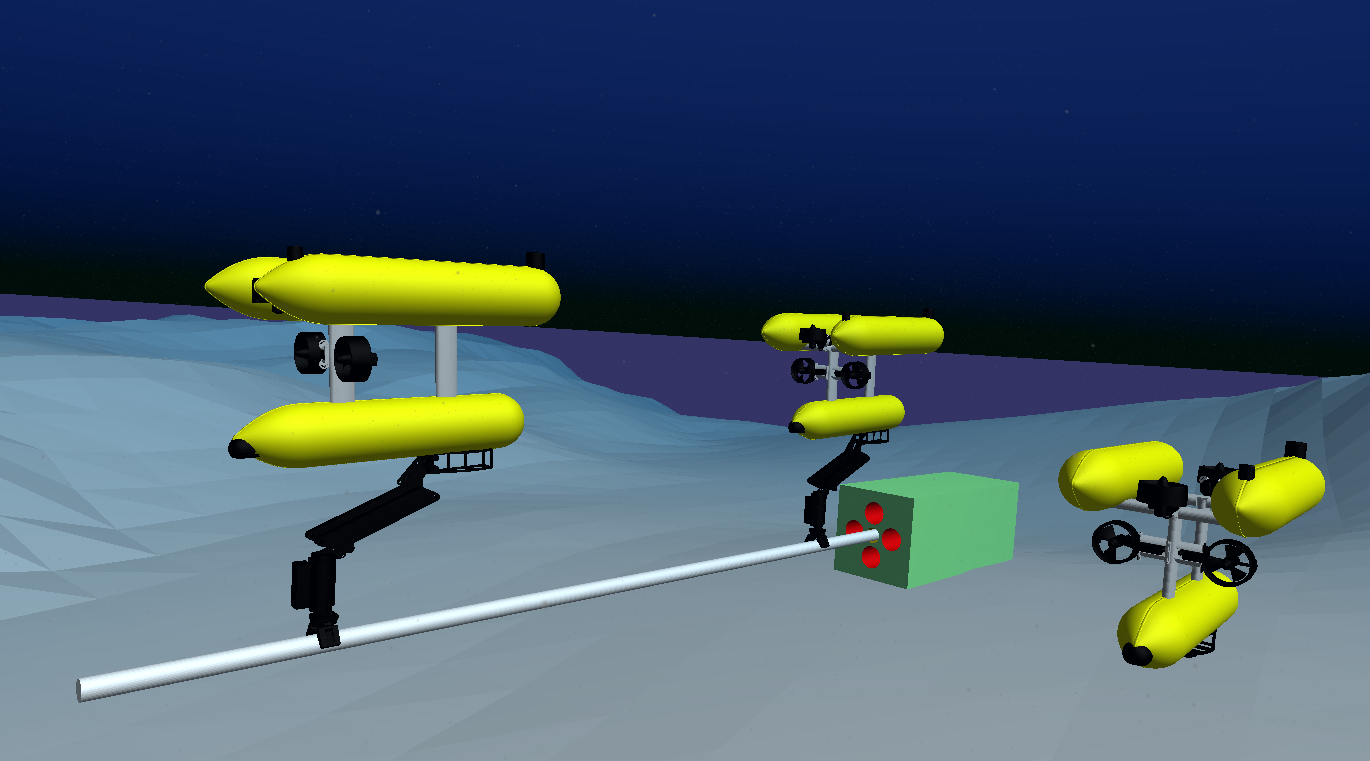}
		\includegraphics[width=6cm, height=3.5cm]{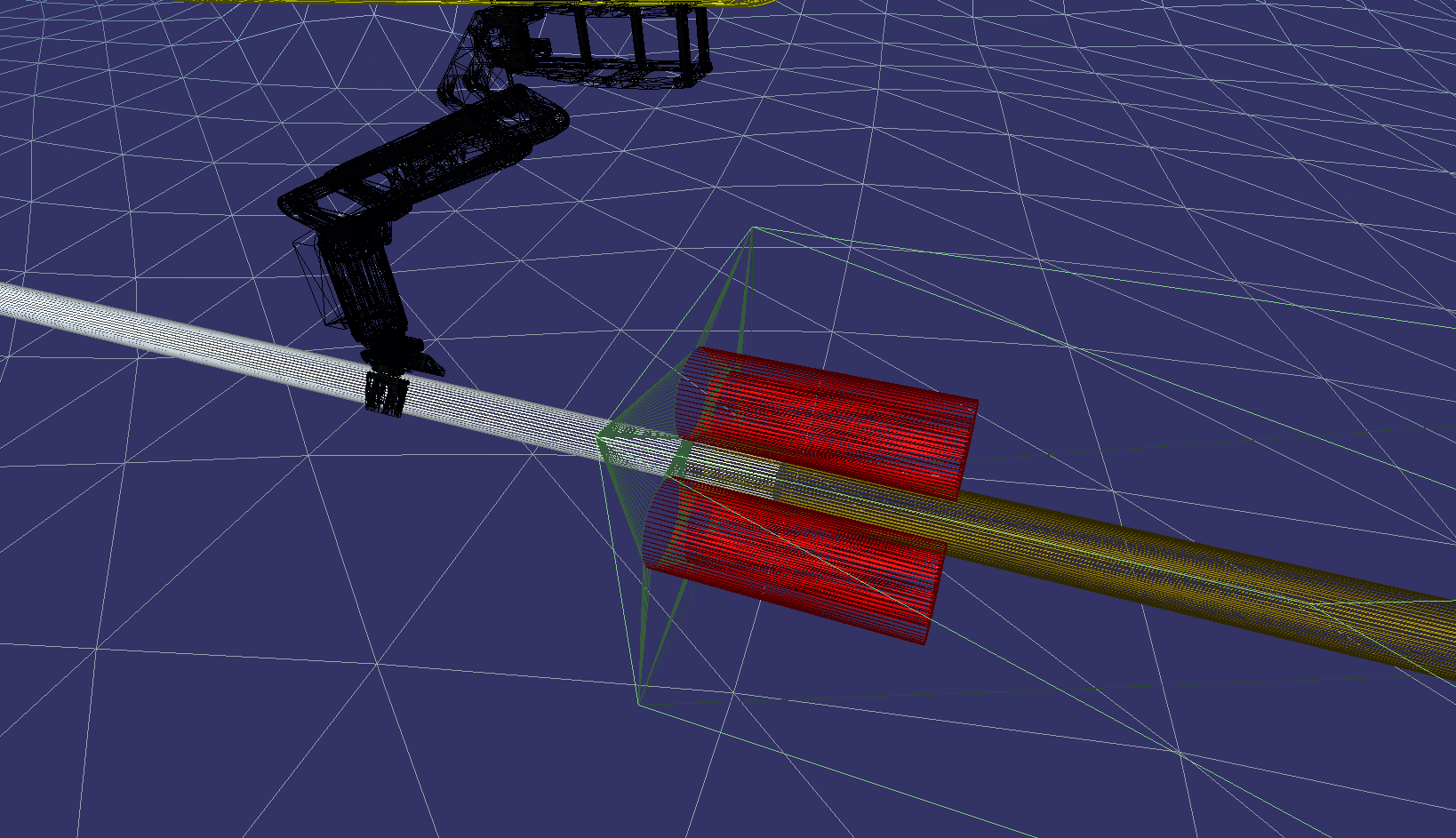}
	}
	\caption[Screenshots from the final experiment]{Screenshots from the final experiment with Vision. From the top to the bottom, from the left to the right: a) The Vision robot has detected the squares and it is tracking the hole's pose; b) The Vision robot is driven away and the carrying robots are ready to begin; c,d) the two robots are cooperatively transporting the peg to the hole; e) The first \enquote{contact} between the \textit{peg} and the \textit{hole}; f) The \textit{peg} is being inserted; g) The robots stop because the tool has reached the desired depth (0.2m); h) A polygon wire-frame view mode of the simulator to see the \textit{peg} inserted.}

	\label{fig:screenSimulation}
\end{figure}

\begin{figure}[H]
	\centering
	\textbf{Results with hole's pose estimation by Vision}\\
	\vspace*{20px}
	\centerline{
		\includegraphics[width=8.5cm]{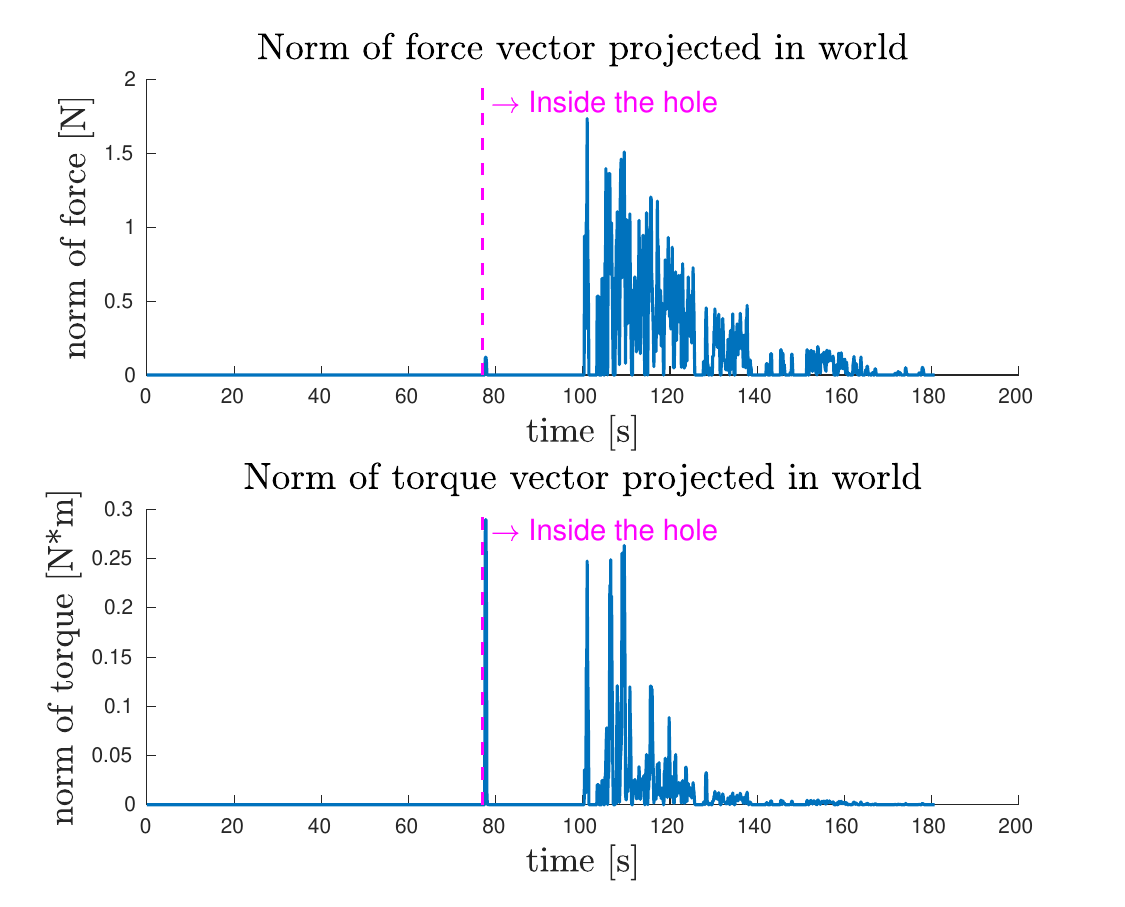}
		\includegraphics[width=8.5cm]{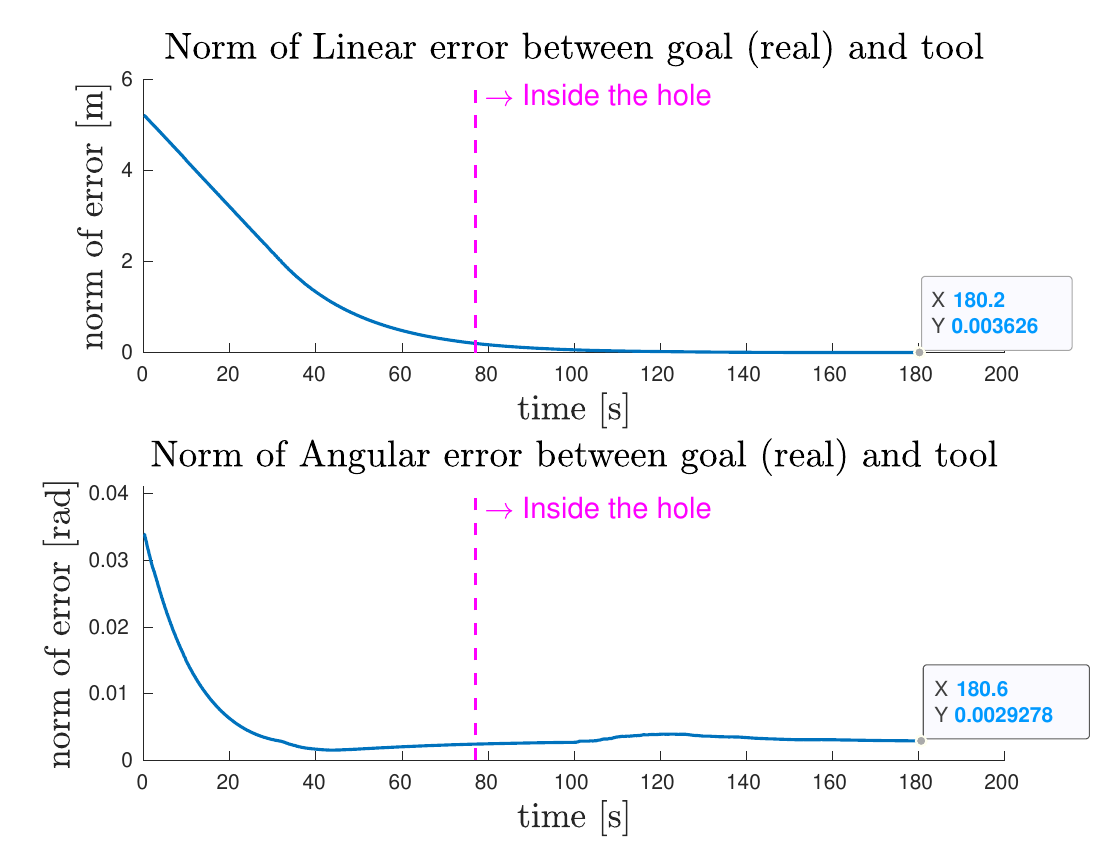}
	}
	\vspace{30px}
	\centerline{
		\includegraphics[width=19.5cm]{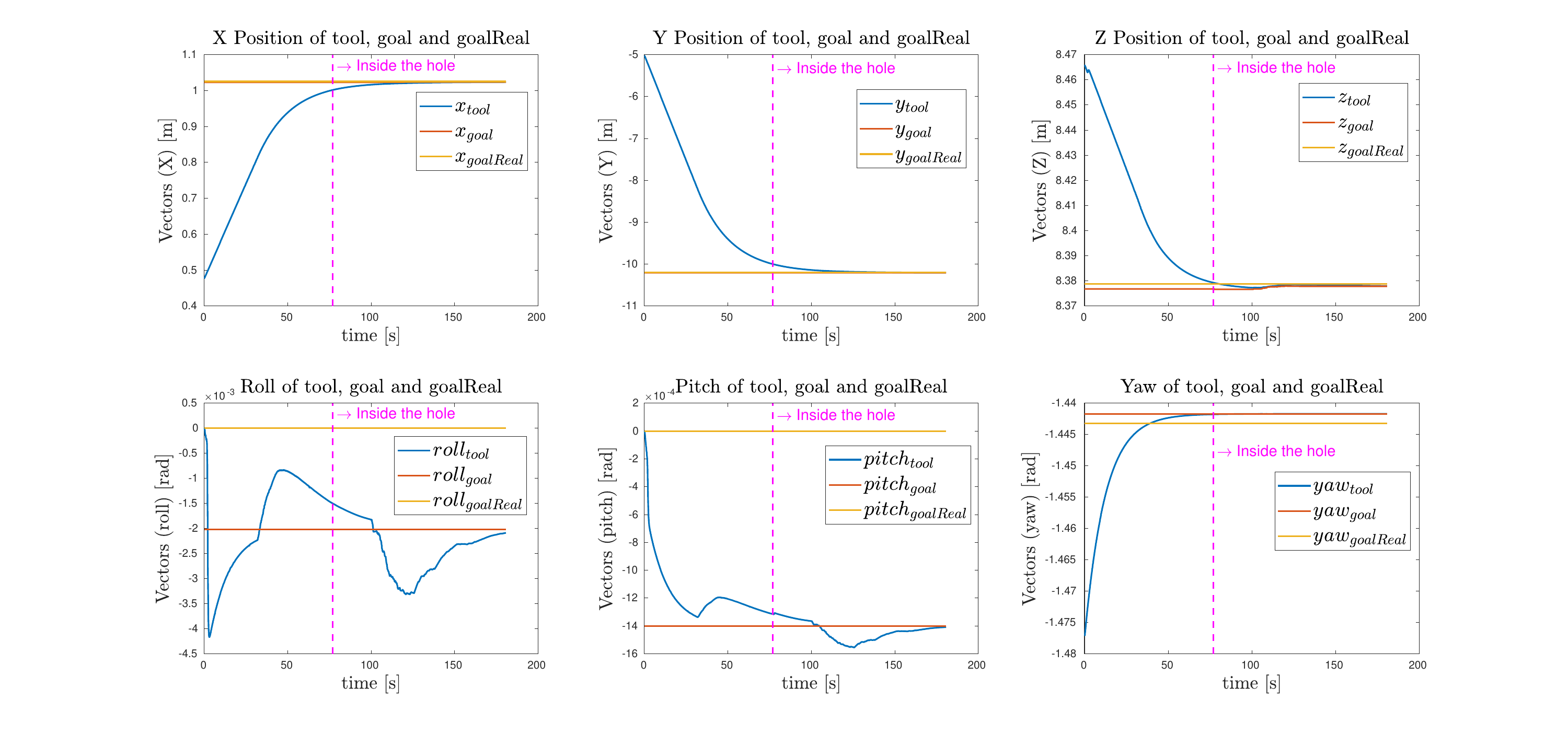}
	}
	\vspace{10px}
	\caption[Plots of results with Hole's pose estimation by Vision]{Results with the hole's pose estimated by the best one of the tested vision algorithms. At second 77 the peg's tip goes inside the hole (magenta vertical lines). Being the pose estimation really good, forces and torques are not so big  as in the previous tests (upper left plot). Furthermore, modification of the goal are almost not noticeable (lower plot). The positional error from goal to peg's tip  converges to a small value (upper right plot). The visible error for the angular part is due to the fact that the orientation of the goal frame has some imprecisions, due to not perfect hole's pose estimation.}
	\label{fig:expWithVisio}
\end{figure}

\begin{figure}[H]
	\centering
	\textbf{Results with hole's pose estimation by Vision}\\
	\textbf{Cooperative system velocities for robots A and B (after cooperation)}
	\vspace{20px}
	\centerline{
		\includegraphics[width=9cm]{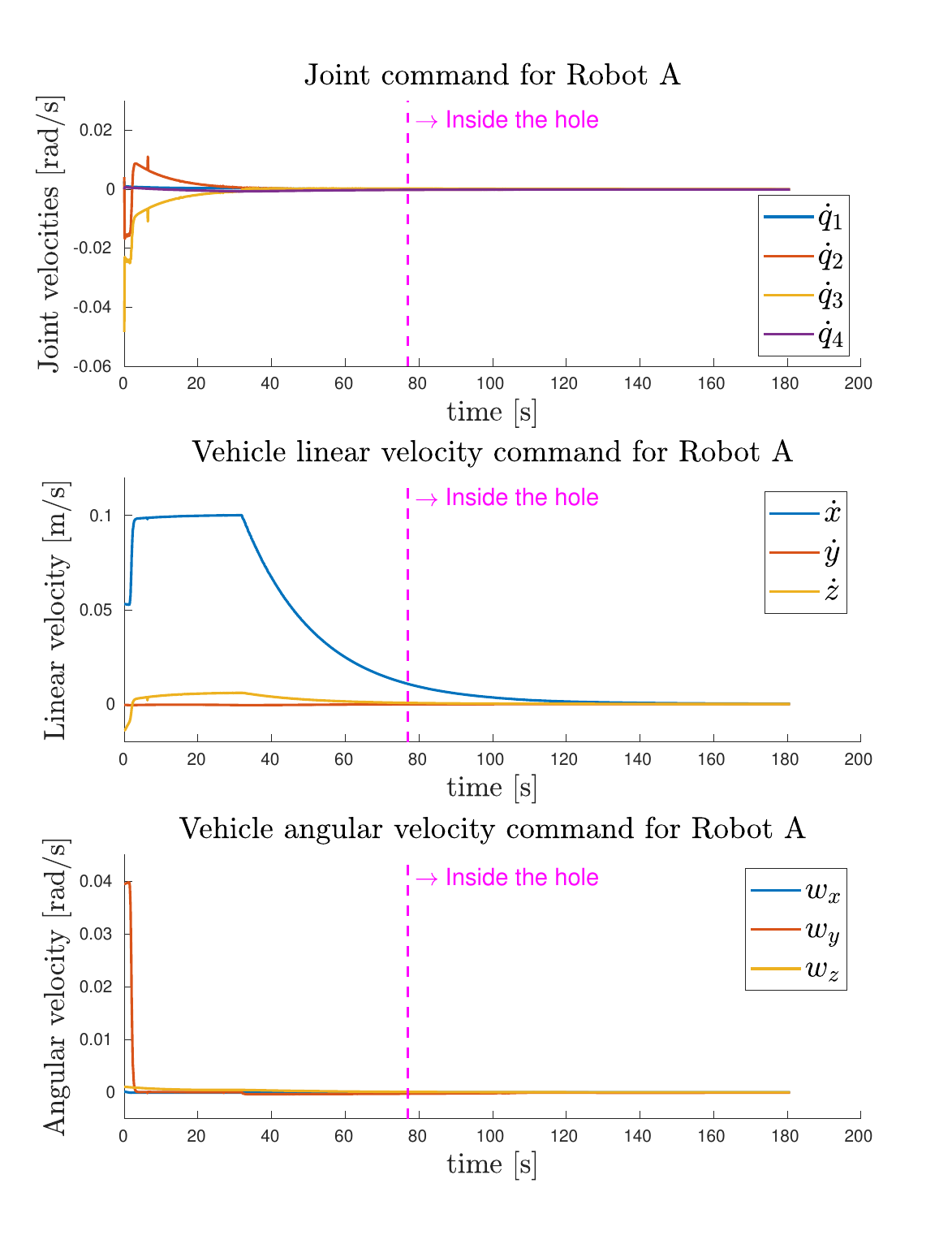}
		\includegraphics[width=9cm]{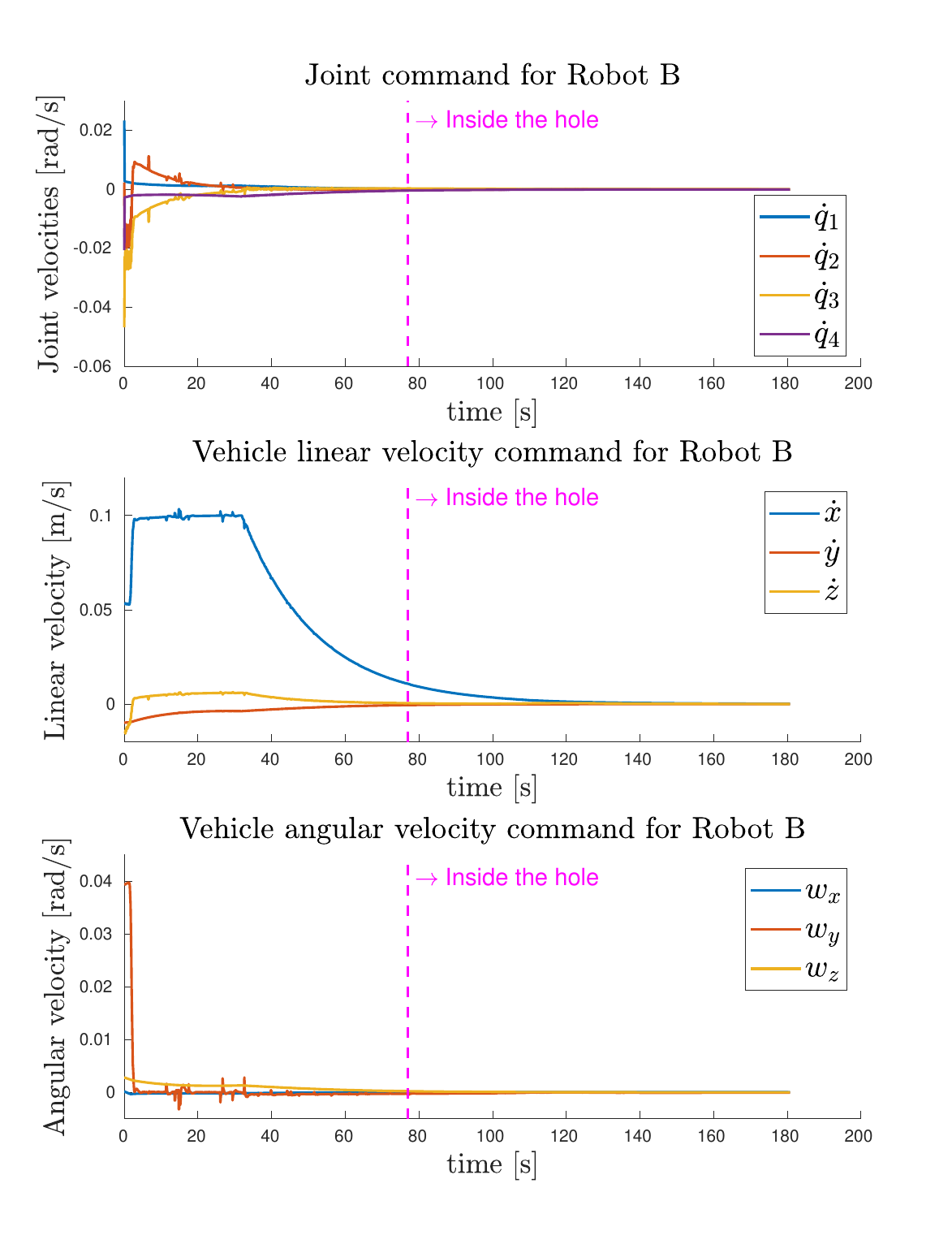}
	}

	\caption[Plots of cooperative robots velocities]{The system velocities $\dot{\hat{\boldsymbol{y}}}_a$ and $\dot{\hat{\boldsymbol{y}}}_b\,$, outputs of the kinematic layer after the cooperation policy (section \ref{sec:coopScheme}) and the arm-vehicle coordination scheme (section \ref{sec:armVehScheme}). These are not always the \emph{real} robot velocities because they do not include velocities caused by collisions (for robot A) and by firm grasp constraint (for robot B).}
	\label{fig:expWithVisioVel}
\end{figure}

\begin{figure}[H]
	\centering
	\textbf{Results with hole's pose estimation by Vision}\\
	\textbf{Cooperative tool velocity (after cooperation)}
	\vspace{20px}
	\centerline{
		\includegraphics[width=12.5cm]{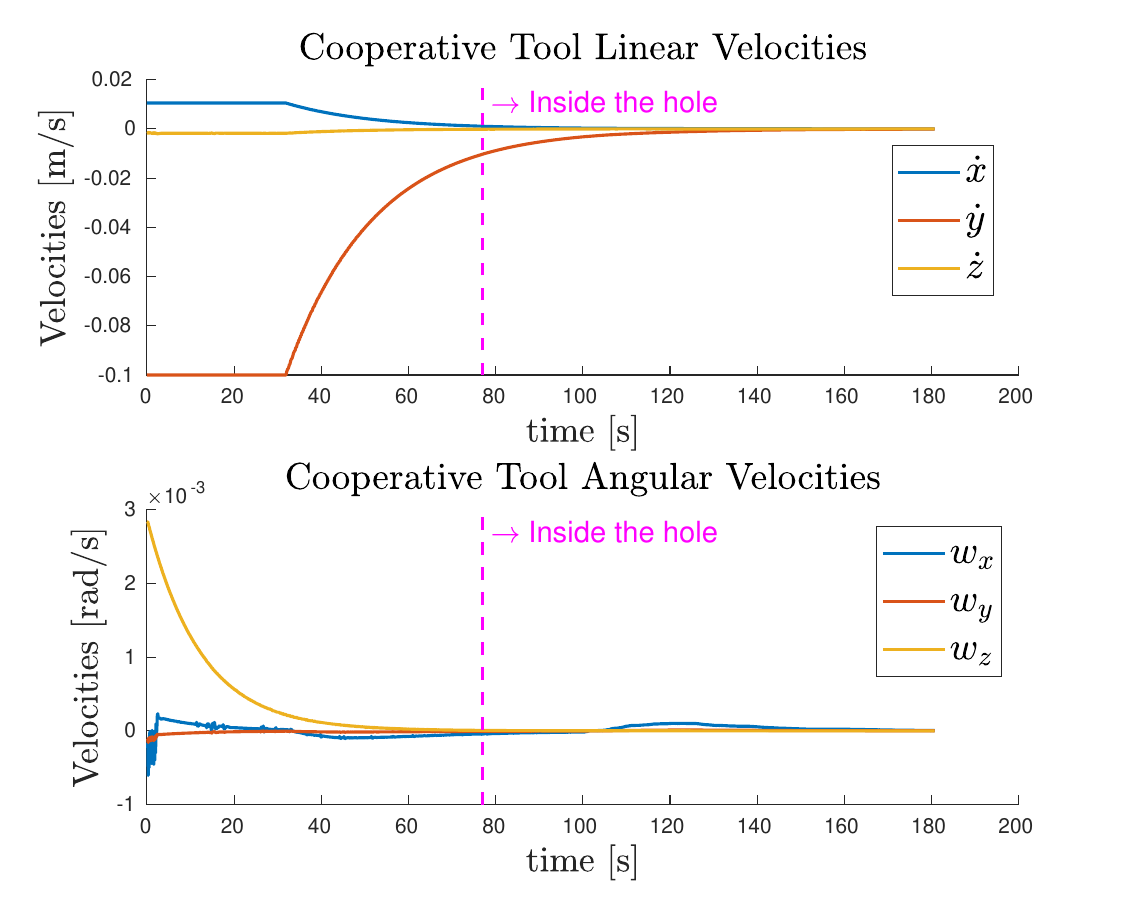}
	}
	\caption[Plots of cooperative tool velocity]{The tool velocity $\dot{\tilde{\boldsymbol{x}}}_t$, the one that the coordinator sends to both robots during the coordination policy. It is the \emph{feasible} one that \emph{both} robots provide to the tool (section \ref{sec:coopScheme}), and it is caused by the system velocities of figure \ref{fig:expWithVisioVel}. So, it is not always the \emph{real} velocity that the tool has because collisions and firm grasp constraint are not included.}
	\label{fig:expWithVisioVelTool}
\end{figure}

In this section, results with the preliminary vision phase are presented.\\
Differently from the previous experiments of section \ref{subsec:resultsControlError}, the error of the hole's pose when using the Vision robot is not so influential on the mission. In norm, the pose estimation error is less than $0.006$ meters for the linear part and $0.01$ radians for the angular part (with the best method, as shown in figure \ref{fig:squareErrors} of section \ref{subsec:trackResult}).\\ 
However, the experiment is interesting not only because it puts all the mission phases together, but also because it shows the outcome when the error is \enquote{spread} among all linear and, especially, angular component (which was not considered before). Even more, the carrying robots start farther from the hole than before. In this simulation, both the Change Goal routine and the Force-Torque objective are used.\\
The test's details described at the beginning of section \ref{sec:resultNoVisio} are still valid, except for the initial position of the two carrying robot. This time, the distance between hole and peg's tip has component $[3.590, 0.039, -0.041]$ for the linear part (and the same as previous for the angular part: $[0, 0, 1.942]$ \textit{roll, pitch, yaw}, in degrees).\\
Some screenshots that show the main phases of the simulation are visible in figure \ref{fig:screenSimulation}. The interesting plots about the performances are shown in figure \ref{fig:expWithVisio}. To give an idea of the magnitude of the velocities involved, cooperative system velocities $\dot{\hat{\boldsymbol{y}}}_a$ and $\dot{\hat{\boldsymbol{y}}}_b\,$, and cooperative tool velocity $\dot{\tilde{\boldsymbol{x}}}_t$ are displayed in figure \ref{fig:expWithVisioVel} and figure \ref{fig:expWithVisioVelTool}. Please note that these velocities do not include the collision propagation and the firm grasp constraint, they are only the output of the kinematic layer.

\section{Results Discussion}
\label{sec:resDisc}
In this section, considerations about the results shown in the previous pages are made.\\

\subsection{Perfectly Known Hole's Pose Discussion}
When the goal frame is known without error, the results are good even without the Force-Torque objective and the Change Goal routine.\\
This is clearly visible in figure \ref{fig:noErrorPlots}. Despite no presence of errors, some collisions happen anyway. This is due to the fact that the peg is driven  directly toward the goal (inside the hole), without considering that there is the hole structure. Anyway, the forces and torques acting on the peg \enquote{help} the insertion phase because they drive \textit{naturally} the tool inside the hole. The peaks of the magnitudes are due to the first contact with the hole: even if the pose is perfect, this does not mean that a perfect alignment is present \textit{before} reaching the goal frame.\\
 
The lower left plot of figure \ref{fig:noErrorPlots} shows the tool velocities caused by the collisions, and in the lower right one we can see that the second tool \enquote{follow} the velocities of the first one, due to the firm grasp constraint (around second 40). In other words, it is as if the robot B would be dragged by motions causes by collisions on robot A.\\
In the same plot, the chaotic velocities caused at the beginning are due to a not-perfectly coincident initial position of the two tools, that is impossible to achieve due to numerical errors of the simulation. Anyway, after a while they go to zero, so this initial error does not affect the rest of the experiment.\\

The upper right plot of figure \ref{fig:noErrorPlots} shows how the error converges along its components. The small peak in roll is caused by the initial collisions with the hole.\\

Even if a perfect goal pose is obviously impossible to have, this experiment is useful to understand how the peg is inserted if ideal conditions are present. This is important: from now we know that, in any case, the forces and torques acting on the peg may help \textit{naturally} the insertion phase.\\

\subsection{Change Goal Routine Discussion}
When an error in the goal pose is present, the results obviously get worse. The first noticeable thing is in figure \ref{fig:comparison_final}: due to the error the collisions are more present during the insertion (first plot). An interesting thing is that the peg is inserted at the wanted depth anyway (as can be seen in the right lower plot, where the norm converges to a small value). An issue is that the forces are not nullified at the end, which means that continuous pressure between the peg and the hole is present.\\

The first idea to reduce the number of the collisions is to introduce the routine which change the goal frame. From the plots in the middle of figure \ref{fig:comparison_final}, we see that the overall behaviour is improved: the forces and torques have a littler norm in general (upper plot) and the final position of the tool is more precise (lower plot).\\

In fact, from figure \ref{fig:Error_nothingandgoal_plot6}, the six lower plots show us that the error in $x$ is corrected by the routine. It can be noticed that also the $z$ component is modified, even if it has no initial error. This is caused by the fact that the routine is active: so, all the components are modified in the direction of the detected forces.\\
Obviously, we can't assume to have an error only along a single axis, so we can't modify only the axis where we know there is the error ($x$-axis in this case).\\
In truth, also the \mbox{$y$-component} is shifted, even if the change is very little because the component of the force acting along the length of the peg is neglected to not modify the wanted depth of insertion (as explained in \ref{sec:changeGoal}).\\

From the same figure \ref{fig:Error_nothingandgoal_plot6}, another thing to notice is that the routine does not correct perfectly the error in all the components. In fact the goal along the z-component is a bit erroneous at the end of the experiment. This happens because the peg has reached a point where the forces are zero, and so no more modifications can be done. This is caused by the fact that the peg has a smaller diameter than the hole, and so there is a tiny tolerance zone inside the cavity where collisions do not happen.\\

\subsection{Force-Torque Objective Discussion}
In general, the goal changes slowly: this means that, especially when the first contact happens, the forces and torques magnitudes are big as in the vanilla case.\\
The Force-Torque objective implemented helps to reduce these peaks: as soon as a force (or torque) is detected, the objective responds to it and generates a reference velocity that moves the peg away from the walls of the hole. The reduction of the magnitudes is visible in the upper plots of figure \ref{fig:comparison_final}. \\

In the lower plots of the same figure \ref{fig:comparison_final}, it is shown that the positional error between the goal and the tool is not affected so much by this objective. In fact this plot is similar to the one of the second experiment. In truth, the convergence of the norm to a small value is a bit slower than the one of the case where this objective is not used.\\
The first factor for this is that now the collisions are considered by the kinematic layer. So, being the new objective at an higher priority respect to the reaching goal objective, the kinematic control tends \textit{first} to nullify the forces and torques and \textit{then} to drive the tool toward the goal. This, sometimes, could increase the time to accomplish the mission (as in this case). Anyway, some other times it can \textit{decrease} the mission time because less stalemates happen, thanks to the fact that this objective may help to drive the peg away from the collision. In the second plot of figure \ref{fig:comparison_final}, we can see that the norm of the force stays at a nearly constant value around the interval $80s-180s$, as well as the norm of the error in the plot below. This shows a situation of stalemate that is then solved. Presence of standoffs like this one are discussed later.\\
A second factor is simply that, being the particularity of the problem, each experiment is different from the others, so sometimes the mission is \enquote{lucky} and it has less difficulties.\\

Figure \ref{fig:forceTaskActRef} shows how the new introduced objective generates references and activations in correspondence of the forces and the torques detected. Thus, obviously, the references and activations shapes (lower four plots) are similar to the forces and torques shape (upper two plots). In this case they are even more similar because the norm (which is the quantity that the objective controls) is given by almost only one component, the $y$ one.\\ 
The activation is a function which assumes only values from $0$ to $1$, so it is always positive. The reference instead, has opposite sign respect to the forces and the torques. This is because the objective wants to \textit{nullify} the forces and the torques, so it provides a velocity in the opposite direction.\\ 
In the reported plots reference and activation are the ones calculated by the \mbox{robot A}. However, the agents receive the same data from the sensor (except small synchronization problems that can happen), so the reference $\dot{\bar{\boldsymbol{x}}}_{ft}$ and the activation 	$\boldsymbol{A}_{ft}$ are the same for the robot B.\\

Figure \ref{fig:forceTaskVelocities} shows the results of $\; \boldsymbol{J}^{\#}_{ft} \; \dot{\bar{\boldsymbol{x}}}_{ft}\;$, for both robots. These are the system velocities that we would give if the Force Torque objective was the only one in the TPIK list, without considering the activation, and a simple pseudoinverse for the Jacobian was used (without any regularization). So, neither the collision propagation, the firm grasp constraint, nor the cooperation are included.\\
As explained before, the reference $\dot{\bar{\boldsymbol{x}}}_{ft}$ is the same vector for both agents. The thing that makes the two velocities so dissimilar, is the Jacobian $\boldsymbol{J}_{ft}\,$, which is obviously different for each agent because they are not in the same configuration. This difference, at this point, it is not a problem for the cooperation because we have still to deal with the coordination policy.\\
 
The plots of figure \ref{fig:forceTaskVelocities} are shown only to point out that the velocities are not so big to be not feasible. In truth, at least for the vehicle part, they could be even too small to be followed well. This can be solved making the objective to control only the arm, or by increasing the gains (but taking into account that bigger gains would mean greater risks of bad behaviours).\\
Another thing we can see is that they are not so smooth, and they have a lot of fast changes. However, firstly we have to consider that the activation, which the main work is smoothing the behaviour, is not present in the $\; \boldsymbol{J}^{\#}_{ft} \; \dot{\bar{\boldsymbol{x}}}_{ft}\;$ formula plotted.\\
Then, magnitudes are so little (in the $0.01rad/s$ order for joints, and in the $0.01m/s$ and $0.01rad/s$ order for vehicle) that these fast changes are not so important.\\
Last, we can't do so much because these changes are given by external data (the forces and the torques) that we have to deal with.\\

\subsection{Discussion About the Experiment with the Vision Part}
The final test of section \ref{sec:finalTest} (which includes the Vision part) is interesting for two main reasons. The first one is that the peg does not start so near the hole as in the previous trials (but it is almost aligned to it as before). So, we can see that the transportation phase is done in a good manner by the cooperative robots. This is noticeable even if no difficulties (e.g. difficult trajectories, an obstacle, a joint limit, and no vehicle and arm dynamics) are encountered.\\
The second reason is that the hole's pose has also some error on the angular components. This affect the orientation of the goal frame where the peg is driven to. Anyway, this error is small and it does not influence too much the insertion. \\

In fact, as we can see in the upper left plot of figure \ref{fig:expWithVisio}, the norm of the force and the norm of the torque are smaller than the ones of the previous experiment. Further, the error converge smoothly (upper right plot); the bigger error in the angular norm is obviously due to the fact that now there are also some imprecisions in the orientation of the hole. Also, little modifications are done to the frame goal, because its origin is almost with no error (lower plots). \\

Figure \ref{fig:expWithVisioVel} shows some plots about the cooperation. No big difficulties are met by the robots during the transportation and the insertion. So, the system velocities, that are the outputs of the final TPIK procedure (after the cooperation policy and the arm-vehicle coordination), are similar between the two robots. It must be remembered that these velocities are not the applied ones because the disturbances caused by collisions and firm grasp are not included. \\
However, for the robot B (right plot) we can see that tiny peaks are present around the interval $20s - 40s$. These can be caused \textit{indirectly} by the firm grasp constraint. When the tool of the robot B is driven away a bit, in the next control loop the kinematic layer must do a little more effort in recovering the trajectory, causing these tiny peaks. Being the cooperation policy included, if these peaks are high (as around second $8$) the robot A helps the other one, in fact a little peak is present also for the latter.\\

The last figure \ref{fig:expWithVisioVelTool} shows the \textit{feasible} \textit{cooperative} tool velocity $\dot{\tilde{\boldsymbol{x}}}_t$, that is the one that the coordinator sends to the two agents, after assuring that is achievable by both (as explained in section \ref{sec:coopScheme}).\\
The \textit{non-cooperative} velocities $\dot{\boldsymbol{x}}_{t,a}$ and $\dot{\boldsymbol{x}}_{t,b}\,$ are not shown because they are very similar to the cooperative one. This is caused by the fact that no robot has difficulties in providing the \textit{ideal} tool velocity.\\
Even if disturbances of collisions and of firm grasp constraint are not present, this plot gives an idea about the tool's speed during the mission. The object is driven slowly because the mission needs so: the insertion phase, made by two kinematic cooperative manipulators, puts big challenges and we can't afford to have too big gains.\\

\subsection{Standoff Discussion}
A last thing to report is that some experiments meet a standoff situation at some point during the insertion. This happens when the collisions continuously make the peg \enquote{bounce} on the inner hole walls, making it moving back and forth. The \enquote{bounce} is due to a bad peg alignment inside the hole, so the tip continuously touches the inner wall.\\

In these cases, sometimes, the methods used do not manage to solve the stalemate in a reasonable time. The problem can be due to different factors.\\
One could be that the simulation, being only kinematic, is not precise. So, even if the gains are really small, in any case the velocities are instantaneously provided to the system, which causes always a bit of chattering. This complicates the work of the simulator (to calculate the collisions) and of the whole mission (that could meet almost instantaneously a strong force or torque).\\
Another problem could be a not so good setting of the many gains that we have to put.\\ Other one can be given by how Collision Propagation work. Collisions propagate on the system through Jacobians, that are mappings derived from linear approximations of non-linear things.\\ 
Other one can be that, with the Change Goal routine, orientation's modifications are not made, so no corrections are doable for the angular part.\\
Additional problem to take into account is that some synchronization issues may happen. The coordination policy, in the way it is implemented, assures the synchronization (section \ref{sec:controlLoop}), but we have also other kinds of exchanged information. For example, the force-torque sensor data is shared using ROS topics in a simple way, so with the actual software we can't know exactly when information arrives to each node. Considering also that on a single machine we run the simulation and the two robot software (plus the coordinator) this could cause even more synchronization problems. This may be solved with further code improvement, but it would go out of the scope of this work and, also, it would not be so useful for the real application (where we don't run everything on a single machine).\\

The standoff problem is more influential when the hole's pose error is bigger, or, for example, when we have also orientation error such as in the last experiment. However, the methods show a good starting point to improve the current state of the art in this unexplored problem underwater.

\chapter{Vision: Methods \& Results}

\begin{figure}[H]
	\centering
	\includegraphics[width=13.5cm]{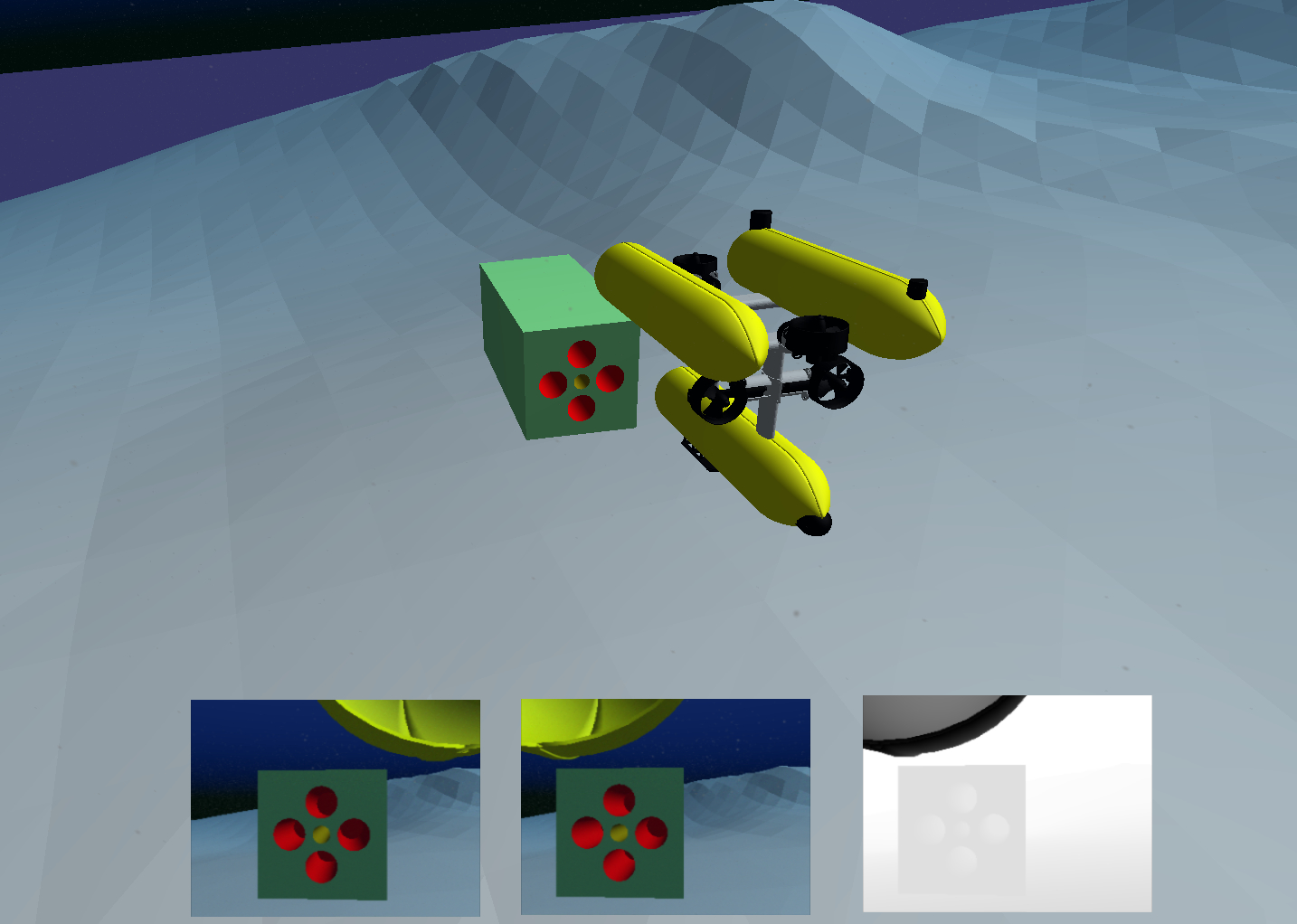}
	\caption[The Vision robot watching the hole]{The \textit{vision} Robot watching the hole. The hole (in yellow) is in the centre of the cuboid. The red holes around are there to help vision algorithms. Below, the right RGB, the left RGB and the left depth cameras are shown. In the methods, only one of the left cameras is used.}
	\label{fig:vision-uwsim}
\end{figure}

This Chapter covers exclusively the Vision part. So, here they are presented methods and tools to estimate the hole's position with computer vision algorithms. To not go outside the scope of this thesis, the methods are only introduced, briefly explained, and compared; no theoretical background is given. So, no mathematical formulas are illustrated for the used vision functions. \\

Before the two carrying robots can approach the hole, obviously the hole's position must be know, at least roughly.
In the considered scenario, a third robot is present, as depicted in figure \ref{fig:vision-uwsim}. Its work is devoted exclusively \textit{to detect} and \textit{to track} the hole. In the simulation, another \href{https://cirs.udg.edu/auvs-technology/auvs/girona-500-auv/}{Girona 500 AUV} is used for this job, without the arm. It is evident that, in a real scenario, a smaller and more efficient robot should be used for the vision, seeing that no manipulation capability are needed. In fact, in the original TWINBOT [\cite{TWINBOT2019}] simulation, a smaller \href{https://bluerobotics.com/product-category/rov/bluerov2/}{BlueROV} is present.
Anyway, for this thesis, another Girona 500 is used to not deal with an additional robot model.\\

\begin{figure}[H]
	\centering
	\includegraphics[width=12.5cm]{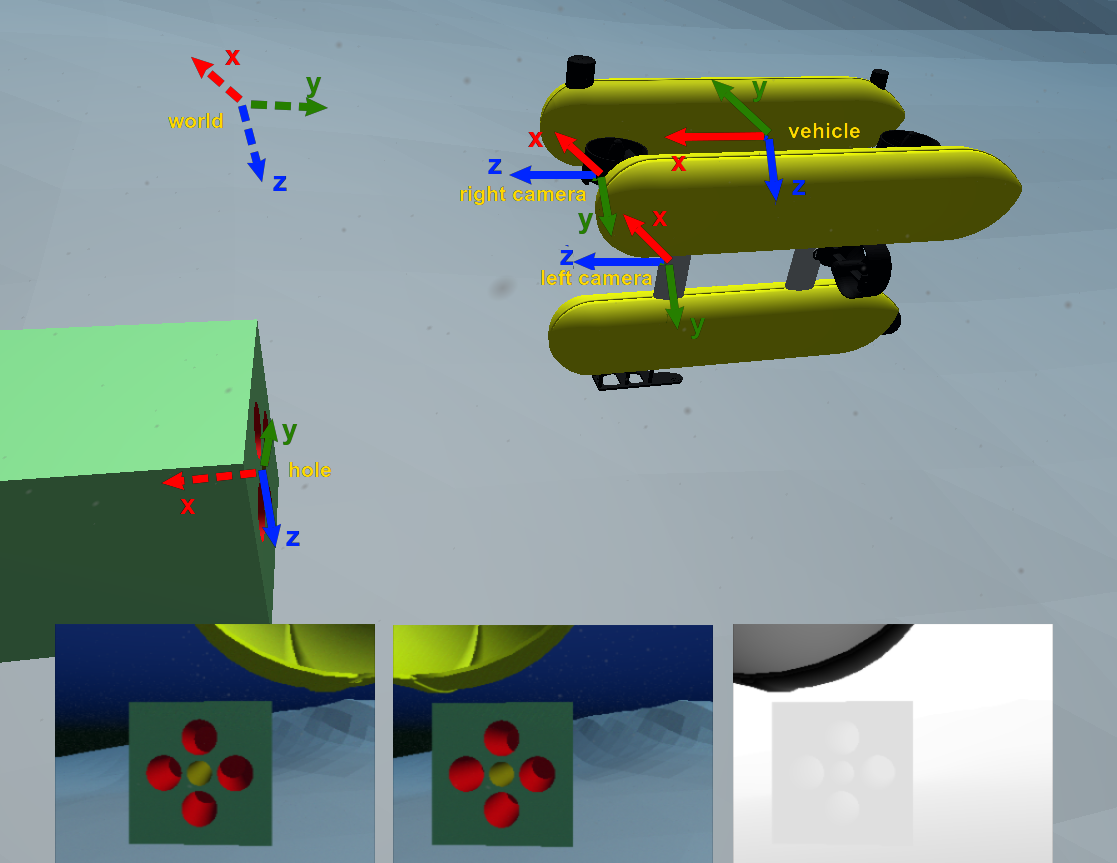}
	\caption[Main frames related to the Vision phase]{Screenshot where the main frames for the Vision are depicted. The cameras 3D models are not present. The left and the right cameras have same orientation, and the origin of the right one is shifted of 0.2 meters along left camera's $x$-axis. Vehicle and cameras axis are parallel. Hole's $x$-axis goes inside the cavity. World's frame is depicted only to understand its orientation, its origin is not in that point.}
	\label{fig:visionFrames}
\end{figure}

The Vision robot is equipped with two cameras which point in front of it, as shown in figure \ref{fig:visionFrames}. They are used for three different methods as:
\begin{itemize}
	\item Two distinct cameras, independent of each other (mono camera case).
	\item As stereo cameras, thus exploiting stereo vision algorithms.
	\item As RGB-D camera, i.e., a stereo vision couple where the left one is a RGB camera and the right one is a Depth camera.  
\end{itemize} 

\noindent The job is divided in two phases: \textit{Detection}  of the hole (section \ref{sec:visDetect}) and \textit{Tracking} of the hole's pose \mbox{(section \ref{sec:visTracking}).} 

\section{Vision Assumptions}
\label{sec:visioAssumption} %
For the sake of simplicity, some assumptions are made:
\begin{itemize}
	\item Known \textit{intrinsic} camera parameters. These parameters are used by algorithms to take into account how the single camera sees the scene. No image distortion is present.
	\item Known \textit{extrinsic} camera parameters, i.e. the position and the orientation of cameras (respect to the vehicle), and thus it is also known the relative pose between the two cameras (needed for stereo vision algorithm).
	\item No external disturbances for Vision, such as underwater light reflections or bad visibility.
	\item Hole's model known. This means that dimensions of the cuboid which contains the hole are known. Further explanation about this are given successively in section \ref{sec:visTracking}.
	\item A "friendly" cuboid structure of the hole. The front face is coloured and additional holes are present, as can be seen in fig. \ref{fig:vision-uwsim}. This helps both the \textit{detection} phase and the \textit{tracking} phase.
\end{itemize}
About the robot, other assumption are:
\begin{itemize}
	\item The pose of the vehicle respect to the inertial frame (the \textit{world}) is known. This is important to share the hole's pose with the carrying agents. Further details about this are given in section \ref{sec:expAssumption}.
	
	\item The initial position of the robot is such that it is facing the front face of the hole. It must be noticed that methods explained in the next sections can be adapted to relax this hypothesis. For example, the vehicle could turn around z-axis until the hole is detected with one of the methods described later. 
	
	\item Once the robot has tracked the hole and the pose is sent to the twin robots, it must go away to not interfere with the insertion phase. This is done through a keyboard (as a ROV) but it is not difficult to improve the code to let it go away autonomously. It also must be noticed that, thanks to the tracking, if the robot moves (because it is commanded to do so, or for water currents) the pose estimation keeps working. 
\end{itemize}

\section{Tools}
\label{sec:visionTools}
To deal with the pose estimation, some external libraries are used. In this section they are listed.

\begin{itemize}

	\item \href{https://opencv.org/}{\textbf{OpenCV}} (Open Source Computer Vision Library) [\cite{opencv}],  an open-source BSD-licensed library that includes several hundreds of computer vision algorithms. In this thesis, it is used mostly for the detection part, even if some of its functionalities are used also by ViSP (for example for the keypoint tracking).
	
	\item \href{https://visp.inria.fr/}{\textbf{ViSP}} (Visual Servoing Platform) [\cite{visp}], another open source library that helps to develop applications which exploit visual tracking and visual servoing techniques. It is interesting because it is more specific than OpenCV for robotic fields. In this work, it is used for the tracking phase.
	
	\item \href{http://www.pointclouds.org/}{\textbf{PCL}} (Point Cloud Library) [\cite{pclLib}] a library for 2D/3D image and point cloud processing. In this work, it is used by ViSP when depth images are used. However, further works can use it as another tool to deal with the vision part.
	
\end{itemize}

\section{Detection}
\label{sec:visDetect}

\textit{Object Detection} means detecting a particular shape (i.e. the \textit{object}) in the scene. This is important to initialize the common tracking algorithms known in the literature.\\

In fact, for the method exploited in this work, the detection step must provide a correspondence between some pixels in the 2D image and some points of the 3D object. It is important to notice that the needed 3D coordinates refer to the object frame (the \textit{hole} in figure \ref{fig:visionFrames}), and not to an "external" frame. Seen that the object model is assumed to be know, the knowledge of some 3D coordinates directly derives from this assumption.\\

Four is the minimum number of points accepted by the tracking algorithm. The more the points are, the more the tracking is good. Plus, points should lie on different surfaces of the object, to have better results. \\
Anyway, in this case, good tracking results are obtained also not considering these two aspects. The four points chosen are the corners of the front face of the cuboid which contains the hole.\\
The 3D coordinates chosen for the simulations are represented in the \textit{.init} file \ref{file:initfile}.
\begin{fileAlgorithm}
	\caption{\textit{The \emph{.init} file which describes the position on the 4 corners of the front face, respect to a frame positioned in the centre of the hole, with x-axis going inside the hole, y lying along the surface pointing on the right, z pointing down to the seafloor (this hole frame is depicted in figure \ref{fig:visionFrames}).}}
	\label{file:initfile}
	\begin{algorithmic}[1]
	\STATE 4         \hspace*{50px}    \# Number of points\\
	          \hspace*{59px}        \# Coordinate order is \textit{x y z}. The unit of measure is the meter\\
	\STATE 0      -0.4     -0.4  \hspace*{5px} \# top right corner
	\STATE 0      0.4      -0.4  \hspace*{8px}   \# top left
	\STATE 0      0.4     0.4   \hspace*{11px} \# bottom left
	\STATE 0      -0.4    0.4    \hspace*{7px} \# bottom right
	\end{algorithmic}
\end{fileAlgorithm}
\vspace{30px}

\noindent The work of the Detection step is to provide pixels' 2D coordinates that correspond to the 3D points of file \ref{file:initfile}. This must be done for each camera, except for the depth one (when used).\\
To provide these correspondences, two methods are evaluated: \textit{Find Square} (section \ref{subsec:findSquare}) and \textit{Template Matching} (section \ref{subsec:templateMatch}). A third method (section \ref{subsec:clickMethod}), where the 2D coordinates are precise as much as possible (i.e. they are selected by hand clicking on the exact pixels), is used as a benchmark. This is also helpful to analyse the tracking results when the 2D coordinates are almost perfect.\\

Other methods and functions for the detection are briefly explained in \mbox{Appendix \ref{chap:AppendixVision}.}\\
Another, not explored, method can be using some code tags (like QR codes) on the cuboid surface. However, in underwater situations this can be difficult to be put in practice.

\subsection{Already Known Coordinates Method}
\label{subsec:clickMethod}
As explained, with this method the 2D coordinates are perfectly known. This is done by letting the user to click on the four pixels which contain the square's corners. Given that the image is made by discrete pixels, it is impossible to have an ideal point which is exactly the corner, but the errors for this are not noticeable.

\subsection{Find Square Method}
\label{subsec:findSquare}
This method is taken from an OpenCV tutorial (\url{https://docs.opencv.org/3.4/db/d00/samples_2cpp_2squares_8cpp-example.html}).\\
A rough explanation of how the method works is presented:
\begin{itemize}
	\item This method looks in each image channel (that is only one if it is a black and white image, or they are three if it is a coloured image) to find squares.
	\item First, it pre-processes the image to reduce noise, using a pyramid scaling. Then, it exploits the Canny Edge Detector [\cite{CannyEdge}] to highlight the edges (results of Canny are visible in figure \ref{fig:HoughStandard} of Appendix's section \ref{sec:HoughTrasf}).
	\item The OpenCV function \href{https://docs.opencv.org/3.4.6/d3/dc0/group__imgproc__shape.html#ga17ed9f5d79ae97bd4c7cf18403e1689a}{\textit{findContours()}} is called to extract contours of shapes with the algorithm described in \cite{findcountors}. The output after this passage is visible in figure \ref{fig:BoundBoxresultOnlyPolig} of Appendix section \ref{sec:boundingBox}.
	\item Each shape's contour is approximated to be more like a regular polygon, i.e. with less vertices and edges.
	\item Finally, the algorithm looks if the shapes founded are similar to a square or to a rectangle. This is done checking if the internal angles of the contours are approximately 90 degrees. Furthermore, shapes with too little area are discarded to eliminate noise.
	
	\item The returned shapes are described by their four corners, that is what we were looking for.

\end{itemize}

An additional function is called to be sure that the order of the returned corners is the same order of the 3D points, otherwise correspondences are obviously erroneous.

\subsection{Template Matching Method}
\label{subsec:templateMatch}
\textit{Template Matching} means to find a pattern (in this case, the face of the hole) inside a scene. This method is a well-known tool used in many applications.\\

It is important to notice that an additional image (the \textit{template}) is necessary. So, we have to assume that an image of the hole's square face is provided.\\

The code implemented follows an OpenCV tutorial (\url{https://docs.opencv.org/3.4.6/de/da9/tutorial\_template\_matching.html}).\\
In brief, the \textit{template matching} finds the point in the scene which has the best similarity (or the least dissimilarity) with the provided template. This is done considering intensity values of the pixels in the neighbourhood area of each point in the image.\\
In practice, the template is shifted all over the scene and a formula is computed for each shifting. Various formulas to compute similarity (or dissimilarity) are provided by OpenCV and are detailed in the library \href{https://docs.opencv.org/master/df/dfb/group__imgproc__object.html#gga3a7850640f1fe1f58fe91a2d7583695dab65c042ed62c9e9e095a1e7e41fe2773}{documentation}. The chosen one in the experiments is the so-called \textit{squared difference}.\\

To have correct results, it is important to scale up and down the template and to compute multiple times the similarity. This because usually the template's size is not equal to the size of the object in the scene.\\
So, for each scaling, a best similarity point is detected. Then, all the similarity points are compared and the best one is taken. At the end, a rectangle with the template (scaled) dimensions is built considering the best point as the centre. The corners of this rectangle are the 4 points which we were looking for.

\subsection{Detection Results}
\label{subsec:detectResult}
\begin{figure}[H]
	\centering
	
	\textbf{Already known coordinates Method}\\
	\includegraphics[width=4.5cm]{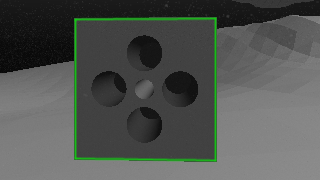}
	\includegraphics[width=4.5cm]{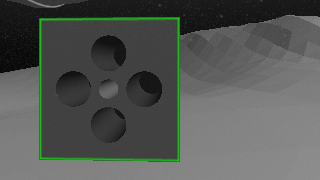}\\
	\vspace{30px}
	
	\textbf{Find Square Method}\\
	\includegraphics[width=4.5cm]{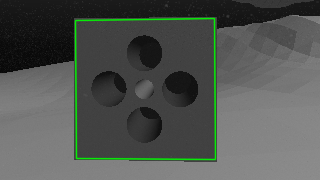}
	\includegraphics[width=4.5cm]{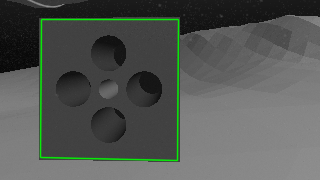}\\
	\vspace{30px}
		
	\textbf{Template Matching Method}\\
	\includegraphics[width=4.5cm]{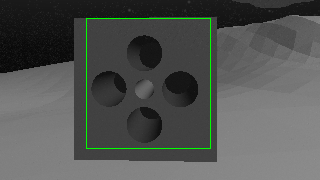}
	\includegraphics[width=4.5cm]{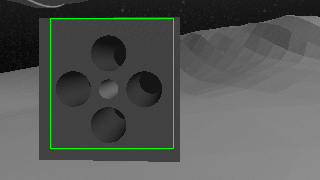}
	
	\caption[Hole Detection results with the three different methods]{Results of the three different detection method. The green rectangle is the estimate position of the square. The output of the detection step are the four corners of the green rectangle.}
	\label{fig:detectResults}
\end{figure}

For this scenario, the Find Square method is better than the Template Matching one.\\
As can be seen in figure \ref{fig:detectResults}, differences from the ideal method and the Find Square one are barely visible.\\

Looking at the way it works, it should be noticed that the Find Square method gives good results only if the camera faces the cuboid structure approximately at the front. If a side face is more visible, it will be the one where the rectangle is detected. This is not a problem because the tracking algorithm can be initialized also by a side face, but we have to give the 3D points which correspond to this side. So we must know which face the robot is looking at.\\

In addition, this method is suitable if no other squares of similar dimensions are present in the visible scene. If this would happen, some further works are needed to take the right one.\\
 
Another problem is that it is not suitable with other kind of shapes (a circle structure, for example). This is obvious because the algorithm only detects squares and rectangles.\\ 

It is also important to point out that, sometimes, the method fails to find any shapes in the right image, when the initial robot position is the one chosen in the experiment. This happens approximately 30\% of the time, and it may show a very bad robustness and a low predictability of the method. However, the fail is detectable (by a human operator but also easily by the software) and another trial can be repeated.\\

As said, the Template Matching is less precise than the first method.\\
Apart this, it has different problems than previous. Firstly, if the face is viewed from a different angle, an other template image is needed, with an orientation similar to what the robot is seeing. In general, lot of template images at different angles are needed. Otherwise, some processing of the template image is necessary to orient it in a different way.\\

Secondly, the orientation can make the face to be a distorted square in the 2D image. With the first method, contours are followed well even if the quadrilateral is not a perfect square (until a certain point obviously). With the Template Matching, the only way to find the corners is by building a shape equal to the template around the similarity point. So, the real edges can be followed badly and the resultant corners can be imprecise (as can be seen in figure \ref{fig:detectResults}).\\

A better quality of the Template Matching method respect to the other is that it can be used for different shapes, not only rectangle ones. Plus, the object to be found can be also a complicated one, full of details that would put a \enquote{geometric shapes detector} (like the Find Square) in big difficulties.\\

\section{Tracking}
\label{sec:visTracking}
\textit{Object Tracking} means to follow the motion of an object of interest during time. Both the object and the camera can be mobile, even if, in this case, only the cameras move (actually, the robot moves with the cameras rigidly attached to its body). Tracking an object is usually done to estimate its pose respect to the camera frame.\\

In this work, we speak about a \textit{markerless} \textit{model-based} tracking. Thus, no markers attached to the object are needed but the object structure must be known. In this scenario, it is sufficient to give to the algorithm the 3D dimensions of the cuboid structure of the hole (i.e., the position of the eight corners respect to the object frame), and the hole's dimension and position in the front face (i.e. position of three points which describe the circumference respect to the object frame).\\
The library chosen for the tracking phase, \href{https://visp.inria.fr/}{ViSP} [\cite{visp}], uses an own format called \textit{.cao}, which syntax is described in the library  \href{https://visp-doc.inria.fr/doxygen/visp-daily/tutorial-tracking-mb-generic.html#mb_generic_advanced_cao}{documentation}.\\
As explained in section \ref{sec:visDetect}, the tracking algorithm must also know the 2D-3D correspondence of \textit{at least} four points belonging to the object. In the experiments, the provided ones are the 4 corners of the front face.\\

Three different trackers have been implemented: a \textit{Mono Cameras Tracker} (section \ref{subsec:monoTrack}), a \textit{Stereo Camera Tracker} (section \ref{subsec:stereoTrack}), and a \textit{Stereo Depth Camera Tracker} (section \ref{subsec:depthTrack}). Results and comparison of them are discussed in \mbox{section \ref{subsec:trackResult}}.\\

A tracker is linked to each camera. For RGB cameras, it can be of three types: \textit{edge-based} [\cite{visp-edge}], \textit{keypoint-based} [\cite{visp-klt}] or a mix of both. During the experiments, the hybrid method emerged as the most precise, so all the results in section \ref{subsec:trackResult} refer to this one.\\
For the depth camera used in the Stereo Depth Camera tracking, the tracker's type can be \textit{normal} or \textit{dense} [\cite{visp-depth}]. Being the \textit{dense} one more robust, it is the chosen one for the trials. Please note that it is also computationally heavier for larger matrix computations, but speed performance is not considered here.

\subsection{Mono Camera Tracking}
\label{subsec:monoTrack}
This method derives from the ViSP tutorial \textit{markerless generic model-based tracking using a colour camera}  (\url{https://visp-doc.inria.fr/doxygen/visp-daily/tutorial-tracking-mb-generic.html}).\\

The implementation is straightforward: after setting the trackers (i.e. giving edge detection and keypoint detection parameters, camera parameters, and 2D-3D correspondences of the four corners), at each loop the tracker estimates the transformation matrix between each camera and the object.\\

In this method, the left and the right cameras are independent. Thus, each one provides a different pose estimation. It is not so easy to understand when one camera provides better results that the other, without taking the real pose as benchmark. So, in the applications could be difficult to choose one pose or the other. A good compromise could be to do a mean of them.\\

\subsection{Stereo Camera Tracking}
\label{subsec:stereoTrack}
This method derives from the ViSP tutorial \textit{Markerless generic model-based tracking using a stereo camera} (\url{https://visp-doc.inria.fr/doxygen/visp-daily/tutorial-tracking-mb-generic-stereo.html}).\\

The code is analogous to the previous one, except that in this case also the relative pose between each camera must be provided. If this is unknown, some method for stereo calibration must be used. Due to the fact that now the cameras are not independent, a unique pose is provided, that, as we will see later, has better precision than the previous one.

\subsection{Stereo Depth Camera Tracking}
\label{subsec:depthTrack}
This method derived from the ViSP tutorial \textit{Markerless generic model-based tracking using a RGB-D camera} (\url{https://visp-doc.inria.fr/doxygen/visp-daily/tutorial-tracking-mb-generic-rgbd.html}).\\

This method is similar to the previous one, except that now the right camera is a depth one, thus it provides range images.\\
The functions used for the depth images need the support of another library, \href{http://www.pointclouds.org/}{PCL} [\cite{pclLib}].\\
Another difference is that the depth camera does not need to initialize the 2D-3D correspondences, so the \textit{Detection} step has to be done only for the left camera.

\subsection{Tracking Results}
\label{subsec:trackResult}

In this section, performances of the three trackers are evaluated. For each one, the three different types of detection initialization (explained in section \ref{sec:visDetect}) are considered to evaluate the effects of detection's error on each kind of tracker.\\

Experiments have been conducted with a lot of simplifications: no disturbances, no camera distortions, very good visibility, nice object shape. Results described here can give only an idea on how to proceed in a more realistic environment.\\

In the scenario, the robot is perfectly still while it is tracking the object, even if the tracking methods can be used with moving objects and/or moving cameras. The vehicle is positioned in front of the square face, slightly on the right. The original images taken from cameras are cut to delete a region where a part of the vehicle is visible. This is done to make this part to not cause useless disturbances to the vision algorithms. In the depth images, this is not necessary, because there is no interference.\\

In figure \ref{fig:photoTracking} the detected shape and the estimated frame are drawn on the camera images. Differences are barely visible when comparing the first two initialization methods (the ideal one and the Find Square one) in all the three tracking methods.\\

\begin{figure}
	\centering
	\textbf{Template Matching Initialization}\\
	\vspace*{20px}
	\centerline{
		\includegraphics[width=9cm]{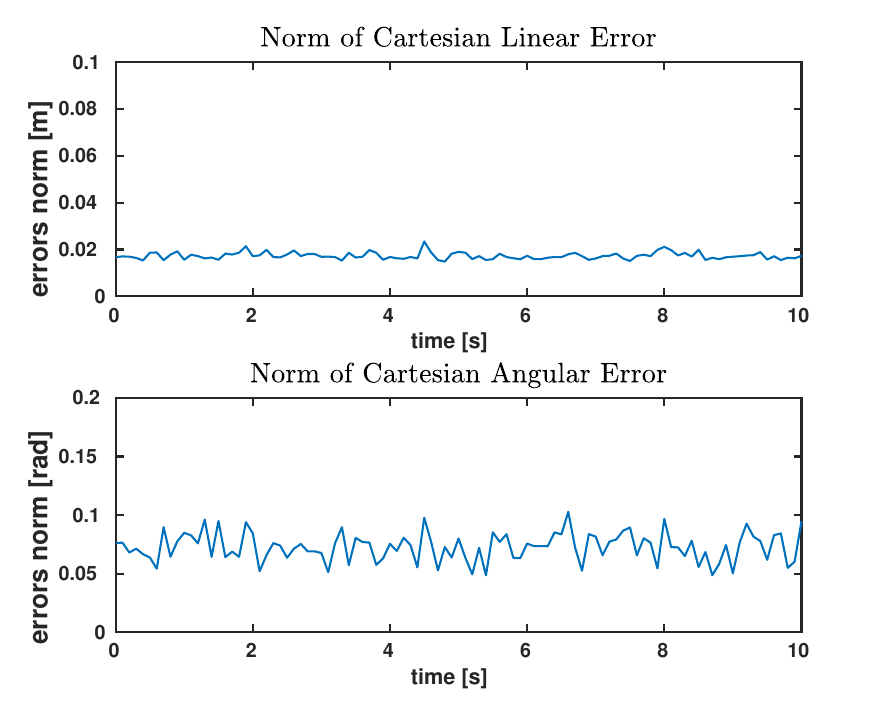}
		\includegraphics[width=9cm]{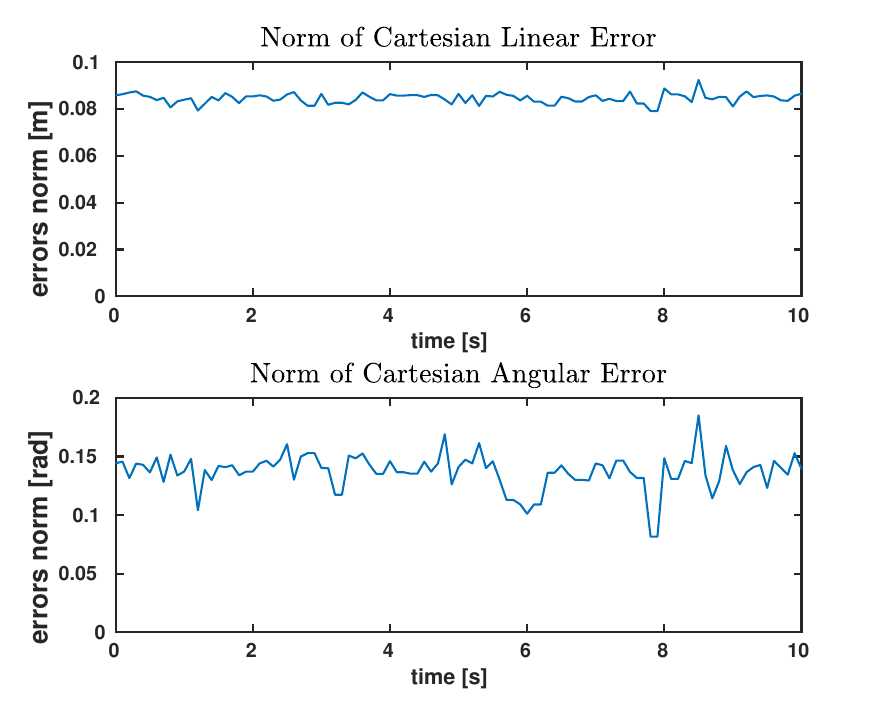}
	}
	\hspace*{15px}\textit{Mono (left Camera) Case} \hspace{125px} \textit{Mono (right Camera) Case}\\
	\vspace{30px}
	\centerline{
		\includegraphics[width=9cm]{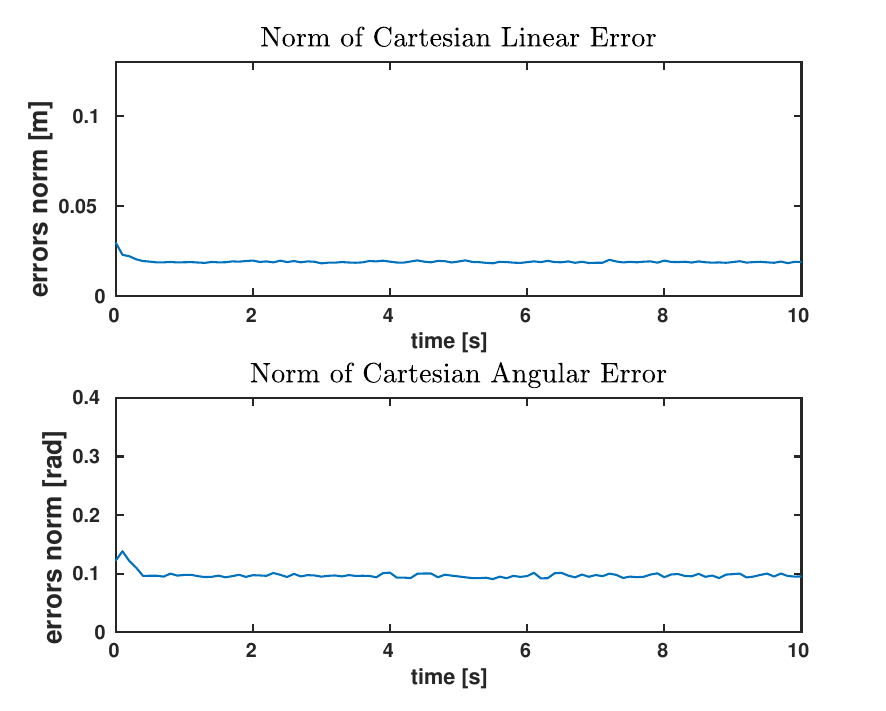}
		\includegraphics[width=9cm]{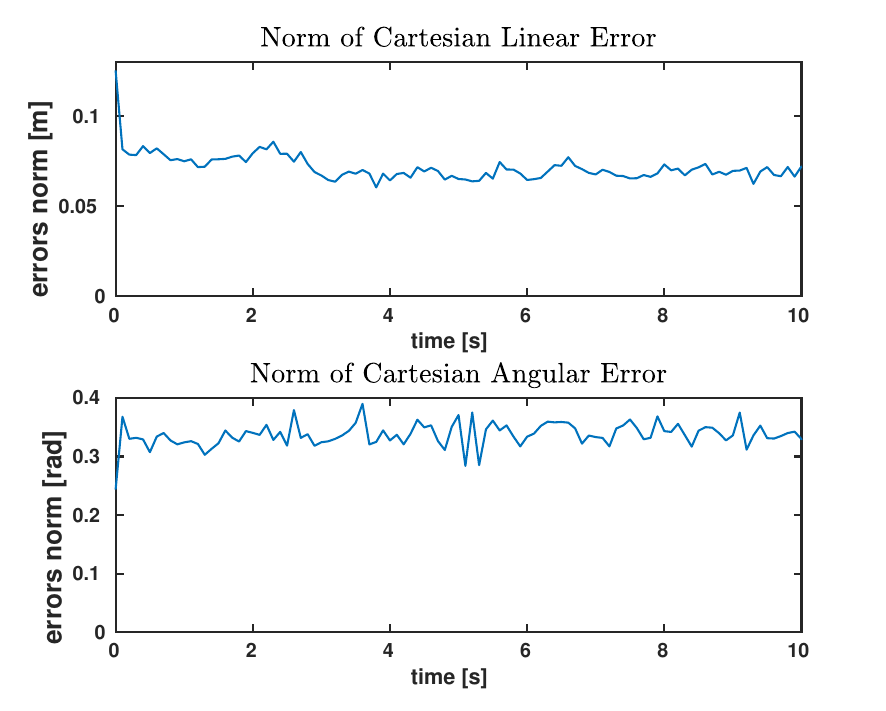}
	}
	\hspace*{20px}\textit{Stereo Camera Case} \hspace{135px} \textit{Stereo Depth Camera Case}\\
	\vspace{30px}
	\caption[Tracking error plots with Template Matching detection initialization]{Linear and angular error (in norm) between the true pose and the estimated one. The tracking is initialized with the Template Matching detection method of \mbox{section \ref{subsec:templateMatch}}.}
	\label{fig:templateErrors}
\end{figure}

Instead with the Template Matching detection initialization (last group of images of figure \ref{fig:photoTracking}), the lower precision (visible in figure \ref{fig:detectResults}) is paid in tracking results, especially in the depth case. This is clearer in figure \ref{fig:templateErrors} where the error is plotted. With this initialization, the depth-stereo method is even worse than the monocular case.\\
This can be due to the fact that the depth image is not initialized with 2D-3D correspondence; thus we pay more the initialization error, being done only in the left image.\\
This demonstrates that it is not always better to have a stereo RGB-D camera instead of a normal stereo RGB. This is an interesting result and should be further explored with more realistic scenes.\\

\begin{figure}[H]
	\begin{center}
		\textbf{Already known coordinates Initialization}
	\end{center}
	\vspace{-10px}	
   	\includegraphics[width=3.4cm]{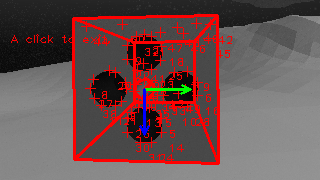}
	\includegraphics[width=3.4cm]{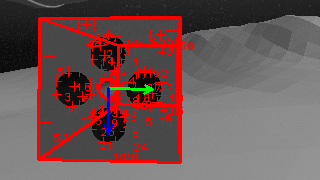}
	\hspace{10px}
	\includegraphics[width=3.4cm]{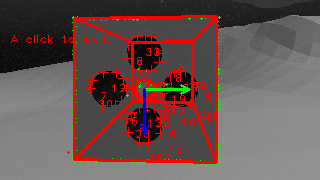}
	\includegraphics[width=3.4cm]{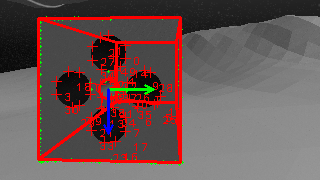}\\
	{\footnotesize \hspace*{20px}\textit{Mono Cameras Case} \hspace{120px} \textit{Stereo Cameras Case}}\\
	\centering{
		
	\includegraphics[width=3.4cm]{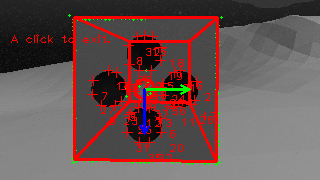}
	\includegraphics[width=3.4cm]{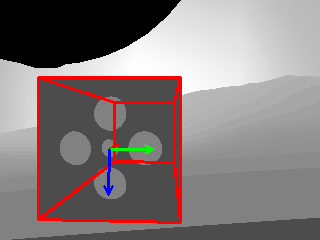}}\\
    {\footnotesize \textit{Stereo Depth Camera Case}}\\  
\end{figure}
\vspace{-12px}
\begin{figure}[H]
	\begin{center}
		 \textbf{Find Square Initialization}
	\end{center}
	\vspace{-10px}
	\includegraphics[width=3.4cm]{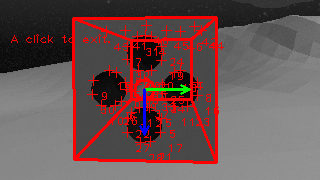}
	\includegraphics[width=3.4cm]{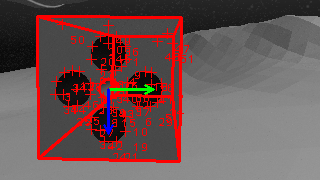}
	\hspace{10px}
	\includegraphics[width=3.4cm]{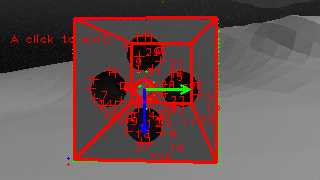}
	\includegraphics[width=3.4cm]{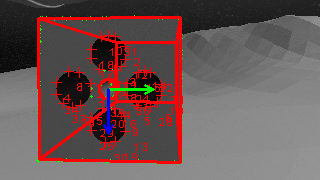}\\
	{\footnotesize \hspace*{20px}\textit{Mono Cameras Case} \hspace{120px} \textit{Stereo Cameras Case}}\\
	\centering{
		
	\includegraphics[width=3.4cm]{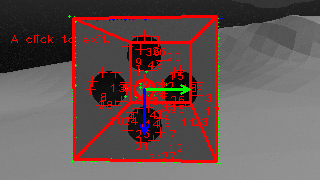}
	\includegraphics[width=3.4cm]{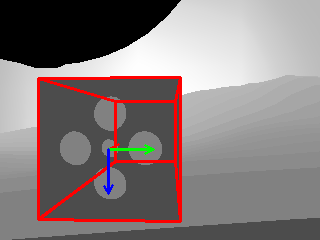}}\\
    {\footnotesize \textit{Stereo Depth Camera Case}}\\  
\end{figure}
\vspace{-12px}
\begin{figure}[H]
	\begin{center}
		\textbf{Template Matching Initialization}
	\end{center}
	\vspace{-10px}
	\includegraphics[width=3.4cm]{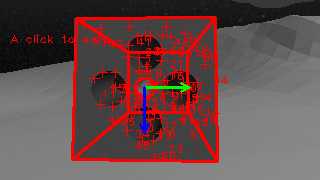}
	\includegraphics[width=3.4cm]{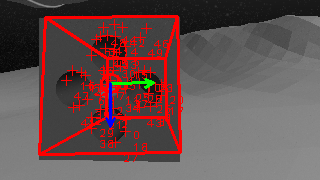}
	\hspace{10px}
	\includegraphics[width=3.4cm]{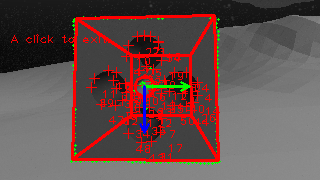}
	\includegraphics[width=3.4cm]{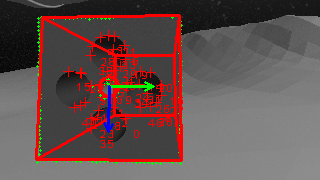}\\
	{\footnotesize \hspace*{20px}\textit{Mono Cameras Case} \hspace{120px} \textit{Stereo Cameras Case}}\\
	\centering{
		
	\includegraphics[width=3.4cm]{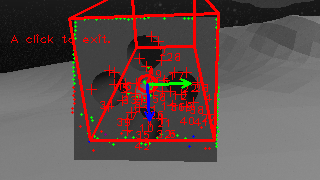}
	\includegraphics[width=3.4cm]{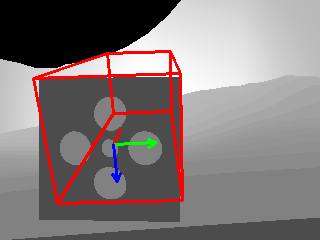}}\\
    {\footnotesize \textit{Stereo Depth Camera Case}}\\
\end{figure}
\begingroup
\captionof{figure}[Tracking results with the three different detection initializations]{Results of the three tracking methods with the three detection initializations. Red lines are the contours of the model which represents where the object is estimated to be; little red $+$ marks are the points detected by the keypoint algorithm. Green dots are the tracked correspondent points between the left and the right images, in fact they are present only in the stereo cases. The arrows represent the estimated object frame: green for y-axis, blue for z-axis, red for x-axis (which goes inside the hole and it is barely visible).}
\label{fig:photoTracking}
\endgroup
\vspace{35px}

\begin{figure}
	\centering
	\textbf{Already known coordinates Initialization}\\
	\vspace*{20px}
	\centerline{
		\includegraphics[width=9cm]{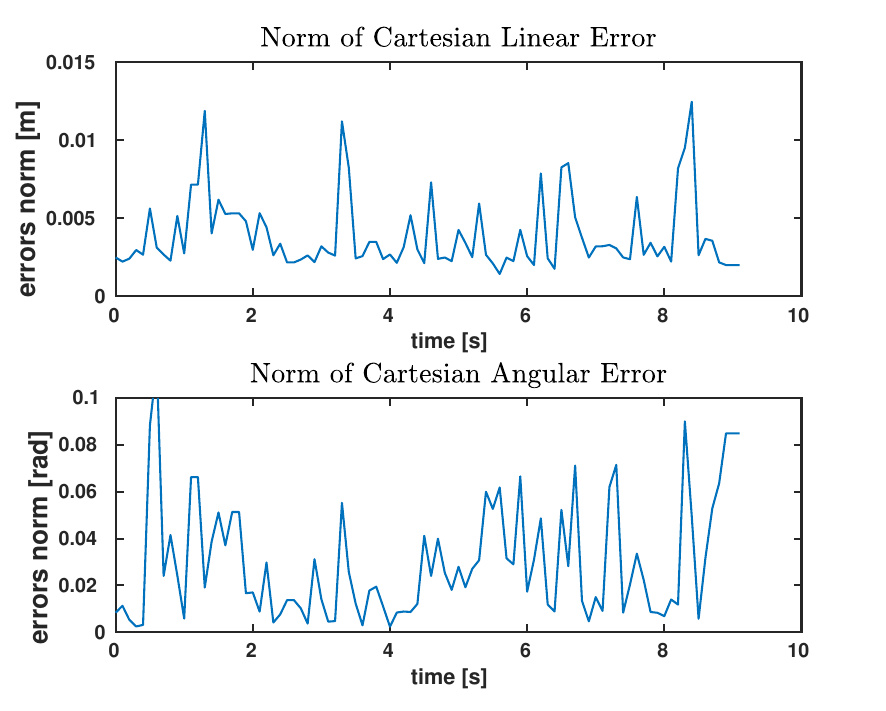}
		\includegraphics[width=9cm]{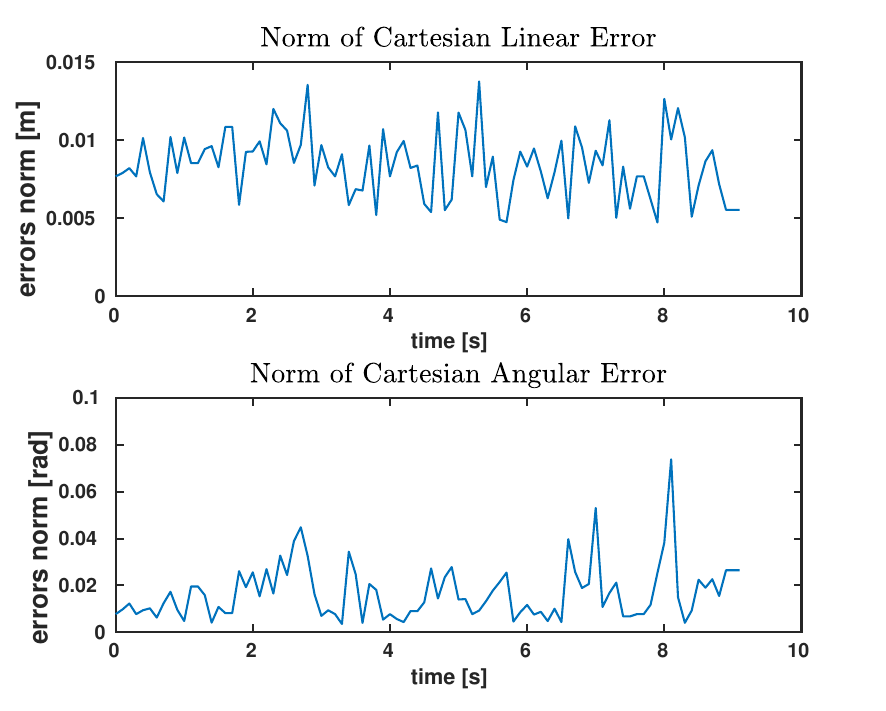}
	}
	\hspace*{15px}\textit{Mono (left Camera) Case} \hspace{125px} \textit{Mono (right Camera) Case}\\
	\vspace{30px}
	\centerline{
		\includegraphics[width=9cm]{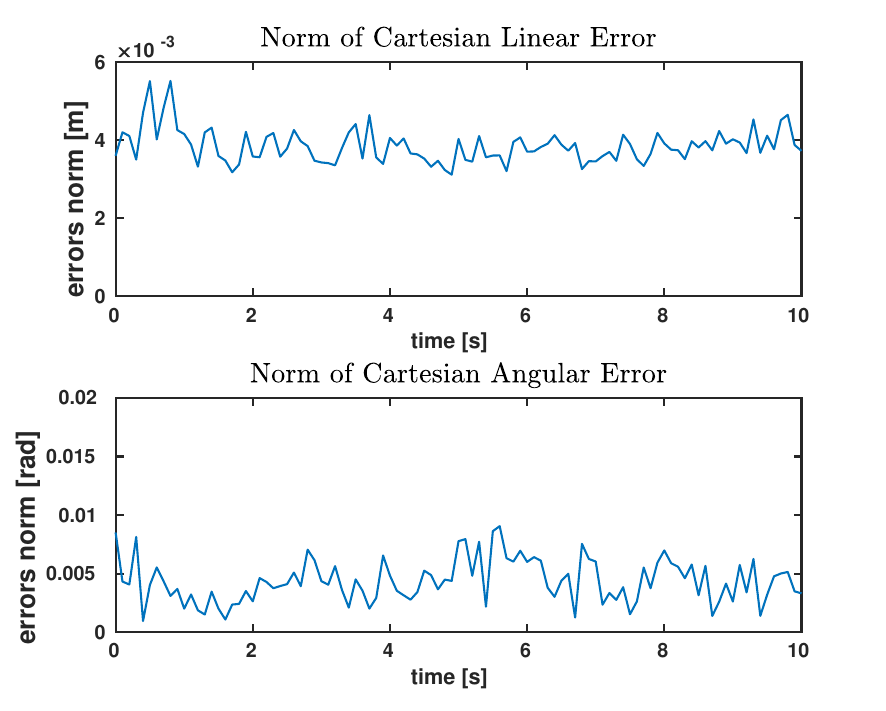}
		\includegraphics[width=9cm]{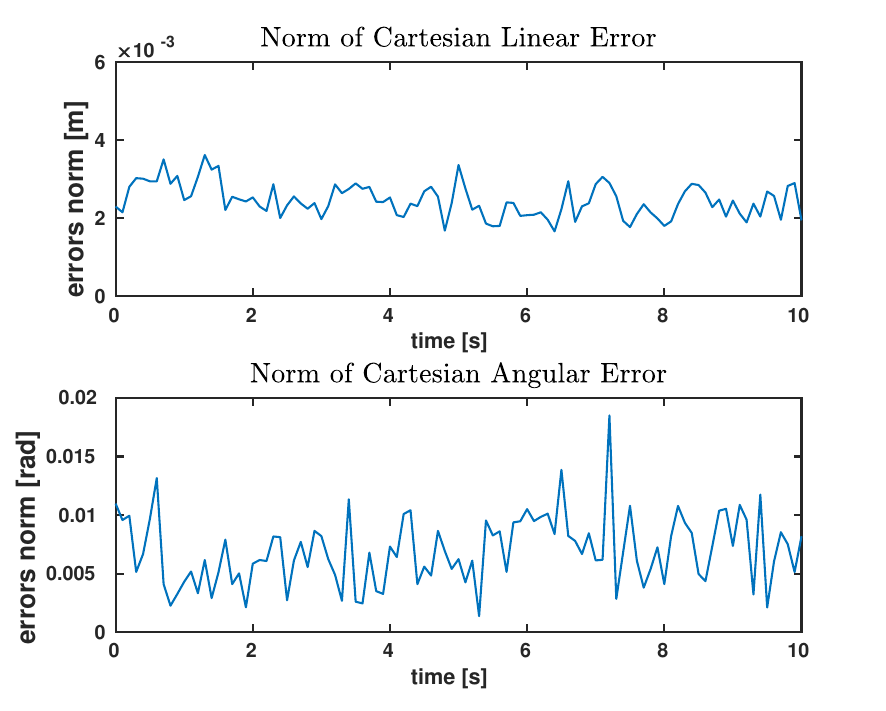}
	}
	\hspace*{20px}\textit{Stereo Camera Case} \hspace{135px} \textit{Stereo Depth Camera Case}\\
	\vspace{30px}
	\caption[Tracking error plots with ideal detection initialization]{Linear and angular error (in norm) between the true pose and the estimated one. The tracking is initialized with the ideal detection method of section \ref{subsec:clickMethod}.}
	\label{fig:clickErrors}
\end{figure}

\begin{figure}
	\centering
	\textbf{Find Square Initialization}\\
	\vspace*{20px}
	\centerline{
		\includegraphics[width=9cm]{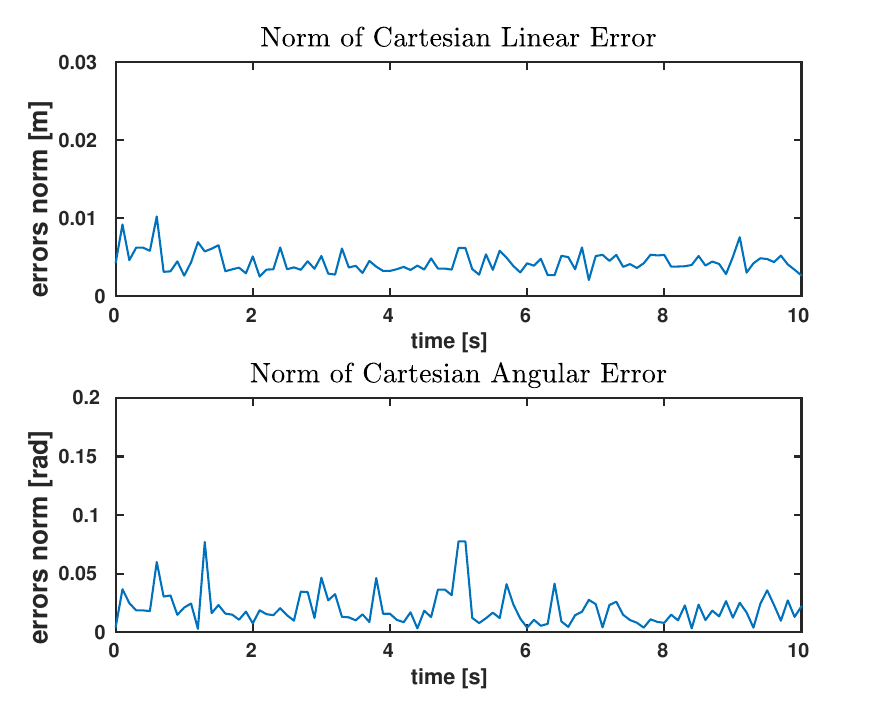}
		\includegraphics[width=9cm]{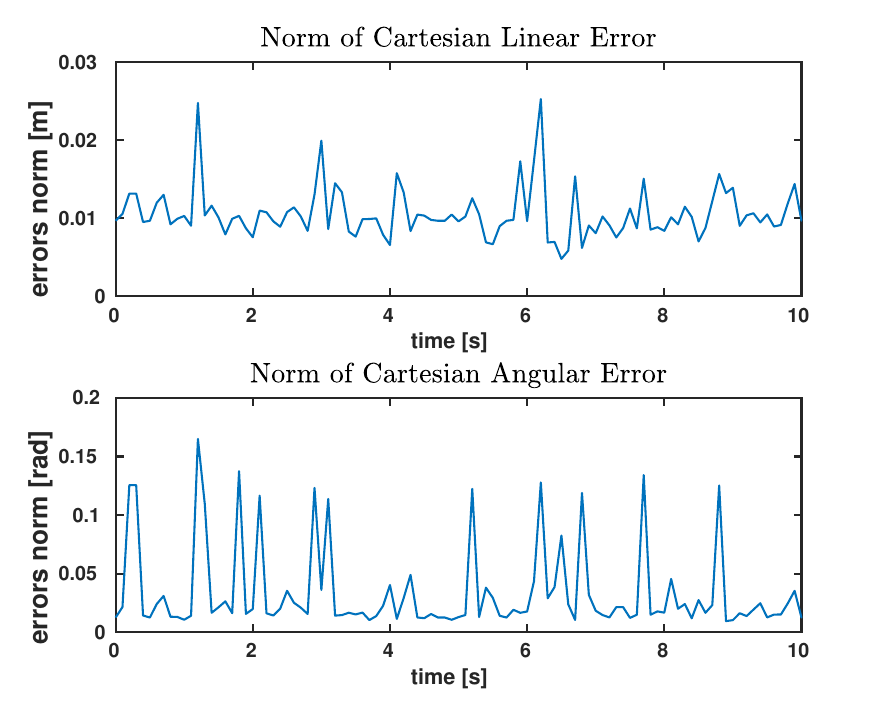}
	}
	\hspace*{15px}\textit{Mono (left Camera) Case} \hspace{125px} \textit{Mono (right Camera) Case}\\
	\vspace{30px}
	\centerline{
		\includegraphics[width=9cm]{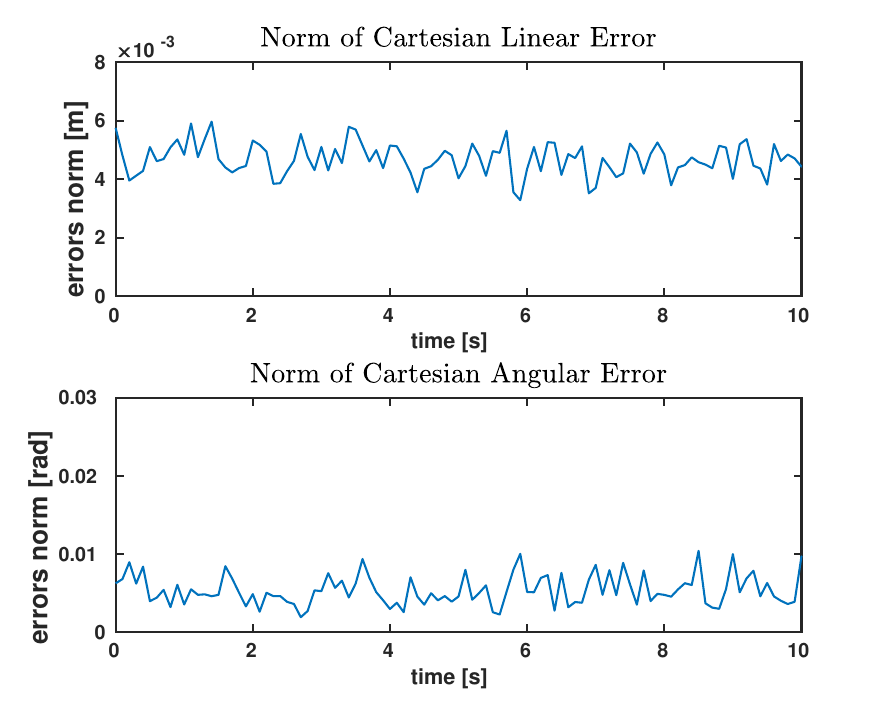}
		\includegraphics[width=9cm]{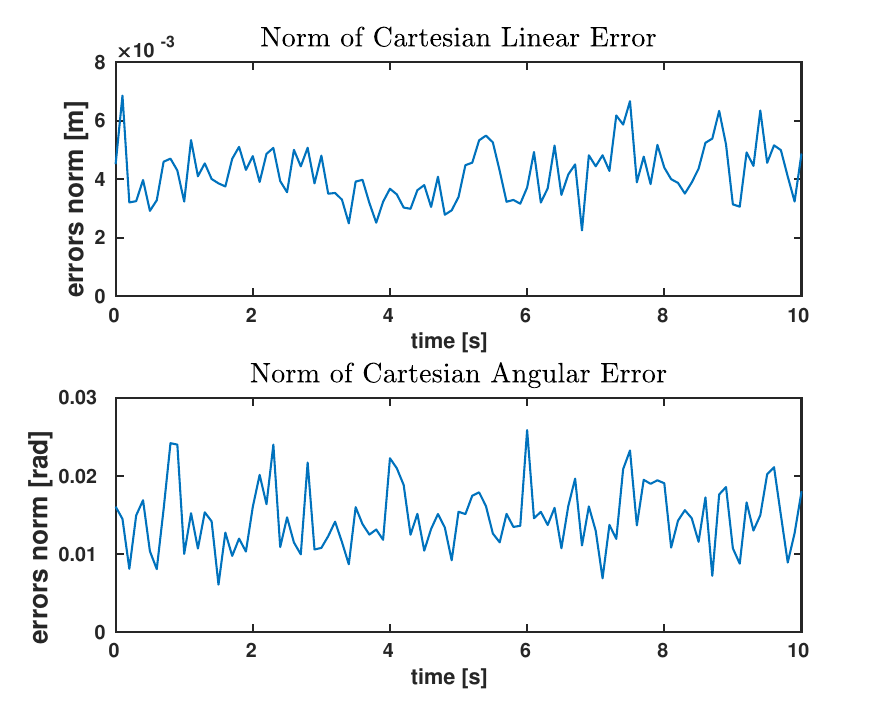}
	}
	\hspace*{20px}\textit{Stereo Camera Case} \hspace{135px} \textit{Stereo Depth Camera Case}\\
	\vspace{30px}
	\caption[Tracking error plots with Find Square detection initialization]{Linear and angular error (in norm) between the true pose and the estimated one. The tracking is initialized with the Find Square detection method of section \ref{subsec:findSquare}.}
	\label{fig:squareErrors}
\end{figure}

With the two good initializations (the ideal one and the Find Square one), the two stereo methods have similar results, overall better than the mono case. This is visible in the errors' plots of figure \ref{fig:clickErrors} and figure \ref{fig:squareErrors}. Anyway, we can also see that the mono camera case has not so bad performance. Considering that a stereo camera is much more expensive, a tracking with a single mono camera can be an advisable deal.\\

Another interesting outcome for the monocular case, is that the position of the camera influences the results. This is because different view angles provide different tracking performance.\\

In the plots of the errors (fig. \ref{fig:clickErrors}, fig. \ref{fig:squareErrors}, and fig. \ref{fig:templateErrors}), a lot of variations can be seen while the time goes on, although the robot and the object are still. This is due to the nature of the tracking algorithm, which continuously updates the pose at each new image received by the camera. Anyway, it must be noticed that the variations are little for both the linear and the angular parts. So, taking the pose at a certain time instead of an another is not so influential.

\chapter{Conclusions}

This thesis has presented a kinematic control architecture for two cooperative autonomous underwater manipulators.\\
The robot collaboration is done at kinematic level, exchanging vectors and matrices to produce a common tool Cartesian velocity. The cooperation scheme takes into account that underwater communication is difficult, and it keeps the amount of exchanged data as low as possible.\\
The experimental results showed how the Task Priority Inverse Kinematics approach used can deal with an assembly task: the \textit{peg-in-hole}. Being an unexplored problem for cooperative underwater manipulator, the scenario is simulated with many simplifications and assumptions.\\
Part of the problem deals with computer vision techniques. The thesis shows how some detection and tracking algorithm can be exploited to estimate the hole's pose.\\
For the insertion phase, a force-torque sensor is used to help the accomplishment of the mission, thanks to the data provided exploited by the control architecture.\\
Both the Control and the Vision part can help other works in various robotics fields, not only related to underwater intervention missions.\\

Code implementation is suited to easily manage objectives (e.g. to delete offline some objectives to try a different method). Big effort has been made in providing a modular and flexible code architecture. For example, the dependency from ROS (Robotics Operating System) is kept at minimum to make possible to easily adapt the code to a different kind of interface and/or simulator (which does not rely on ROS to communicate).\\

In the Chapter \ref{chap:introduction}, an introduction about the context is given, and the previous works in the relative fields has been recalled.\\
In Chapter \ref{chap:control}, the principal points of the theory behind the Control Architecture are summarized. Here, the mathematical foundations for the Task Priority Inverse Kinematic approach are recalled, considering also a coordination policy between multiple agents that fits in the chosen approach.\\
In Chapter \ref{chap:method}, the theory explained before is exploited to deal with the scenario stated by this thesis. A Force-Torque objective is inserted in the TPIK list to reduce the magnitudes of forces and torques that act on the peg during the insertion phase. This is noticeable because it is used a \enquote{dynamic} information (the force and the torque) in a pure-kinematic method. A simple, but suitable for the scenario, list of objectives is then described. An additional routine, part not of the kinematic layer but more of the Mission Managing layer, is explained. Considering the goal frame where the peg is driven to by the kinematic control, this new routine shifts its origin according to the direction of the forces acting on the peg. This is an additional method to further exploit the information given by the force-torque sensor.\\
In Chapter \ref{chap:results}, the simulated environment is detailed and the experimental results are discussed. The chosen simulator, UWSim, is introduced, along with some other underwater simulators that can be useful for the interested reader. To make the insertion phase realistic, collisions between the peg and the hole have been inserted in the simulation. These collisions propagates through the whole robotic system, thus affecting the arm and the vehicle. Another routine is implemented to fake a firm grasp of the tool by the two robots. In real environment, this constraint is assured by frictional forces, but in a pure-kinematic simulator like UWSim the frictions are not present. The tuning of the gains permits to not hide bad cooperation between the agents.  In the same Chapter \ref{chap:results}, assumptions to simplify the problem are explained, and an idea of how the Control Loop runs is given. Finally, results of the experiments done are presented and discussed. Three main experiments have been carried out: without hole's pose error, with a fixed error of 0.015m along one axis, and one final test which includes the Vision part with the hole's pose error given by the \textit{Detection} and the \textit{Tracking} algorithms used.\\
Chapter \ref{chap:vision} covers exclusively methods and tools used by the Vision robot. No theoretical background is given because it would take out of the scope of this thesis. The Chapter gives an idea on how two computer vision libraries, \href{https://opencv.org/}{\textbf{OpenCV}} (Open Source Computer Vision Library) [\cite{opencv}] and \href{https://visp.inria.fr/}{\textbf{ViSP}} (Visual Servoing Platform) [\cite{visp}], are used. The Vision part is divided into two phases: Detection and Tracking. For both, different algorithms have been tested, compared and discussed. 
In particular, for the Detection part, some methods have been discarded and they have not been used any more for further trials, but they are anyway presented in Appendix \ref{chap:AppendixVision}. They are all OpenCV algorithms that can help in other applications.\\
This Chapter \ref{chap:conclusions} concludes the thesis and it gives some starting points for possible future works.\\
The Appendix \ref{chap:AppendixCode} gives some details on how the software is implemented, together with a list of some useful libraries used that surely can help to develop a control architecture in the C++ programming language.\\

For some on-going progresses in this scenario, it can be useful for the reader to follow the TWINBOT project [\cite{TWINBOT2019}]. This thesis' context derives from the scenario of this project, but it evolves independently because at the time this thesis was being developed, TWINBOT was in a very early stage.

\section{Future Works}
Since the novelty of the application, further works can be pursued in various directions.\\
For what concerns the experiments, a dynamic simulation, along with a dynamic controller, can be introduced to better analyse the methods adopted. This would mean to include effects that would increase the realism of the simulation, such as buoyancy, thrusters modelling, disturbances of the arm to the vehicle and vice-versa, real tool-grasping effects, even some water currents. Some work has been done in this direction but then it has not been pursued due to the lack of time. These efforts, even if they are not presented in this thesis, showed that the first step to introduce dynamics could be using the plugin \href{https://github.com/freefloating-gazebo/freefloating_gazebo}{FreeFloatingGazebo} [\cite{freeFloatingGazebo}], mentioned in section \ref{sec:simulators}. This one is the most suitable tool to be used from the actual work because its scope is to solve the lack of dynamics of UWSim, expanding the functionalities of the simulator. So, it would be easy to adapt the code to the new simulations. For example, the scene (i.e. the file which describes the simulated scenario) would be the same, and ROS would be always used as interface.\\

Regarding the actual chosen architecture, the Force Torque objective idea can be improved. For example, we can consider a different task reference, calculated not only as a proportional error between the desired force (that is zero) and the actual detected one.\\
Another improvement can be for the Change Goal routine. We could let the forces and the torques modify also the orientation of the goal frame, to reduce/eliminate the angular error between the real goal frame and the one estimated.\\

Regarding the insertion phase, additional problems can be explored. This thesis focused only on collisions that may happen when the peg is already inside the hole. If some contact between the external surface of the hole's structure and the peg's tip happens, the mission fails (i.e., it occurs a stalemate where the peg bounces forever against the hole's surface, unable to find the hole). In the literature, various methods have been explored toward this point. For example, researchers have considered the cases when the peg meets the hole with a bad alignment that creates a two or three points contact (as briefly explained in \ref{sec:artPeg}). However, to the best of this author's knowledge, the \textit{peg-in-hole} problem has never been studied when the protagonists are two autonomous mobile manipulator (in any scenarios, not specifically underwater ones). So, it can be interesting to adapt old tools to this particular (cooperative and underwater) field.\\

Towards a more realistic intervention missions, efforts can be spent to consider ways to localize the robot under the water's surface. In this thesis, it is assumed that all agents have a reference frame in common, but, usually, underwater localization is really an issue. Some cooperative methods (this time not at kinematic level) can be considered to make some surface vessels help the underwater agents to localize themselves (a problem explored in the WiMUST project [\cite{wimust}]).\\

In this thesis, some assumptions have been made for the Vision part. Further works could consider to relax some of them, for example to increase the difficulty of the detection and tracking phases. In an underwater scenario, not always the water permits to watch from afar, and illumination and distortions can be other important issues.\\

There is always a lot of work towards increasing the capacity of \mbox{intervention-AUVs}. For example, we can consider more specifically communication issues and, so, new techniques to exchange data between agents; especially in an underwater scenario, we can't share too much information among robots and with too high frequency. Also, other kinds of assembly problems can be addressed: for instance, a \textit{peg-in-hole} one where the \textit{peg} is held by only one robot and the \textit{hole} by another one.\\ Some objectives of the TWINBOT project [\cite{TWINBOT2019}] aim to study these two just mentioned problems.

\appendix

\chapter{C++ Code Scheme}
\label{chap:AppendixCode}

Some effort has been spent to implement a flexible and modular C++ code architecture. The main focus is directed to facilitate the use of the TPIK approach. With this scheme, adding and removing tasks from the code is easy and straightforward. Furthermore, even if ROS is used as interface, other communication methods (for different simulators, or even for real robots) can be used easily, changing a little part of the code. This is due to the fact that all the ROS parts (the \textit{interface} classes) are written in separate files from the ones of the main blocks.\\
Only a rough idea about the code scheme is given here, without too much details. Further explanations can be found in the github page (\url{https://github.com/torydebra/AUV-Coop-Assembly}) and in the code documentation (\url{https://torydebra.github.io/AUV-Coop-Assembly/}).

\section{Tools}
The Control Architecture is implemented in C++ language. Some external support libraries have been used, and, in this section, the most relevant ones are described. Please note that a particular section (section \ref{sec:simulators}) is dedicated to the choice of the simulator, and another lists the main tools used by Vision (section \ref{sec:visionTools}).

\begin{itemize}
	\item \href{http://www.ros.org/}{\textbf{ROS}} (Robot Operating System), the well-known robotic middleware. It is used to communicate with the simulator, which means sending commands to robots and receiving information by the on-going simulations (e.g. robots states, data from sensors, streaming images from cameras).
	
	\item \href{http://eigen.tuxfamily.org/index.php?title=Main_Page}{\textbf{Eigen}} [\cite{eigen}], a C++ library for linear algebra. It is very useful to deal with matrix computation and management in any C++ software.
	
	\item \textbf{CMAT}, another C++ library, developed at GRAAL laboratory of University of Genova (\url{http://www.graal.dibris.unige.it/}). It implements the core functions for the TPIK method, detailed in \cite{IntroMaris1}.
	
	\item \href{http://www.orocos.org/kdl}{\textbf{Orocos KDL}}, a package to deal with kinematic and dynamic chains. Here it is used to compute the Jacobian of the robots, given the arm and the vehicle configuration.
\end{itemize}

\section{The Single Robot Code Scheme}
\begin{figure}[H]
	\begin{center}
		\includegraphics[width=1\columnwidth]{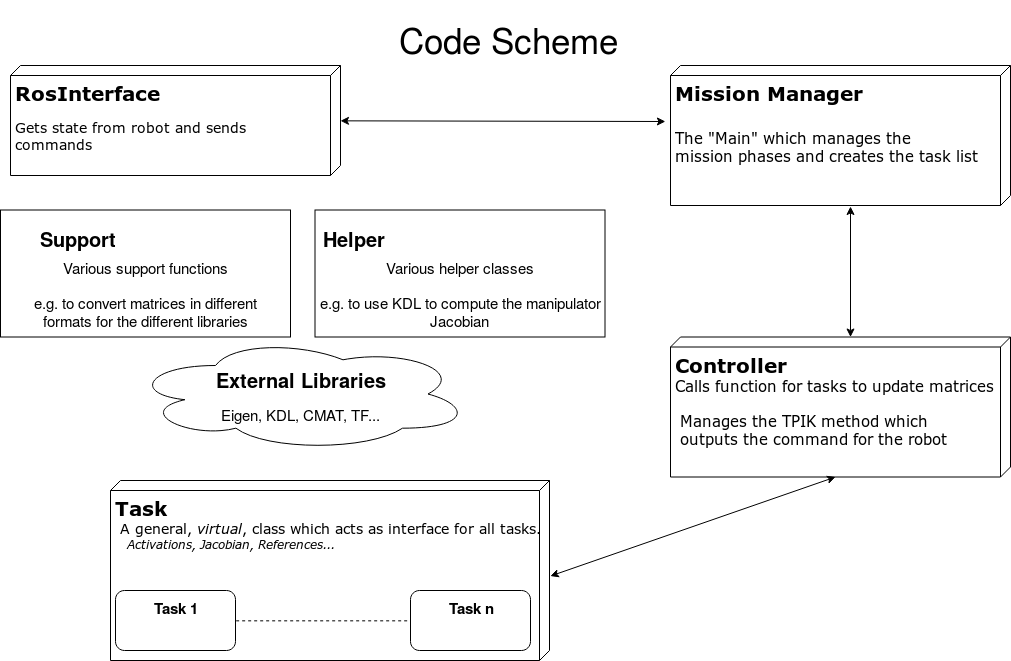}
		\caption[C++ code scheme for the single robot]{The C++ code scheme that shows the relationship among the main blocks of the single robot. Parallelepipeds represent the principal blocks, while rectangles contain useful functions used by them. Arrows indicate the communication between the main blocks.}
		\label{fig:codeSchemeSingle}
	\end{center}
\end{figure}

In this section it is presented the scheme for the single carrying robot. The two cooperative robots share the same code, and they are differentiated (when necessary) thanks to their names (\textit{g500\_A} and \textit{g500\_B}).

\begin{itemize}
	\item \textbf{RosInterface}. This class is used to communicate with the simulator, e.g. sending commands to the robot, receiving state and sensor information, and so on. It is obviously ROS-dependent, and it should be replaced if another middleware is used.
	
	\item \textbf{Mission Manager}. This block is the \enquote{main}. It initializes all the useful classes, it creates the tasks list, and it manages the whole mission.
	
	\item \textbf{Controller}. This class is the core of the control; in practice it is the kinematic layer. It generates commands for the vehicle according to the prioritized list of tasks. It is where the iCAT algorithm based on the TPIK approach (section \ref{sec:tpik}) is used.
	
	\item \textbf{Task}. This is an \textit{abstract} class. It acts as a base class for all the specific \textit{concrete} classes of tasks. In this way, the controller and the mission manager simply handle a vector of pointer to Task. The Mission Manager creates a concrete class for each task and then it fills the vector. This vector is the prioritized list of the TPIK method. The controller can iterate the element of this vector and can call the abstract methods of Task without worrying of which real concrete task is actually inside the list.
	
	\item \textbf{Support}. It is a group of various \textit{namespaces}, used for conversions, to print to file, and to use some mathematical formulas.
	
	\item \textbf{Helper}. It contains a group of classes and headers used to help the control scheme (e.g. to compute the Jacobian with KDL) and to log results.
\end{itemize}

\section{The Whole Code Scheme}
\begin{figure}[H]
	\begin{center}
		\includegraphics[width=1\columnwidth]{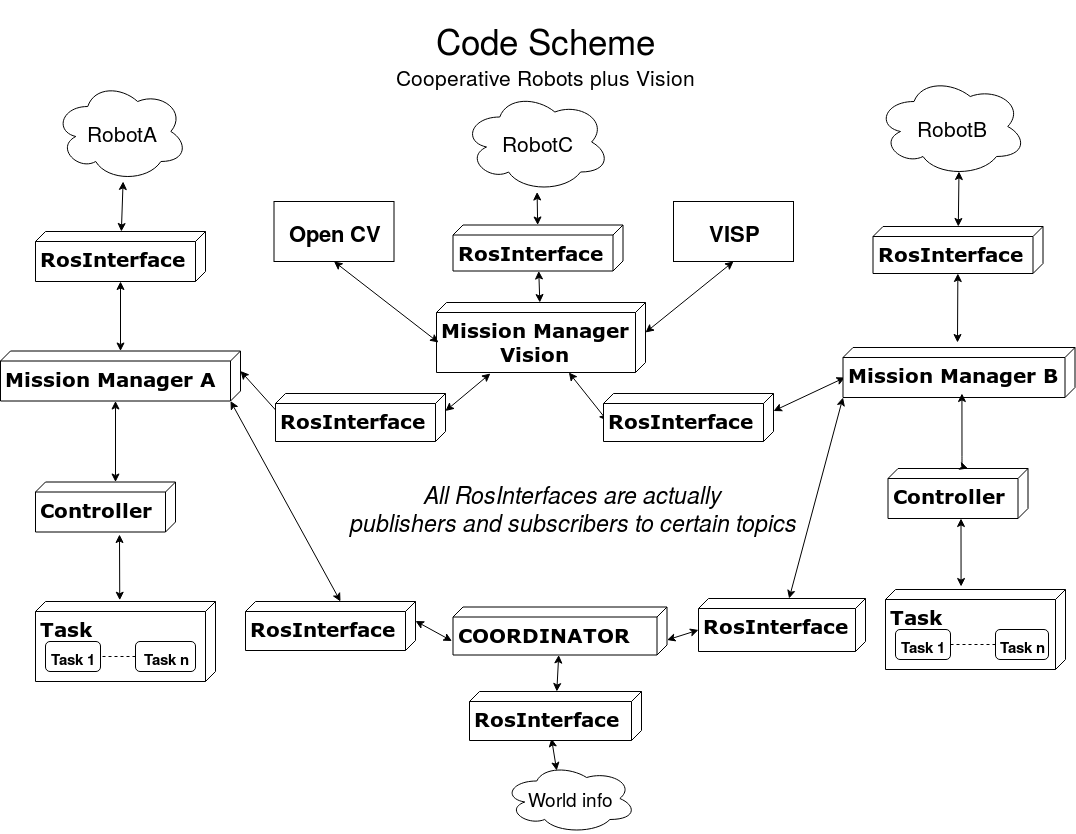}
		\caption[C++ code scheme for the whole architecture]{The C++ code scheme that shows the relationship among the main blocks of the three robots. On the left and on the right side there is a zoomed out view of the previous figure \ref{fig:codeSchemeSingle}, corresponding to the scheme for the two carrying agents. In the centre, there are the blocks for the Vision robot and for the Coordinator.}
		\label{fig:codeSchemeWhole}
	\end{center}
\end{figure}
	
The two robots must communicate between them and with the Vision robot. Like explained before, ROS is used to communicate with the simulator, but it is also used to make different nodes (e.g. Coordinator and Robots) to communicate.\\
Please note that the Coordinator is not a physical agent: it can be put as a software routine on one robot. This would help with the communication issues typical of underwater scenarios, because only data-exchange between the Coordinator and the other robot will pass through the water.\\
The scheme for the Vision robot is simple because it is driven as a ROV: it is not autonomous and so no TPIK is implemented for it. Anyway, it can be switched easily into an autonomous robot, with or without TPIK (that it is not really necessary for this agent).\\
The Coordinator is a node in charge of dealing with the coordination policy explained in section \ref{sec:coopScheme}. It also needs information from the world (i.e. the simulation) to compute the cooperative velocity.

\chapter{Other Algorithms for Object Detection}
\label{chap:AppendixVision}

During the simulations, several trials have been done to find a suitable algorithm for the detection of the hole structure. In Chapter \ref{chap:vision} two methods have been discussed as the successful ones. In this appendix, others that have been discarded are briefly presented. Even if they are not used in the last versions of experiments, they can be useful for other purposes, such as detection of other kind of shapes.\\
Each algorithm is taken from OpenCV Detection tutorials (\url{https://docs.opencv.org/3.4/d9/d97/tutorial_table_of_content_features2d.html}), where also other interesting methods can be found.

\section{Corner Detection with the Shi-Harris Method}

\begin{figure}[H]
	\centering
	\includegraphics[width=8.0cm]{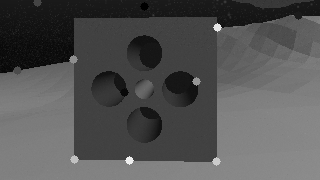}
	\caption[Result of \textit{goodFeaturesToTrack()}]{\textit{goodFeaturesToTrack()} result. Only two detected points are the real corners, and the upper ones are not detected.} 
	\label{fig:goodFeatToTrack}
\end{figure}

Following the tutorial (\url{https://docs.opencv.org/3.4.6/d8/dd8/tutorial_good_features_to_track.html}), it has been implemented a corner detector with the Shi-Harris method [\cite{Shi94goodfeatures}] using the OpenCV function \href{https://docs.opencv.org/3.4.6/dd/d1a/group__imgproc__feature.html#ga1d6bb77486c8f92d79c8793ad995d541}{goodFeaturesToTrack()}.\\
This function, in our case, can be useful to find the corners of the hole structure, which is the necessary initialization for the Tracking method used after (section \ref{sec:visDetect}).\\

The original example lets change the number of maximum points to be found. This is useful to reduce the number of false positive corners. The main problem is that the real corners of the square are not the "best" ones. So we can't simply put this parameter equal to 4. On the other side, with a large number of points, would be then difficult to discriminate the right corners from the others.\\
Other interesting parameters are:
\begin{itemize}
	\item \textbf{minDistance}. The minimum distance among the corners to be found.
	\item \textbf{qualityLevel}. A parameter which characterizes the minimum accepted quality of image corners.
	\item \textbf{blockSize}. Size of an average block for computing a derivative covariance matrix over each pixel's neighbourhood.
	\item \textbf{mask}. To specify a certain region of interest in the image. In such a way, corners are searched only in this region. The problem in our case is that without prior works we can't know where the interesting region is (i.e. the region which contains the hole).
\end{itemize}
The points detected are effectively good feature points (as can be seen in figure \ref{fig:goodFeatToTrack}). But, the best ones are not the ones that we want to detect (i.e., the corner of the square).\\
This method should be used as a low level algorithm, to then help higher level ones. For example, to construct some polygons and to check if these polygons are square/rectangles. However, to follow this direction should be better to start from the edges (as done in section \ref{subsec:findSquare}).\\
  
\section{Canny Edge and Hough Transform}
\label{sec:HoughTrasf}

\begin{figure}[H]
	\centering
	\centerline{
		\includegraphics[width=7.0cm]{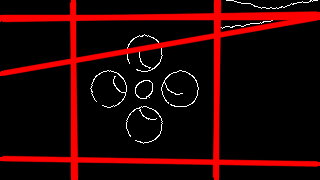}
		\qquad
		\includegraphics[width=7.0cm]{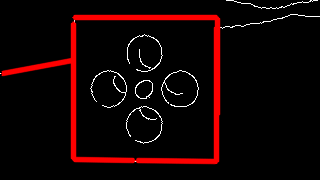}
	}
	\caption[Results of the Standard Hough Transform and the Probabilistic one]{Results of the Hough Standard Transform (left) and the Probabilistic one (right). In white they are depicted all the edges detected with Canny; the red lines are the detected straight lines, outputs of the method.}
	\label{fig:HoughStandard}
\end{figure}

The Hough Transform [\cite{DudaHoughTrasf}] is a method to detect straight lines in an image. Usually, a preprocessing of the image with an edge detector is used to improve the results, for example with a Canny Edge Detector [\cite{CannyEdge}].\\

The OpenCV tutorial (\url{https://docs.opencv.org/3.4/d9/db0/tutorial_hough_lines.html}) makes use of two types of Hough Transform: the standard \href{https://docs.opencv.org/3.4/dd/d1a/group__imgproc__feature.html#ga46b4e588934f6c8dfd509cc6e0e4545a}{\textit{HoughLines()}} and the probabilistic \href{https://docs.opencv.org/3.4/dd/d1a/group__imgproc__feature.html#ga8618180a5948286384e3b7ca02f6feeb}{\textit{HoughLinesP()}} [\cite{houghprob}].\\
Results are visible in figure \ref{fig:HoughStandard}. The results on the right shows that the probabilistic version is good to detect the square structure of the hole. 
Thus, this method can be used as a good preliminary step to then extract the corner from the detected square.

\newpage
\section{Bounding Box Detection}
\label{sec:boundingBox}

\begin{figure}[H]
	\centering
	\centerline{
	\includegraphics[width=7.0cm]{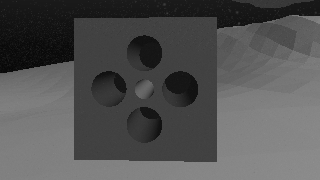}
	\qquad
	\includegraphics[width=7.0cm]{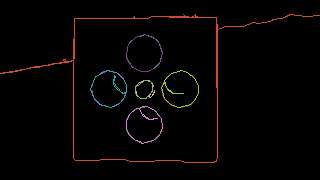}
	}
	\caption[Result of \emph{findContours()}]{Result of \emph{findContours()}: on the left the original image, on the right the contours detected.}
	\label{fig:BoundBoxresultOnlyPolig}
\end{figure}

\begin{figure}[H]
	\centering
	\includegraphics[width=7.0cm]{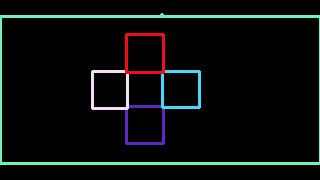}
	\caption[Result of Bounding Box detection]{Result of Bounding Box detection, where they are visible the drawn bounding boxes around the holes.}
	\label{fig:BoundBoxresultOnlyRect}
\end{figure}

The code derived from an OpenCV tutorial (\url{https://docs.opencv.org/3.4.6/de/d62/tutorial_bounding_rotated_ellipses.html}).\\
First, a Canny edge detector is used to preprocess the image. Then, the function \href{https://docs.opencv.org/3.4.6/d3/dc0/group__imgproc__shape.html#ga17ed9f5d79ae97bd4c7cf18403e1689a}{\textit{findContours()}} is called to retrieve contours with the algorithm described in \cite{findcountors}. As can be noticed, these initial steps are the same of the implemented method of section \ref{subsec:findSquare}. The difference in this tutorial is that, after these passages, bounding boxes are drawn around some particular shapes detected.\\

The result after the first step is shown in figure \ref{fig:BoundBoxresultOnlyPolig}. We can see that this passage already reveals the square, that is important for the method described in \mbox{section \ref{subsec:findSquare}.}\\
Instead, the algorithm presented here continues in a different direction, which brings us to the final result of figure \ref{fig:BoundBoxresultOnlyRect}.\\

This tutorial is recalled because it can be useful to find other kinds of shapes, for example an hole structure which is not a rectangle or a square.

\section{2D Feature Matching \& Homography}
\begin{figure}[H]
	\centering
	\centerline{
		\includegraphics[width=3.8cm]{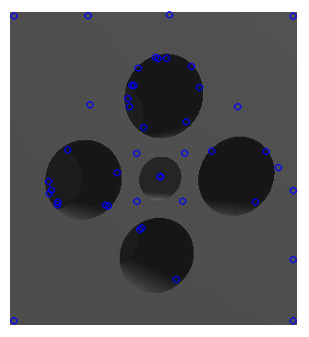}
		\qquad
		\includegraphics[width=7.1cm]{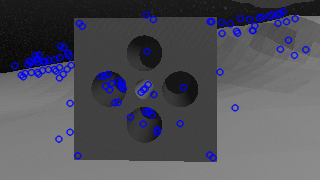}
	}
	\vspace{10px}
	\includegraphics[width=9cm]{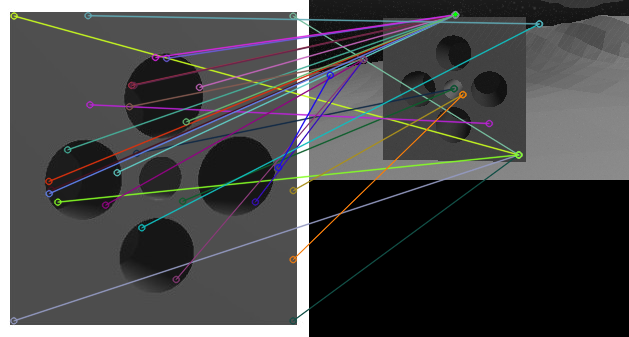}
	\caption[Result of 2D Feature Matching]{Result of the 2D Feature Matching. Above, the blue circles are the detected features in the \textit{object image} (on the left) and in the \textit{scene image} (on the right). Below, the output of the matching passage, which shows clearly a bad outcome.}
	\label{fig:featHomog}
\end{figure}

\textit{Image features} are small patches that are useful to compute similarities between images. These are different from corner points; they indicates particular details that stand out in the image. Detecting these areas is useful to recognize objects of interest, as a sort of \textit{template matching} (please note that this method is not a template matching as the one of section \ref{subsec:templateMatch}).\\
The \textit{descriptors} of these features contain the visual description of the patches, which are used to recognize similarities between different images.\\

This method needs an \textit{object image} and a \textit{scene image}. The first is a sort of template which contains only the object to be found (in this case, the square face of the hole). The second is the image in which we want to detect this object (in this case, what the camera is seeing).\\
After good features are extracted from both images, the \textit{descriptors} are used to \textit{match} them, thus, detecting the object in the scene.
Then, it is necessary to find the perspective transformation between the object image and the scene (i.e. find the homography). This is needed to take into account that usually the pose and the scaling of the object in the scene are not the same of the ones of the object image.\\
The OpenCV tutorial (\url{https://docs.opencv.org/3.4/d7/dff/tutorial_feature_homography.html}) uses different tools:
\begin{itemize}
	\item \textbf{SURF} (Speeded Up Robust Features) Detector [\cite{surfDet}] to extract the features, and to compute the descriptors.
	\item \textbf{FLANN} (Fast Library for Approximate Nearest Neighbors) matcher [\cite{flannMatch}] to match the features of the two images.
	\item \textbf{Lowe's ratio test} [\cite{loweTest}] to take only the best matches.
	\item \textbf{RANSAC} (RANdom SAmple Consensus) [\cite{ransacHomog}] method to find the homography with the function \href{https://docs.opencv.org/3.4/d9/d0c/group__calib3d.html#ga4abc2ece9fab9398f2e560d53c8c9780}{\textit{findHomography()}}.
\end{itemize}

\vspace{20px}
In this case, results are unsatisfactory, as can be seen in figure \ref{fig:featHomog}. The main problem is that in this particular scene there are not nice distinct features. Also, the symmetry of the structure does not help, because there are a lot of particulars that are similar (like the square sides and the holes). As can be seen in the results of the previously cited \href{https://docs.opencv.org/3.4/d7/dff/tutorial_feature_homography.html}{tutorial}, good outcomes are obtained for food boxes. In fact, this method is often exploited for scenes where a lot of details are present (e.g. graffiti painting, supermarket shelf). In underwater cases, realistic infrastructures have not so many details, and so also the simulated scenario of this thesis.\\

There are a lot of parameters to set for the three main tools (SURF, FLANN, and RANSAC). Various options have been tried but no-one was satisfactory. Also, different detectors (like SWIFT [\cite{loweTest}]) and matchers (like Brute-Force), have been tried, again with poor results.\\
Anyway, the variety of tools and parameters makes this method suitable for a lot of applications, and it should be taken into consideration in other applications.

\bibliographystyle{Classes/RoboticsBiblio}    %
\renewcommand{\bibname}{References}           %
\bibliography{References/references}          %

\begin{thebibliography}{107}
\expandafter\ifx\csname natexlab\endcsname\relax\def\natexlab#1{#1}\fi

\bibitem[{Abdullah {\em et~al.\/}(2015)Abdullah, Roth, Weyrich \&
  Wahrburg}]{IntroPeg8}
{\sc Abdullah, M., Roth, H., Weyrich, M. \& Wahrburg, J.} (2015). An approach
  for peg-in-hole assembling using intuitive search algorithm based on human
  behavior and carried by sensors guided industrial robot. {\em
  IFAC-PapersOnLine\/}, {\bf 48}, 1476--1481.

\bibitem[{Abreu {\em et~al.\/}(2016)Abreu, Morishita, Pascoal, Ribeiro \&
  Silva}]{wimust}
{\sc Abreu, P., Morishita, H., Pascoal, A., Ribeiro, J. \& Silva, H.} (2016).
  {Marine Vehicles with Streamers for Geotechnical Surveys: Modeling,
  Positioning, and Control}. {\em IFAC-PapersOnLine\/}.

\bibitem[{Antonelli(2009)}]{IntroTpik20}
{\sc Antonelli, G.} (2009). Stability analysis for prioritized closed-loop
  inverse kinematic algorithms for redundant robotic systems. {\em IEEE
  Transactions on Robotics\/}, {\bf 25}, 985--994.

\bibitem[{Antonelli \& Chiaverini(1998)}]{IntroTpik12}
{\sc Antonelli, G. \& Chiaverini, S.} (1998). Task-priority redundancy
  resolution for underwater vehicle-manipulator systems. In {\em Proceedings.
  1998 IEEE International Conference on Robotics and Automation (Cat.
  No.98CH36146)\/}, vol.~1, 768--773 vol.1.

\bibitem[{Antonelli \& Chiaverini(2003)}]{IntroTpik13}
{\sc Antonelli, G. \& Chiaverini, S.} (2003). Fuzzy redundancy resolution and
  motion coordination for underwater vehicle-manipulator systems. {\em IEEE
  Transactions on Fuzzy Systems\/}, {\bf 11}, 109--120.

\bibitem[{Antonelli {\em et~al.\/}(2008)Antonelli, Arrichiello \&
  Chiaverini}]{IntroTpik9}
{\sc Antonelli, G., Arrichiello, F. \& Chiaverini, S.} (2008). The
  null-space-based behavioral control for autonomous robotic systems. {\em
  Intelligent Service Robotics\/}, {\bf 1}, 27--39.

\bibitem[{Antonelli {\em et~al.\/}(2009)Antonelli, Indiveri \&
  Chiaverini}]{antoSat}
{\sc Antonelli, G., Indiveri, G. \& Chiaverini, S.} (2009). Prioritized
  closed-loop inverse kinematic algorithms for redundant robotic systems with
  velocity saturations. 5892--5897.

\bibitem[{Baerlocher \& Boulic(2004)}]{IntroTpik27}
{\sc Baerlocher, P. \& Boulic, R.} (2004). An inverse kinematics architecture
  enforcing an arbitrary number of strict priority levels. {\em The Visual
  Computer\/}, {\bf 20}, 402--417.

\bibitem[{Bay {\em et~al.\/}(2006)Bay, Tuytelaars \& Van~Gool}]{surfDet}
{\sc Bay, H., Tuytelaars, T. \& Van~Gool, L.} (2006). {SURF}: Speeded up robust
  features. In A.~Leonardis, H.~Bischof \& A.~Pinz, eds., {\em Computer Vision
  -- ECCV 2006\/}, 404--417, Springer Berlin Heidelberg, Berlin, Heidelberg.

\bibitem[{Bingham {\em et~al.\/}(2010)Bingham, Foley, Singh, Camilli,
  Delaporta, Eustice, Mallios, Mindell, Roman \& Sakellariou}]{IntroApp4}
{\sc Bingham, B., Foley, B., Singh, H., Camilli, R., Delaporta, K., Eustice,
  R., Mallios, A., Mindell, D., Roman, C. \& Sakellariou, D.} (2010). Robotic
  tools for deep water archaeology: Surveying an ancient shipwreck with an
  autonomous underwater vehicle. {\em Journal of Field Robotics\/}, {\bf 27},
  702--717.

\bibitem[{Bradski(2000)}]{opencv}
{\sc Bradski, G.} (2000). {The OpenCV Library}. {\em Dr. Dobb's Journal of
  Software Tools\/}.

\bibitem[{{Canny}(1986)}]{CannyEdge}
{\sc {Canny}, J.} (1986). A computational approach to edge detection. {\em IEEE
  Transactions on Pattern Analysis and Machine Intelligence\/}, {\bf PAMI-8},
  679--698.

\bibitem[{Capocci {\em et~al.\/}(2017)Capocci, Dooly, Omerdic, Coleman, Newe \&
  Toal}]{IntroApp5}
{\sc Capocci, R., Dooly, G., Omerdic, E., Coleman, J., Newe, T. \& Toal, D.}
  (2017). Inspection-class remotely operated vehicles—a review. {\em Journal
  of Marine Science and Engineering\/}, {\bf 5}, 13.

\bibitem[{Carrera {\em et~al.\/}(2014)Carrera, Palomeras, Hurtos, Kormushev \&
  Carreras}]{IntroPandora2}
{\sc Carrera, A., Palomeras, N., Hurtos, N., Kormushev, P. \& Carreras, M.}
  (2014). Learning by demonstration applied to underwater intervention. vol.
  269.

\bibitem[{Casalino {\em et~al.\/}(2001)Casalino, Angeletti, Bozzo \&
  Marani}]{IntroAMADEUS3}
{\sc Casalino, G., Angeletti, D., Bozzo, T. \& Marani, G.} (2001). Dexterous
  underwater object manipulation via multi-robot cooperating systems. In {\em
  Proceedings 2001 ICRA. IEEE International Conference on Robotics and
  Automation (Cat. No.01CH37164)\/}, vol.~4, 3220--3225 vol.4.

\bibitem[{Casalino {\em et~al.\/}(2002)Casalino, Angeletti, Cannata \&
  Marani}]{IntroAMADEUS2}
{\sc Casalino, G., Angeletti, D., Cannata, G. \& Marani, G.} (2002). The
  functional and algorithmic design of {AMADEUS} multirobot workcell. In S.K.
  Choi \& J.~Yuh, eds., {\em Underwater Vehicle Technology\/}, vol.~12.

\bibitem[{Casalino {\em et~al.\/}(2009)Casalino, Turetta, Sorbara \&
  Simetti}]{IntroTpik17}
{\sc Casalino, G., Turetta, A., Sorbara, A. \& Simetti, E.} (2009).
  Self-organizing control of reconfigurable manipulators: A distributed dynamic
  programming based approach. In {\em 2009 ASME/IFToMM International Conference
  on Reconfigurable Mechanisms and Robots\/}, 632--640.

\bibitem[{Casalino {\em et~al.\/}(2012)Casalino, Zereik, Simetti, Torelli,
  Sperind\'e \& Turetta}]{IntroTrident2}
{\sc Casalino, G., Zereik, E., Simetti, E., Torelli, S., Sperind\'e, A. \&
  Turetta, A.} (2012). Agility for underwater floating manipulation task and
  subsystem priority based control strategy. In {\em International Conference
  on Intelligent Robots and Systems (IROS 2012)\/}, 1772--1779.

\bibitem[{Casalino {\em et~al.\/}(2014)Casalino, Caccia, Caiti, Antonelli,
  Indiveri, Melchiorri \& Caselli}]{IntroMaris0}
{\sc Casalino, G., Caccia, M., Caiti, A., Antonelli, G., Indiveri, G.,
  Melchiorri, C. \& Caselli, S.} (2014). {MARIS}: A national project on marine
  robotics for interventions. In {\em 22nd Mediterranean Conference on Control
  and Automation\/}, 864--869.

\bibitem[{Chang {\em et~al.\/}(2011)Chang, Y.~Lin \& S.~Lin}]{IntroPeg3}
{\sc Chang, R.J., Y.~Lin, C. \& S.~Lin, P.} (2011). Visual-based automation of
  peg-in-hole microassembly process. {\em Journal of Manufacturing Science and
  Engineering\/}, {\bf 133}, 041015.

\bibitem[{Cheng~Chang {\em et~al.\/}(2004)Cheng~Chang, Yuan~Chang \&
  Ting~Cheng}]{IntroSpecApp1}
{\sc Cheng~Chang, C., Yuan~Chang, C. \& Ting~Cheng, Y.} (2004). Distance
  measurement technology development at remotely teleoperated robotic
  manipulator system for underwater constructions. 333 -- 338.

\bibitem[{Chhatpar \& Branicky(2001)}]{IntroPeg10}
{\sc Chhatpar, S.R. \& Branicky, M.S.} (2001). Search strategies for
  peg-in-hole assemblies with position uncertainty. In {\em Proceedings 2001
  IEEE/RSJ International Conference on Intelligent Robots and Systems.
  Expanding the Societal Role of Robotics in the the Next Millennium (Cat.
  No.01CH37180)\/}, vol.~3, 1465--1470 vol.3.

\bibitem[{Chiaverini(1997)}]{IntroTpik8}
{\sc Chiaverini, S.} (1997). Singularity-robust task-priority redundancy
  resolution for real-time kinematic control of robot manipulators. {\em IEEE
  Transactions on Robotics and Automation\/}, {\bf 13}, 398--410.

\bibitem[{Christ \& Wernli(2013)}]{IntroSpecApp4}
{\sc Christ, R. \& Wernli, R.} (2013). {\em The {ROV} Manual: A User Guide for
  Remotely Operated Vehicles\/}. Elsevier, 2nd edn.

\bibitem[{Cieslak {\em et~al.\/}(2015)Cieslak, Ridao \&
  Giergiel}]{IntroPandora3}
{\sc Cieslak, P., Ridao, P. \& Giergiel, M.} (2015). Autonomous underwater
  panel operation by {GIRONA500} {UVMS}: A practical approach to autonomous
  underwater manipulation. In {\em 2015 IEEE International Conference on
  Robotics and Automation (ICRA)\/}, 529--536.

\bibitem[{Comport {\em et~al.\/}(2006)Comport, Marchand, Pressigout \&
  Chaumette}]{visp-edge}
{\sc Comport, A., Marchand, E., Pressigout, M. \& Chaumette, F.} (2006).
  {Real-time markerless tracking for augmented reality: the virtual visual
  servoing framework}. {\em {IEEE Transactions on Visualization and Computer
  Graphics}\/}, {\bf 12}, 615--628.

\bibitem[{{Cook} {\em et~al.\/}(2014){Cook}, {Vardy} \&
  {Lewis}}]{simComparisonCook}
{\sc {Cook}, D., {Vardy}, A. \& {Lewis}, R.} (2014). A survey of {AUV} and
  robot simulators for multi-vehicle operations. In {\em 2014 IEEE/OES
  Autonomous Underwater Vehicles ({AUV})\/}, 1--8.

\bibitem[{Di~Lillo {\em et~al.\/}(2016)Di~Lillo, Simetti, De~Palma, Cataldi,
  Indiveri, Antonelli \& Casalino}]{IntroDexrov}
{\sc Di~Lillo, P.A., Simetti, E., De~Palma, D., Cataldi, E., Indiveri, G.,
  Antonelli, G. \& Casalino, G.} (2016). Advanced {ROV} autonomy for efficient
  remote control in the {DexROV} project. {\em Marine Technology Society
  Journal\/}, {\bf 50}.

\bibitem[{Diaz~Ledezma {\em et~al.\/}(2015)Diaz~Ledezma, Amer, Abdellatif,
  Outa, Trigui, Patel \& Binyahib}]{IntroSpecApp6}
{\sc Diaz~Ledezma, F., Amer, A., Abdellatif, F., Outa, A., Trigui, H., Patel,
  S. \& Binyahib, R.} (2015). A market survey of offshore underwater robotic
  inspection technologies for the oil and gas industry.

\bibitem[{Dietrich {\em et~al.\/}(2010)Dietrich, Buchholz, Wobbe, Sowinski,
  Raatz, Schumacher \& Wahl}]{IntroPeg9}
{\sc Dietrich, F., Buchholz, D., Wobbe, F., Sowinski, F., Raatz, A.,
  Schumacher, W. \& Wahl, F.M.} (2010). On contact models for assembly tasks:
  Experimental investigation beyond the peg-in-hole problem on the example of
  force-torque maps. In {\em 2010 IEEE/RSJ International Conference on
  Intelligent Robots and Systems\/}, 2313--2318.

\bibitem[{Djapic {\em et~al.\/}(2013)Djapic, Na\dj, Ferri, Omerdic, Dooly, Toal
  \& Vukić}]{IntroSpecApp3}
{\sc Djapic, V., Na\dj, D., Ferri, G., Omerdic, E., Dooly, G., Toal, D. \&
  Vukić, Z.} (2013). Novel method for underwater navigation aiding using a
  companion underwater robot as a guiding platforms. In {\em 2013 MTS/IEEE
  OCEANS - Bergen\/}, 1--10.

\bibitem[{Drap(2012)}]{IntroApp3}
{\sc Drap, P.} (2012). Underwater photogrammetry for archaeology. {\em Special
  Applications of Photogrammetry\/}.

\bibitem[{Duda \& Hart(1972)}]{DudaHoughTrasf}
{\sc Duda, R.O. \& Hart, P.E.} (1972). Use of the hough transformation to
  detect lines and curves in pictures. {\em Commun. ACM\/}, {\bf 15}, 11--15.

\bibitem[{E.~Rohmer(2013)}]{vrep}
{\sc E.~Rohmer, M.F., S. P. N.~Singh} (2013). {V-REP}: a versatile and scalable
  robot simulation framework. In {\em Proc. of The International Conference on
  Intelligent Robots and Systems (IROS)\/}.

\bibitem[{Escande {\em et~al.\/}(2014)Escande, Mansard \& Wieber}]{IntroTpik29}
{\sc Escande, A., Mansard, N. \& Wieber, P.B.} (2014). Hierarchical quadratic
  programming: Fast online humanoid-robot motion generation. {\em I. J.
  Robotics Res.\/}, {\bf 33}, 1006--1028.

\bibitem[{Evans {\em et~al.\/}(2003)Evans, Redmond, Plakas, Hamilton \&
  Lane}]{IntroAlive1}
{\sc Evans, J., Redmond, P., Plakas, C., Hamilton, K. \& Lane, D.} (2003).
  Autonomous docking for {Intervention-AUVs} using sonar and video-based
  real-time {3D} pose estimation. In {\em Oceans 2003. Celebrating the Past ...
  Teaming Toward the Future (IEEE Cat. No.03CH37492)\/}, vol.~4, 2201--2210
  Vol.4.

\bibitem[{Evans {\em et~al.\/}(2001)Evans, Keller, Smith, Marty \&
  Rigaud}]{IntroSwimmer1}
{\sc Evans, J.C., Keller, K.M., Smith, J.S., Marty, P. \& Rigaud, O.V.} (2001).
  {Docking techniques and evaluation trials of the SWIMMER AUV: an autonomous
  deployment AUV for work-class ROVs}. In {\em MTS/IEEE Oceans 2001. An Ocean
  Odyssey. Conference Proceedings (IEEE Cat. No.01CH37295)\/}, vol.~1, 520--528
  vol.1.

\bibitem[{Faverjon \& Tournassoud(1987)}]{IntroTpik25}
{\sc Faverjon, B. \& Tournassoud, P.} (1987). A local based approach for path
  planning of manipulators with a high number of degrees of freedom. In {\em
  Proceedings. 1987 IEEE International Conference on Robotics and
  Automation\/}, vol.~4, 1152--1159.

\bibitem[{Fischler \& Bolles(1981)}]{ransacHomog}
{\sc Fischler, M.A. \& Bolles, R.C.} (1981). Random sample consensus: A
  paradigm for model fitting with applications to image analysis and automated
  cartography. {\em Commun. ACM\/}, {\bf 24}, 381--395.

\bibitem[{Flacco {\em et~al.\/}(2012)Flacco, De~Luca \& Khatib}]{IntroTpik10}
{\sc Flacco, F., De~Luca, A. \& Khatib, O.} (2012). Prioritized multi-task
  motion control of redundant robots under hard joint constraints. 3970--3977.

\bibitem[{Fletcher(2000)}]{IntroSpecApp2}
{\sc Fletcher, B.} (2000). Worldwide undersea {MCM} vehicle technologies. 10.

\bibitem[{Gilmour {\em et~al.\/}(2012)Gilmour, Niccum \&
  O'Donnell}]{IntroSpecApp5}
{\sc Gilmour, B., Niccum, G. \& O'Donnell, T.} (2012). Field resident {AUV}
  systems — chevron's long-term goal for {AUV} development. In {\em 2012
  IEEE/OES Autonomous Underwater Vehicles (AUV)\/}, 1--5.

\bibitem[{Guennebaud {\em et~al.\/}(2010)Guennebaud, Jacob {\em
  et~al.\/}}]{eigen}
{\sc Guennebaud, G., Jacob, B. {\em et~al.\/}} (2010). Eigen v3 [online;
  accessed 29-june-2019].

\bibitem[{Kanoun {\em et~al.\/}(2011)Kanoun, Lamiraux \& Wieber}]{IntroTpik26}
{\sc Kanoun, O., Lamiraux, F. \& Wieber, P.} (2011). Kinematic control of
  redundant manipulators: Generalizing the task-priority framework to
  inequality task. {\em IEEE Transactions on Robotics\/}, {\bf 27}, 785--792.

\bibitem[{Kermorgant(2014)}]{freeFloatingGazebo}
{\sc Kermorgant, O.} (2014). {A dynamic simulator for underwater
  vehicle-manipulators}. In {\em {International Conference on Simulation,
  Modeling, and Programming for Autonomous Robots Simpar}\/}, {Springer},
  Bergamo, Italy.

\bibitem[{Khatib(1986)}]{IntroTpik7}
{\sc Khatib, O.} (1986). Real-time obstacle avoidance for manipulators and
  mobile robots. {\em The International Journal of Robotics Research\/}, {\bf
  5}, 90--98.

\bibitem[{Khatib(1987)}]{IntroTpik2}
{\sc Khatib, O.} (1987). A unified approach for motion and force control of
  robot manipulators: The operational space formulation. {\em IEEE Journal on
  Robotics and Automation\/}, {\bf 3}, 43--53.

\bibitem[{Lane {\em et~al.\/}(1997)Lane, Davies, Casalino, Bartolini, Cannata,
  Veruggio, Canals, Smith, O'Brien, Pickett, Robinson, Jones, Scott, Ferrara,
  Angelleti, Coccoli, Bono, Virgili, Pallas \& Gracia}]{IntroAMADEUS1}
{\sc Lane, D.M., Davies, J.B.C., Casalino, G., Bartolini, G., Cannata, G.,
  Veruggio, G., Canals, M., Smith, C., O'Brien, D.J., Pickett, M., Robinson,
  G., Jones, D., Scott, E., Ferrara, A., Angelleti, D., Coccoli, M., Bono, R.,
  Virgili, P., Pallas, R. \& Gracia, E.} (1997). {AMADEUS}: advanced
  manipulation for deep underwater sampling. {\em IEEE Robotics Automation
  Magazine\/}, {\bf 4}, 34--45.

\bibitem[{Lane {\em et~al.\/}(2012)Lane, Maurelli, Kormushev, Carreras, Fox \&
  Kyriakopoulos}]{IntroPandora1}
{\sc Lane, D.M., Maurelli, F., Kormushev, P., Carreras, M., Fox, M. \&
  Kyriakopoulos, K.} (2012). Persistent autonomy: the challenges of the
  {PANDORA} project. {\em IFAC Proceedings Volumes\/}, {\bf 45}, 268 -- 273,
  9th IFAC Conference on Manoeuvring and Control of Marine Craft.

\bibitem[{Lee \& Park(2014)}]{IntroPeg11}
{\sc Lee, H. \& Park, J.} (2014). An active sensing strategy for contact
  location without tactile sensors using robot geometry and kinematics. {\em
  Autonomous Robots\/}, {\bf 36}, 109--121.

\bibitem[{Lee {\em et~al.\/}(2012)Lee, Mansard \& Park}]{IntroTpik21}
{\sc Lee, J., Mansard, N. \& Park, J.} (2012). Intermediate desired value
  approach for task transition of robots in kinematic control. {\em IEEE
  Transactions on Robotics\/}, {\bf 28}, 1260--1277.

\bibitem[{Lowe(2004)}]{loweTest}
{\sc Lowe, D.G.} (2004). Distinctive image features from scale-invariant
  keypoints. {\em International Journal of Computer Vision\/}, {\bf 60},
  91--110.

\bibitem[{Lozano-P\'{e}rez {\em et~al.\/}(1984)Lozano-P\'{e}rez, Mason \&
  Taylor}]{IntroPeg13}
{\sc Lozano-P\'{e}rez, T., Mason, M.T. \& Taylor, R.H.} (1984). Automatic
  synthesis of fine-motion strategies for robots. {\em The International
  Journal of Robotics Research\/}, {\bf 3}, 3--24.

\bibitem[{Maciejewski \& Klein(1985)}]{IntroTpik5}
{\sc Maciejewski, A.A. \& Klein, C.A.} (1985). Obstacle avoidance for
  kinematically redundant manipulators in dynamically varying environments.
  {\em The International Journal of Robotics Research\/}, {\bf 4}, 109--117.

\bibitem[{Mansard {\em et~al.\/}(2009{\natexlab{a}})Mansard, Khatib \&
  Kheddar}]{IntroTpik23}
{\sc Mansard, N., Khatib, O. \& Kheddar, O.} (2009{\natexlab{a}}). A unified
  approach to integrate unilateral constraints in the stack of tasks. {\em IEEE
  Transactions on Robotics\/}, {\bf 25}, 670--685.

\bibitem[{Mansard {\em et~al.\/}(2009{\natexlab{b}})Mansard, Remazeilles \&
  Chaumette}]{IntroTpik22}
{\sc Mansard, N., Remazeilles, A. \& Chaumette, F.} (2009{\natexlab{b}}).
  Continuity of varying-feature-set control laws. {\em IEEE Transactions on
  Automatic Control\/}, {\bf 54}, 2493--2505.

\bibitem[{Marani {\em et~al.\/}(2003)Marani, Kim, Yuh \& Chung}]{IntroTpik11}
{\sc Marani, G., Kim, J., Yuh, J. \& Chung, W.} (2003). Algorithmic
  singularities avoidance in task-priority based controller for redundant
  manipulators. vol.~4, 3570 -- 3574 vol.3.

\bibitem[{Marani {\em et~al.\/}(2009)Marani, Choi \& Yuh}]{IntroSauvim2}
{\sc Marani, G., Choi, S.K. \& Yuh, J.} (2009). Underwater autonomous
  manipulation for intervention missions {AUVs}. {\em Ocean Engineering\/},
  {\bf 36}, 15 -- 23, autonomous Underwater Vehicles.

\bibitem[{Marchand {\em et~al.\/}(2005)Marchand, Spindler \& Chaumette}]{visp}
{\sc Marchand, E., Spindler, F. \& Chaumette, F.} (2005). {ViSP} for visual
  servoing: a generic software platform with a wide class of robot control
  skills. {\em IEEE Robotics and Automation Magazine\/}, {\bf 12}, 40--52.

\bibitem[{Marty(2004)}]{IntroAlive2}
{\sc Marty, P.} (2004). {ALIVE}: An autonomous light intervention vehicle. {\em
  Scandinavian Oil-Gas Magazine\/}, {\bf 32}.

\bibitem[{Matas {\em et~al.\/}(2000)Matas, Galambos \& Kittler}]{houghprob}
{\sc Matas, J., Galambos, C. \& Kittler, J.} (2000). Robust detection of lines
  using the progressive probabilistic hough transform. {\em Computer Vision and
  Image Understanding\/}, {\bf 78}, 119--137.

\bibitem[{Michel(2004)}]{webots}
{\sc Michel, O.} (2004). Webots: Professional mobile robot simulation. {\em
  Journal of Advanced Robotics Systems\/}, {\bf 1}, 39--42.

\bibitem[{Miura \& Ikeuchi(1998)}]{IntroPeg2}
{\sc Miura, J. \& Ikeuchi, K.} (1998). Task-oriented generation of visual
  sensing strategies in assembly tasks. {\em Pattern Analysis and Machine
  Intelligence, IEEE Transactions on\/}, {\bf 20}, 126 -- 138.

\bibitem[{Muja \& Lowe(2012)}]{flannMatch}
{\sc Muja, M. \& Lowe, D.G.} (2012). Fast matching of binary features. In {\em
  Computer and Robot Vision {(CRV)}\/}, 404--410.

\bibitem[{Nakamura(1990)}]{IntroTpik3}
{\sc Nakamura, Y.} (1990). {\em Advanced Robotics: Redundancy and
  Optimization\/}. Addison-Wesley Longman Publishing Co., Inc., Boston, MA,
  USA, 1st edn.

\bibitem[{Nakamura \& Hanafusa(1986)}]{IntroTpik1}
{\sc Nakamura, Y. \& Hanafusa, H.} (1986). Inverse kinematic solutions with
  singularity robustness for robot manipulator control. {\em Journal of Dynamic
  Systems, Measurement, and Control\/}.

\bibitem[{Nenchev \& Sotirov(1994)}]{IntroTpik28}
{\sc Nenchev, D.N. \& Sotirov, Z.M.} (1994). Dynamic task-priority allocation
  for kinematically redundant robotic mechanisms. In {\em Proceedings of
  IEEE/RSJ International Conference on Intelligent Robots and Systems
  (IROS'94)\/}, vol.~1, 518--524 vol.1.

\bibitem[{Newman {\em et~al.\/}(2001)Newman, Zhao \& Pao}]{IntroPeg7}
{\sc Newman, W., Zhao, Y. \& Pao, Y.H.} (2001). Interpretation of force and
  moment signals for compliant peg-in-hole assembly. vol.~1, 571 -- 576 vol.1.

\bibitem[{Oh \& Oh(2015)}]{IntroPeg6}
{\sc Oh, J. \& Oh, J.H.} (2015). A modified perturbation/correlation method for
  force-guided assembly. {\em Journal of Mechanical Science and Technology\/},
  {\bf 29}, 5437--5446.

\bibitem[{Padir(2005)}]{IntroTpik15}
{\sc Padir, T.} (2005). Kinematic redundancy resolution for two cooperating
  underwater vehicles with on-board manipulators. vol.~4, 3137 -- 3142 Vol. 4.

\bibitem[{Paravisi {\em et~al.\/}(2019)Paravisi, H.~Santos, Jorge, Heck,
  Gonçalves \& Amory}]{usvsim}
{\sc Paravisi, M., H.~Santos, D., Jorge, V., Heck, G., Gonçalves, L.M. \&
  Amory, A.} (2019). Unmanned surface vehicle simulator with realistic
  environmental disturbances. {\em Sensors\/}, {\bf 19}.

\bibitem[{Park {\em et~al.\/}(2013)Park, Bae, Park, Baeg \& Park}]{IntroPeg12}
{\sc Park, H., Bae, J.H., Park, J.H., Baeg, M.H. \& Park, J.} (2013). Intuitive
  peg-in-hole assembly strategy with a compliant manipulator. In {\em IEEE ISR
  2013\/}, 1--5.

\bibitem[{Park {\em et~al.\/}(2017)Park, Park, Lee, Park, Baeg \&
  Bae}]{IntroPeg15}
{\sc Park, H., Park, J., Lee, D.H., Park, J.H., Baeg, M.H. \& Bae, J.H.}
  (2017). Compliance-based robotic peg-in-hole assembly strategy without force
  feedback. {\em IEEE Transactions on Industrial Electronics\/}, {\bf PP},
  1--1.

\bibitem[{Pauli {\em et~al.\/}(2001)Pauli, Schmidt \& Sommer}]{IntroPeg1}
{\sc Pauli, J., Schmidt, A. \& Sommer, G.} (2001). Vision-based integrated
  system for object inspection and handling. {\em Robotics and Autonomous
  Systems\/}, {\bf 37}, 297 -- 309.

\bibitem[{Perez \& Fossen(2007)}]{fossenAnglesSeq}
{\sc Perez, T. \& Fossen, T.I.} (2007). Kinematic models for manoeuvring and
  seakeeping of marine vessels. {\em Modeling, Identification and Control\/},
  {\bf 28}, 19--30.

\bibitem[{{Prats} {\em et~al.\/}(2012){Prats}, {Pérez}, {Fernández} \&
  {Sanz}}]{uwsim}
{\sc {Prats}, M., {Pérez}, J., {Fernández}, J.J. \& {Sanz}, P.J.} (2012). An
  open source tool for simulation and supervision of underwater intervention
  missions. In {\em 2012 IEEE/RSJ International Conference on Intelligent
  Robots and Systems\/}, 2577--2582.

\bibitem[{Prats {\em et~al.\/}(2012)Prats, Romagós, Palomeras,
  García~Sánchez, Nannen, Wirth, Fernandez, P.~Beltrán, Campos, Ridao, Sanz,
  Oliver, Carreras, Gracias, Marín~Prades \& Ortiz}]{IntroRauvi}
{\sc Prats, M., Romagós, D., Palomeras, N., García~Sánchez, J.C., Nannen,
  V., Wirth, S., Fernandez, J., P.~Beltrán, J., Campos, R., Ridao, P., Sanz,
  P., Oliver, G., Carreras, M., Gracias, N., Marín~Prades, R. \& Ortiz, A.}
  (2012). Reconfigurable {AUV} for intervention missions: A case study on
  underwater object recovery. {\em Intelligent Service Robotics\/}, {\bf 5},
  19--31.

\bibitem[{Pressigout \& Marchand(2007)}]{visp-klt}
{\sc Pressigout, M. \& Marchand, E.} (2007). Real-time hybrid tracking using
  edge and texture information. {\em The International Journal of Robotics
  Research\/}, {\bf 26}, 689--713.

\bibitem[{PROMETEO(2016)}]{IntroPrometeo}
{\sc PROMETEO} (2016). \url{http://www.irs.uji.es/prometeo/}, [online; accessed
  25-october-2018].

\bibitem[{Rigaud {\em et~al.\/}(1998)Rigaud, Coste-Maniere, Aldon, Probert,
  Perrier, Rives, Simon, Lang, Kiener, Casal, Amar, Dauchez \&
  Chantler}]{IntroUNION}
{\sc Rigaud, V., Coste-Maniere, E., Aldon, M.J., Probert, P., Perrier, M.,
  Rives, P., Simon, D., Lang, D., Kiener, J., Casal, A., Amar, J., Dauchez, P.
  \& Chantler, M.} (1998). {UNION}: underwater intelligent operation and
  navigation. {\em IEEE Robotics Automation Magazine\/}, {\bf 5}, 25--35.

\bibitem[{ROBUST(2016)}]{IntroRobust}
{\sc ROBUST} (2016). \url{http://eu-robust.eu}, [online; accessed
  25-october-2018].

\bibitem[{Rusu \& Cousins(2011)}]{pclLib}
{\sc Rusu, R.B. \& Cousins, S.} (2011). {3D is here: Point Cloud Library
  (PCL)}. In {\em {IEEE International Conference on Robotics and Automation
  (ICRA)}\/}, Shanghai, China.

\bibitem[{Sanz {\em et~al.\/}(2012)Sanz, Ridao, Oliver, Casalino, Insaurralde,
  Silvestre, Melchiorri \& Turetta}]{IntroTrident1}
{\sc Sanz, P., Ridao, P., Oliver, G., Casalino, G., Insaurralde, C., Silvestre,
  C., Melchiorri, C. \& Turetta, A.} (2012). {TRIDENT}: Recent improvements
  about autonomous underwater intervention missions. vol.~3, 1--10.

\bibitem[{Schempf \& Yoerger(1992)}]{IntroTeleopRov}
{\sc Schempf, H. \& Yoerger, D.R.} (1992). Coordinated vehicle/manipulation
  design and control issues for underwater telemanipulation. {\em IFAC
  Proceedings Volumes\/}, {\bf 25}, 259 -- 267, iFAC Workshop on Artificial
  Intelligence Control and Advanced Technology in Marine Automation (CAMS '92),
  Genova, Italy, April 8-10.

\bibitem[{Sentis \& Khatib(2005)}]{IntroTpik18}
{\sc Sentis, L. \& Khatib, O.} (2005). Control of free-floating humanoid robots
  through task prioritization. In {\em Proceedings of the 2005 IEEE
  International Conference on Robotics and Automation\/}, 1718--1723.

\bibitem[{Shi \& Tomasi(2000)}]{Shi94goodfeatures}
{\sc Shi, J. \& Tomasi, C.} (2000). Good features to track. {\em Proceedings /
  CVPR, IEEE Computer Society Conference on Computer Vision and Pattern
  Recognition\/}, {\bf 600}, 593–600.

\bibitem[{Shirinzadeh {\em et~al.\/}(2011)Shirinzadeh, Zhong, Tilakaratna, Tian
  \& Dalvand}]{IntroPeg4}
{\sc Shirinzadeh, B., Zhong, Y., Tilakaratna, P.D.W., Tian, Y. \& Dalvand,
  M.M.} (2011). A hybrid contact state analysis methodology for robotic-based
  adjustment of cylindrical pair. {\em The International Journal of Advanced
  Manufacturing Technology\/}, {\bf 52}, 329--342.

\bibitem[{Siciliano \& Slotine(1991)}]{IntroTpik4}
{\sc Siciliano, B. \& Slotine, J..E.} (1991). A general framework for managing
  multiple tasks in highly redundant robotic systems. In {\em Fifth
  International Conference on Advanced Robotics 'Robots in Unstructured
  Environments\/}, 1211--1216 vol.2.

\bibitem[{Siciliano {\em et~al.\/}(2009)Siciliano, Sciavicco, Villani \&
  Oriolo}]{bookSiciliano}
{\sc Siciliano, B., Sciavicco, L., Villani, L. \& Oriolo, G.} (2009). {\em
  Robotics: Modelling, Planning and Control\/}, 147--151. Springer-Verlag
  London.

\bibitem[{Simetti \& Casalino(2016)}]{IntroMaris1}
{\sc Simetti, E. \& Casalino, G.} (2016). A novel practical technique to
  integrate inequality control objectives and task transitions in priority
  based control. {\em Journal of Intelligent \& Robotic Systems\/}, {\bf 84}.

\bibitem[{Simetti \& Casalino(2017)}]{IntroMaris2}
{\sc Simetti, E. \& Casalino, G.} (2017). Manipulation and transportation with
  cooperative underwater vehicle manipulator systems. {\em IEEE Journal of
  Oceanic Engineering\/}, {\bf 42}, 782--799.

\bibitem[{Simetti {\em et~al.\/}(2009)Simetti, Turetta \&
  Casalino}]{IntroTpik16}
{\sc Simetti, E., Turetta, A. \& Casalino, G.} (2009). {\em Distributed Control
  and Coordination of Cooperative Mobile Manipulator Systems\/}, 315--324.
  Springer Berlin Heidelberg, Berlin, Heidelberg.

\bibitem[{Simetti {\em et~al.\/}(2014{\natexlab{a}})Simetti, Casalino, Torelli,
  Sperind\'e \& Turetta}]{IntroTrident4}
{\sc Simetti, E., Casalino, G., Torelli, S., Sperind\'e, A. \& Turetta, A.}
  (2014{\natexlab{a}}). Floating underwater manipulation: Developed control
  methodology and experimental validation within the {TRIDENT} project. {\em
  Journal of Field Robotics\/}, {\bf 31(3)}, 364--385.

\bibitem[{Simetti {\em et~al.\/}(2014{\natexlab{b}})Simetti, Casalino, Torelli,
  Sperind\'e \& Turetta}]{IntroTpik30}
{\sc Simetti, E., Casalino, G., Torelli, S., Sperind\'e, A. \& Turetta, A.}
  (2014{\natexlab{b}}). Underwater floating manipulation for robotic
  interventions. {\em IFAC Proceedings Volumes\/}, {\bf 47}, 3358 -- 3363, 19th
  IFAC World Congress.

\bibitem[{Simetti {\em et~al.\/}(2018)Simetti, Casalino, Wanderlingh \&
  Aicardi}]{IntroRecent}
{\sc Simetti, E., Casalino, G., Wanderlingh, F. \& Aicardi, M.} (2018). Task
  priority control of underwater intervention systems: Theory and applications.
  {\em Ocean Engineering\/}, {\bf 164}, 40 -- 54.

\bibitem[{Song {\em et~al.\/}(2016)Song, Kim \& Song}]{IntroPeg5}
{\sc Song, H., Kim, Y. \& Song, J.B.} (2016). Guidance algorithm for
  complex-shape peg-in-hole strategy based on geometrical information and force
  control. {\em Advanced Robotics\/}, 1--12.

\bibitem[{Sugiura {\em et~al.\/}(2007)Sugiura, Gienger, Janssen \&
  Goerick}]{IntroTpik19}
{\sc Sugiura, H., Gienger, M., Janssen, H. \& Goerick, C.} (2007). Real-time
  collision avoidance with whole body motion control for humanoid robots. In
  {\em 2007 IEEE/RSJ International Conference on Intelligent Robots and
  Systems\/}, 2053--2058.

\bibitem[{Suzuki \& Be(1985)}]{findcountors}
{\sc Suzuki, S. \& Be, K.A.} (1985). Topological structural analysis of
  digitized binary images by border following. {\em Computer Vision, Graphics,
  and Image Processing\/}, {\bf 30}, 32 -- 46.

\bibitem[{Trinh {\em et~al.\/}(2018)Trinh, Spindler, Marchand \&
  Chaumette}]{visp-depth}
{\sc Trinh, S., Spindler, F., Marchand, E. \& Chaumette, F.} (2018). A modular
  framework for model-based visual tracking using edge, texture and depth
  features. In {\em {IEEE/RSJ Int. Conf. on Intelligent Robots and Systems,
  IROS'18}\/}, Madrid, Spain.

\bibitem[{TWINBOT(2019)}]{TWINBOT2019}
{\sc TWINBOT} (2019). \url{http://www.irs.uji.es/twinbot/twinbot.html},
  [online; accessed 29-june-2019].

\bibitem[{Urabe {\em et~al.\/}(2015)Urabe, Ura, Tsujimoto \& Hotta}]{IntroApp2}
{\sc Urabe, T., Ura, T., Tsujimoto, T. \& Hotta, H.} (2015). Next-generation
  technology for ocean resources exploration (zipangu-in-the-ocean) project in
  japan. 1--5.

\bibitem[{Wanderlingh(2018)}]{tesiWander}
{\sc Wanderlingh, F.} (2018). {\em \emph{Cooperative Robotic Manipulation for
  the Smart Factory}\/}. Ph.D. thesis, \emph{Universit\`{a} degli Studi di
  Genova}.

\bibitem[{Wynn {\em et~al.\/}(2014)Wynn, Huvenne, Le~Bas, Murton, Connelly,
  Bett, Ruhl, Morris, Peakall, Parsons, J.~Sumner, E.~Darby, Dorrell \&
  Hunt}]{IntroApp1}
{\sc Wynn, R., Huvenne, V., Le~Bas, T., Murton, B., Connelly, D., Bett, B.,
  Ruhl, H., Morris, K., Peakall, J., Parsons, D., J.~Sumner, E., E.~Darby, S.,
  Dorrell, R. \& Hunt, J.} (2014). Autonomous underwater vehicles ({AUVs}):
  Their past, present and future contributions to the advancement of marine
  geoscience. {\em Marine Geology\/}, {\bf 352}.

\bibitem[{Xu(2015)}]{IntroPeg14}
{\sc Xu, Q.} (2015). Robust impedance control of a compliant microgripper for
  high-speed position/force regulation. {\em IEEE Transactions on Industrial
  Electronics\/}, {\bf 62}, 1201--1209.

\bibitem[{Yoshikawa(1984)}]{IntroTpik6}
{\sc Yoshikawa, T.} (1984). {Analysis and Control of Robot Manipulators with
  Redundancy}. In M.~Brady \& R.~Paul, eds., {\em Robotics Research The First
  International Symposium\/}, 735--747, MIT Press.

\bibitem[{Yuh {\em et~al.\/}(1998)Yuh, Choi, Ikehara, Kim, McMurty,
  Ghasemi-Nejhad, Sarkar \& Sugihara}]{IntroSauvim1}
{\sc Yuh, J., Choi, S.K., Ikehara, C., Kim, G.H., McMurty, G., Ghasemi-Nejhad,
  M., Sarkar, N. \& Sugihara, K.} (1998). Design of a semi-autonomous
  underwater vehicle for intervention missions {(SAUVIM)}. In {\em Proceedings
  of 1998 International Symposium on Underwater Technology\/}, 63--68.

\bibitem[{Zereik {\em et~al.\/}(2011)Zereik, Sorbara, Merlo, Simetti, Casalino
  \& Didot}]{IntroTpik14}
{\sc Zereik, E., Sorbara, A., Merlo, A., Simetti, E., Casalino, G. \& Didot,
  F.} (2011). Space robotics supporting exploration missions: vision, force
  control and coordination strategy for crew assistants. {\em Intelligent
  Service Robotics\/}, {\bf 4}, 39--60.

\end{thebibliography}
\addcontentsline{toc}{chapter}{References}    %

\end{document}